\documentclass{article}

\usepackage{etoolbox}
\newtoggle{arxiv}
\toggletrue{arxiv}

\iftoggle{arxiv}{}
{
\PassOptionsToPackage{numbers}{natbib}
\usepackage[final]{neurips_2021}
}

\usepackage[utf8]{inputenc} %
\usepackage[T1]{fontenc}    %
\usepackage{hyperref}       %
\usepackage{url}            %
\usepackage{booktabs}       %
\usepackage{amsfonts}       %
\usepackage{nicefrac}       %
\usepackage{microtype}      %
\usepackage{xcolor}         %
\usepackage{amsmath,amsthm,amssymb}

\usepackage{subcaption}

\usepackage[capitalise]{cleveref}  %
\usepackage{comment}
\usepackage[inline]{enumitem}

\usepackage{graphicx}
\usepackage{grffile}
\usepackage{wrapfig,epsfig}
\usepackage{epstopdf}
\usepackage{algpseudocode}
\usepackage{multirow}
\usepackage[T1]{fontenc}
\usepackage{bbm}
\usepackage{comment}
\usepackage{dsfont}
\usepackage{makecell}
\usepackage{enumitem}
\usepackage{booktabs}

\usepackage{amsmath}
\usepackage{amsthm}
\usepackage{amssymb}
\usepackage{algorithm}
\usepackage{color}
\usepackage[english]{babel}
\usepackage{graphicx}
\usepackage{grffile}
\iftoggle{arxiv}{
\usepackage[numbers,sort]{natbib}
}{}
\usepackage{wrapfig,epsfig}
\usepackage{epstopdf}
\usepackage{algpseudocode}
\usepackage{multirow}
\usepackage[T1]{fontenc}
\usepackage{bbm}
\usepackage{comment}
\usepackage{dsfont}
\usepackage{makecell}
\usepackage{enumitem}
\usepackage{booktabs}
\usepackage{afterpage}

\usepackage{minitoc}

\usepackage{algorithm}
\usepackage{color}
\usepackage[english]{babel}

\usepackage{import}

\newtoggle{todo}
\togglefalse{todo} %

\newtoggle{comment}
\toggletrue{comment}

\newcommand{\R}{\mathbb{R}}
\newcommand{\rank}{\mathrm{rank}}
\newcommand{\abs}[1]{\left\lvert#1\right\rvert}
\newcommand{\softmax}{\mathrm{softmax}}
\newcommand{\supp}{\mathrm{supp}}

\newcommand{\norm}[1]{\left\|{#1}\right\|} %

\newcommand{\E}{\mathbb{E}} %
\renewcommand{\P}{\mathbb{P}} %
\newcommand{\var}{\mathbb{Var}} %

\newcommand{\defeq}{:=}

\newtheorem{theorem}{Theorem}
\newtheorem*{theorem*}{Theorem}
\newtheorem{corollary}[theorem]{Corollary}
\newtheorem{definition}{Definition}
\newtheorem{lemma}[theorem]{Lemma}

\newtheorem{example}{Example}
\newtheorem{matrixclass}{Matrix Class}
\newtheorem{process}{Process}

\def\eps{{\epsilon}}
\newcommand{\wh}{\widehat}
\newcommand{\wt}{\widetilde}

\renewcommand{\varepsilon}{\epsilon}
\renewcommand{\tilde}{\wt}
\renewcommand{\hat}{\wh}

\DeclareMathOperator{\diag}{diag}

\newcommand*\samethanks[1][\value{footnote}]{\footnotemark[#1]}
\iftoggle{arxiv}{
  \setlength{\textwidth}{6.5in}
  \setlength{\textheight}{9in}
  \setlength{\oddsidemargin}{0in}
  \setlength{\evensidemargin}{0in}
  \setlength{\topmargin}{-0.5in}
  \newlength{\defbaselineskip}
  \setlength{\defbaselineskip}{\baselineskip}
  \setlength{\marginparwidth}{0.8in}
}{

}

\title{Scatterbrain: Unifying Sparse and Low-rank Attention Approximation}
\iftoggle{arxiv}{
  \usepackage{authblk}
  \author[$\dagger$]{Beidi Chen\thanks{Equal contribution. Order determined by coin flip.}}
  \author[$\dagger$]{Tri Dao\samethanks}
  \author[$\dagger$]{Eric Winsor}
  \author[$\S$]{Zhao Song}
  \author[$\ddagger$]{Atri Rudra}
  \author[$\dagger$]{Christopher R{\'e}}
  \affil[$\dagger$]{Department of Computer Science, Stanford University}
  \affil[$\S$]{Adobe Research}
  \affil[$\ddagger$]{Department of Computer Science and Engineering, University at Buffalo, SUNY\vspace{4pt}}
  \affil[ ]{{\texttt{\{beidic,trid,winsor\}@stanford.edu}, \texttt{zsong@adobe.com}, \texttt{atri@buffalo.edu}, \texttt{chrismre@cs.stanford.edu}}}
}{

\author{%
  Beidi Chen\thanks{Equal contribution. Order determined by coin flip.}\: $^\dagger$,
  Tri Dao$^{* \dagger}$, Eric Winsor $^\dagger$, Zhao Song $^\S$, Atri Rudra $^\ddagger$, Christopher R\'{e} $^\dagger$\\
  $^\dagger$ Department of Computer Science, Stanford University\\
  $^\S$ Adobe Research\\
  $^\ddagger$ Department of Computer Science and Engineering, University at Buffalo, SUNY\\
  {\small\texttt{\{beidic,trid,winsor,chrismre\}@stanford.edu}, \texttt{zsong@adobe.com}, \texttt{atri@buffalo.edu}}
}
}

\begin{document}

\maketitle
\doparttoc %
\faketableofcontents 

\begin{abstract}
Recent advances in efficient Transformers have exploited either the sparsity or low-rank properties of attention matrices to reduce the computational and memory bottlenecks of modeling long sequences. However, it is still challenging to balance the trade-off between model quality and efficiency to perform a one-size-fits-all approximation for different tasks. To better understand this trade-off, we observe that sparse and low-rank approximations excel in different regimes, determined by the softmax temperature in attention, and sparse + low-rank can outperform each individually. Inspired by the classical robust-PCA algorithm for sparse and low-rank decomposition, we propose Scatterbrain, a novel way to unify sparse (via locality sensitive hashing) and low-rank (via kernel feature map) attention for accurate and efficient approximation. The estimation is unbiased with provably low error. We empirically show that Scatterbrain can achieve $2.1 \times$ lower error than baselines when serving as a drop-in replacement in BigGAN image generation and pre-trained T2T-ViT. On a pre-trained T2T Vision transformer, even without fine-tuning, Scatterbrain can reduce $98\%$ of attention memory at the cost of only $1\%$ drop in accuracy. We demonstrate Scatterbrain for end-to-end training with up to $4$ points better perplexity and 5 points better average accuracy than sparse or low-rank efficient transformers on language modeling and long-range-arena tasks.
\end{abstract}

\section{Introduction}
\label{sec:intro}

Transformer models~\citep{vaswani2017attention} have been adapted in a wide variety of applications, including natural language processing~\citep{devlin2018bert, brown2020language, raffel2019exploring}, image processing~\citep{carion2020end, parmar2018image}, and speech recognition~\citep{luo2021simplified}.
Training large Transformers requires extensive computational and memory resources, especially when modeling long sequences, mainly due to the quadratic complexity (w.r.t.\ sequence length) in attention layers. 
Recent advances in efficient transformers~\citep{kitaev2020reformer, choromanski2020rethinking, katharopoulos2020transformers, wang2020linformer, daras2020smyrf} leverage attention approximation to overcome the bottleneck by approximating the attention matrices.
However, it is challenging to find a robust approximation method that balances the efficiency-accuracy trade-off on a wide variety of tasks~\citep{tay2020long, tay2020efficient}.

We categorize most of the existing approaches for efficient attention matrix
computation into two major groups: exploiting either the sparsity, e.g.,\ Reformer~\citep{kitaev2020reformer}, SMYRF~\citep{daras2020smyrf}, or
low-rank properties of the attention matrices, e.g.,\ Linformer~\citep{wang2020linformer}, Linear Transformer~\citep{katharopoulos2020transformers}, and Performer~\citep{choromanski2020rethinking}. However, 
these techniques usually have different strengths and focus on the performance of specific tasks, so their approximations still cause accuracy degradation on many other tasks. For instance, according to a recent benchmark paper~\citep{tay2020long} and our experiments, low-rank-based attention might be less effective on hierarchically structured data or language modeling tasks, while sparse-based variants do not perform well on classification tasks.

\begin{figure}
    \begin{center}
	\scriptsize
		\begin{tabular}{cc}
		\hspace{-1cm}
    \includegraphics[width=0.32274\linewidth]{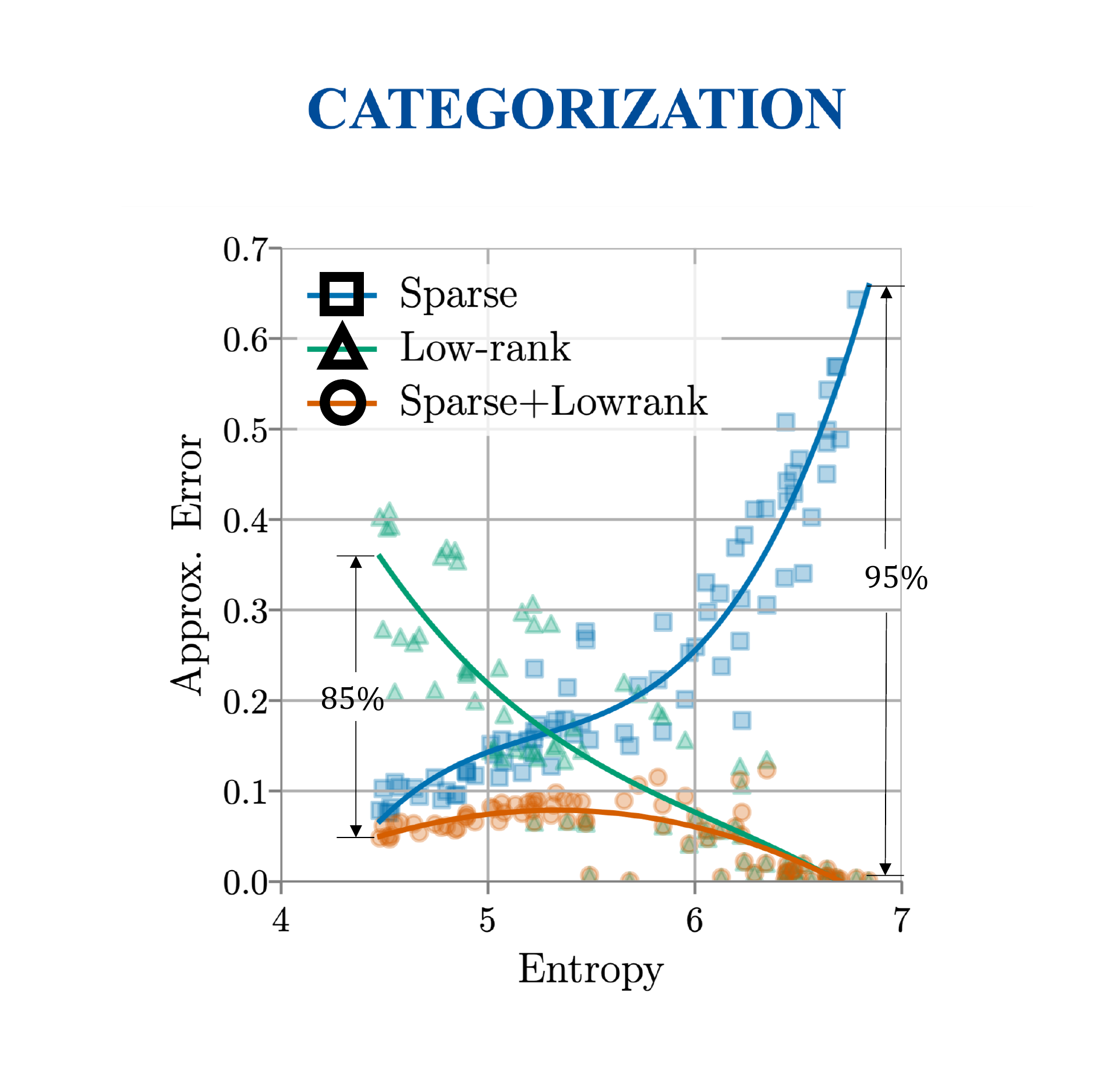}
    \hspace{-0.3cm}
    \includegraphics[width=0.7773\linewidth]{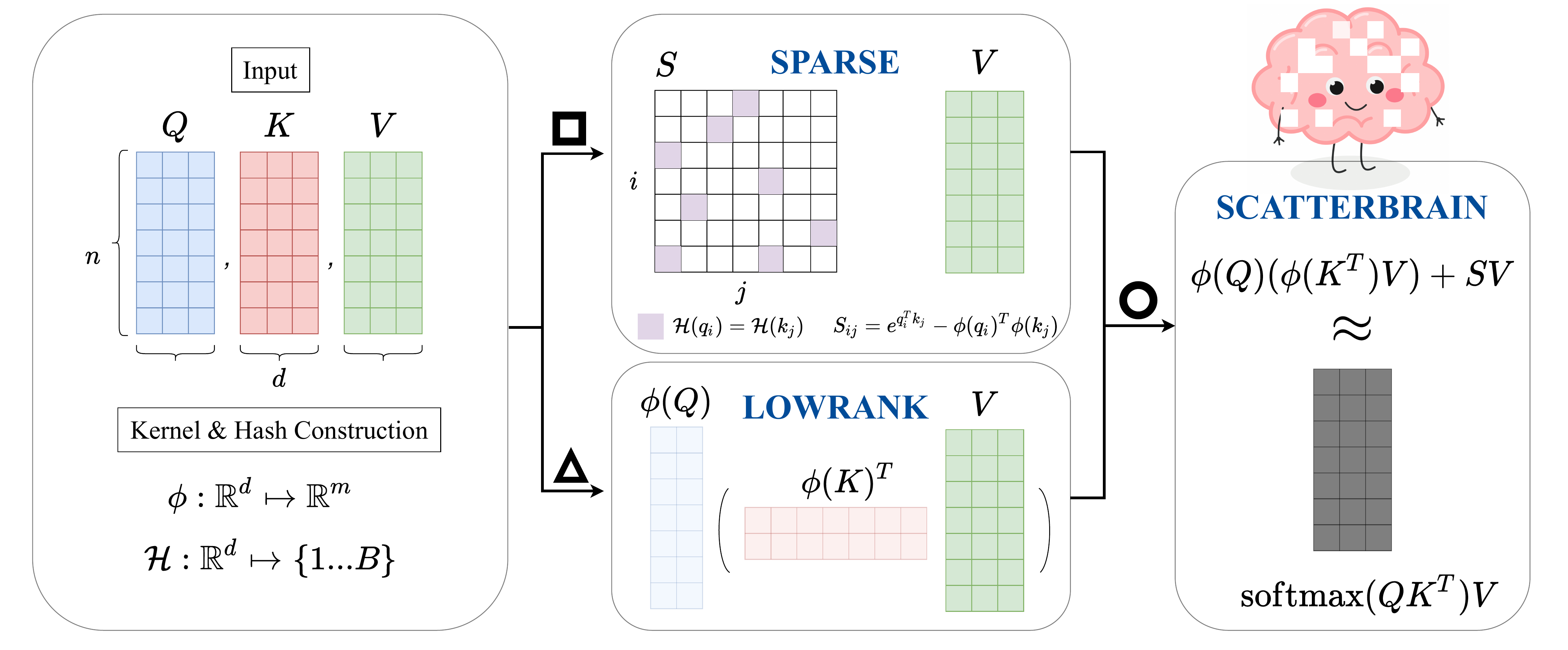}
		\end{tabular}
	\end{center}
    \caption{Left: regimes that sparse+low-rank approximation is more accurate, based on the entropy of the attention matrices. Right: Scatterbrain Workflow. For the attention layer in Transformers, after computing Query $Q$, Key $K$, and Value $V$ matrices, we approximate $\softmax(QK^\top)V$ with two components: (i) sparse $SV$ (ii) low-rank $\phi(Q)(\phi(K)^\top V)$.}
    \label{fig:workflow}
    \vspace{-5mm}
\end{figure}

We observe that sparse and low-rank approximations are complementary for many attention matrices in practice, and sparse+low-rank could outperform each individually (Figure~\ref{fig:workflow} left).
We empirically categorize the regimes in which sparse or low-rank approximation achieves better error based on the softmax temperature of attention (of which the entropy of softmax distribution can be used as a proxy).
We expect that sparse methods perform well if the attention depends on a few entries (low entropy softmax). In contrast, low-rank methods do better if the attention depends on a mixture of many components (high entropy softmax). This explains the phenomenon that current sparse and low-rank-based approaches excel on different kinds of tasks. A natural question is whether one could understand and unify the strength of both approaches. While it is NP-hard to find the optimal combination of sparse and low-rank approximations, Robust PCA~\citep{candes2011robust} is a polynomial-time solution with tight approximation error. 
We observe that Robust PCA achieves lower approximation error than sparse or low-rank alone on attention matrices. The difference is most pronounced for ``mid-range'' entropy, where we observe that up to $95\%$ error reduction is possible.

The connection between Robust PCA and attention matrix estimation provides an opportunity to realize a more robust approximation. Specifically, given an attention matrix, one could adaptively perform sparse+low-rank approximation to obtain a low error. However, it comes with three challenges: (i) How to decompose the attention matrices into sparse and low-rank components and estimate them efficiently and accurately; Robust PCA is accurate but slow and requires materializing the full attention, while straightforward addition of sparse and low-rank attention will be inaccurate due to double counting. (ii) It is not clear if there is a theoretical guarantee that sparse + low-rank approximation is strictly better than sparse or low-rank in some regimes, though we observe the separation empirically. (iii) How does the lower approximation error transfer to end-to-end performance in real tasks.

In this paper, we propose Scatterbrain, an accurate and efficient robust estimation of attention matrices with theoretical guarantees to address the above challenges. Specifically:
\begin{itemize}[leftmargin=*,nosep,nolistsep]
  \item In Section~\ref{sec:theory}, we observe that sparse and low-rank approximation are complementary and demonstrate that sparse + low-rank structure arises naturally when elements in the input sequence form clusters.
  We theoretically characterize and analyze the regimes where sparse, low-rank, and sparse+low-rank excel, dictated by the softmax temperature of attention.
  \item In Section~\ref{sec:algorithm}, inspired by the classical Robust PCA algorithm, we propose Scatterbrain, which efficiently combines sparse and low-rank matrices to approximate attention. In particular, we use Locality Sensitive Hashing (LSH) to identify large entries of the attention matrix (after softmax) without materializing the full matrix and then leverage kernel approximation to parameterize the low-rank part.
  We prove that our method has a strictly lower approximation error than the low-rank baseline.
\item In Section~\ref{sec:experiments}, we empirically validate our theory and the proposed method, showing that Scatterbrain accurately approximates the attention matrix, is memory efficient for long sequences, and works well across different tasks.
  First, we show that its approximation accuracy is close to our oracle Robust PCA and achieves 2.1$\times$ lower error compared to other efficient baselines on real benchmarks.
  This leads to a direct application of Scatterbrain as a drop-in replacement to pre-trained full attention, thus reducing up to $98\%$ of the memory required for attention computations in pre-trained T2T-ViT and BigGAN while maintaining similar quality.
  Last we show that its superior accuracy and efficiency can improve the efficiency-accuracy trade-offs of Transformer end-to-end training.
  On the WikiText-103 language modeling task, Scatterbrain achieves up to 1 point better perplexity compared to Reformer and Performer. On 5 benchmark long-range tasks, Scatterbrain improves the average accuracy by up to 5 points.$\footnote{Scatterbrain code is available at \url{https://github.com/HazyResearch/scatterbrain}}$
\end{itemize}

\section{Problem Setting and Related Work}
We first define the approximation problem we aim to solve in this paper. Then we discuss the applications of sparse and low-rank techniques in efficient Transformers and introduce robust PCA algorithm. 

\textbf{Problem Formulation:} In the attention matrix approximation problem, we are given three matrices, query, key, and value, $Q, K, V \in \R^{n \times d}$ to compute $\softmax(QK^\top)V$.
We seek to reduce the quadratic complexity of $\softmax(QK^\top)$ (applied row-wise) with low approximation error. 
More precisely, for an approximation procedure $f$, we minimize two objectives, the approximation error $\mathbf{E}\left[\Vert f(Q, K)-\softmax(QK^\top) \Vert_{F}^2\right]$, and the computation/memory cost $\mathcal{C}(f(\cdot))$.

\textbf{Sparse, Low-rank Approximation for Attention Matrices:} Recent work exploits the sparsity patterns or finds a low-rank mapping of the original attention matrices to overcome the computational and memory bottlenecks in Transformers~\citep{kitaev2020reformer,daras2020smyrf,roy2021efficient, choromanski2020rethinking, katharopoulos2020transformers, wang2020linformer}. Generally, we can divide most of the techniques into two categories -- sparse and low-rank approximations. Reformer~\citep{kitaev2020reformer} is a representative sparse variant that uses LSH~\citep{andoni2015practical} to retrieve or detect the locations of the attention matrices with large values and reduce the computation from $O(n^2)$ to $O(n \log n)$. Performer~\citep{choromanski2020rethinking} is an example of the low-rank variant, which uses kernelization to avoid explicit $O(n^2d)$ computation. One problem of either the sparse or low-rank approximation is that the structure of the attention matrices varies in practice, and it is challenging to perform robust approximation on a wide range of attention matrices. For example,~\citet{wang2020linformer} observes that attentions tend to have more low-rank structures in lower layers and~\citet{ramsauer2020hopfield} shows that they are sparser in the later stage of the training. Ideally, we want to unify the strength of both techniques, but it is NP-hard to find the best combination of sparse and low-rank approximation. 

\textbf{Sparse + Low-rank and Robust PCA:} Fortunately, classical Robust PCA~\citep{candes2011robust} presents a polynomial algorithm to find the approximately optimal or good combinations of sparse and low-rank approximation of the matrices. The sparse + low-rank matrix structure has been well studied in statistics and signal processing since the late 2000s~\citep{candes2011robust}. This structure naturally generalizes low-rank~\citep{hotelling1933analysis, udell2019big}, and sparse~\citep{tewarson1973sparse} matrices. Scatterbrain is built on a line of work, e.g., Bigbird~\citep{zaheer2020bigbird}, Longformer~\citep{beltagy2020longformer} with the theme of combining multiple types of attention. However, despite the multitude of papers, this sparse + low-rank matrix approximation has not been rigorously studied in the context of attention matrices. We undertake this study and show how we can relax the sparse + low-rank approximation from robust PCA, making it efficient while still retaining PCA’s accuracy. In fact, our results shed further light on why Bigbird or Longformer work, as they are special cases of a single principled structure. An extended discussion of related work is in Appendix~\ref{sec:extended_related_work}.

\section{Characterization of Sparse + Low-rank Approx. to Attention Matrices}
\label{sec:theory}

We motivate the use of sparse + low-rank approximation of the attention matrices
with the key observation that for many attention matrices, sparse and low-rank
approximation are complementary, and their ideal combination (via Robust PCA)
can outperform both (\cref{sec:observation}).
Furthermore, we argue that the sparse + low-rank structure
can arise naturally when elements in the input sequence form clusters, as
dictated by the softmax temperature (\cref{sec:generative}).

\subsection{Motivating Observations: Low-rank and Sparse Structures of Attention Matrices}
\label{sec:observation}
We empirically characterize regimes where sparse and low-rank approximation are well-suited, based on the softmax temperature (for which we use the softmax distribution entropy is a proxy). Specifically, in ~\cref{fig:workflow} (left), we present the approximation error of the original attention matrices and the approximation (sparse or low-rank) of matrices sampled from a 4-layer Transformer trained on IMDb reviews classification~\citep{tay2020long}.
We make two observations:
\begin{enumerate}[leftmargin=*,nosep,nolistsep]
  \item Sparse and low-rank approximation are complementary: sparse excels when the softmax temperature scale is low (i.e., low entropy), and low-rank excels when the softmax temperature is high (i.e., high entropy).
  \item An ideal combination of sparse and low-rank (orange line in \cref{fig:workflow} left), obtained with robust PCA, can achieve lower error than both.
\end{enumerate}
Similar observations on other benchmarks and details are presented in Appendix~\ref{sec:observation_details}.

\subsection{A Generative Model of How Sparse + Low-rank Structure Can Arise}
\label{sec:generative}
\begin{wrapfigure}{}{0.5\textwidth}
  \small
  \vspace{-1em}
  \centering
  \includegraphics[width=\linewidth]{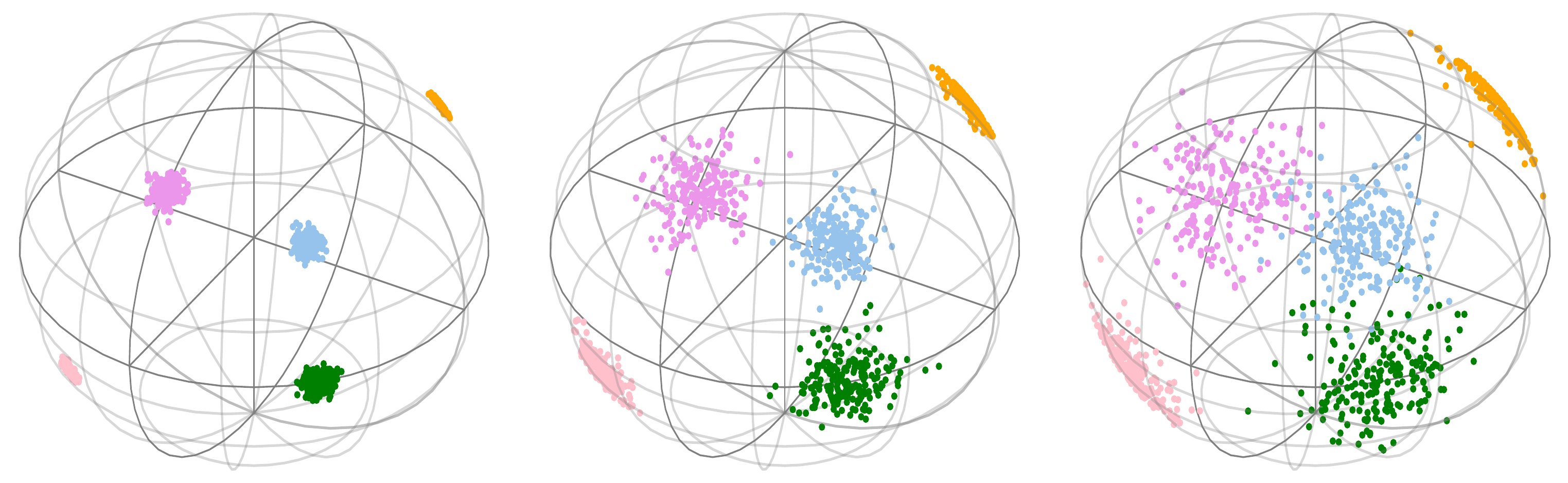}
  \caption{Visualization of the generative process, for three different values of the intra-cluster distance $\Delta$ (small, medium, and large). The vectors from the input sequence (rows of $Q$) form clusters that lie approximately on the unit sphere. Different colors represent different clusters.}
  \label{fig:generative_vis}
  \vspace{-1em}
\end{wrapfigure}
Sparse + low-rank parameterization is more expressive than either sparse or low-rank alone.
Indeed, in the Appendix, we construct a family of attention matrices to show the separation between the approximation capability of sparse + low-rank vs. sparse or low-rank alone: for an $n \times n$ attention matrix, sparse or low-rank alone requires a $O(n^2)$ parameters to get $\epsilon$ approximation error in Frobenius norm, while sparse + low-rank only requires $O(n)$ parameters.

Moreover, we argue here that sparse + low-rank is a natural candidate to approximate generic attention matrices.
We describe a generative model of how the sparse + low-rank structure in attention matrices could arise when the elements of the input sequence form clusters.
Under this process, we characterize how the softmax temperature dictates when we would need sparse, low-rank, or sparse + low-rank matrices to approximate the attention matrix.
This result corroborates the observation in \cref{sec:observation}.

\paragraph{Generative process of clustered elements in input sequence}

We describe here a generative model of an input sequence to attention, parameterized by the inverse temperature $\beta \in \mathbb{R}$ and the intra-cluster distance $\Delta \in \mathbb{R}$.
\begin{process}
  \label{ex:generative}
  Let $Q \in \mathbb{R}^{n \times d}$, where $d\geq\Omega(\log^{3/2}(n))$, with every row of $Q$ generated randomly as follows:
  \begin{enumerate}[leftmargin=*,nosep,nolistsep]
    \item For $C = \Omega(n)$, sample $C$ number of cluster centers $c_1, \dots, c_C \in \mathbb{R}^{d}$ independently from $\mathcal{N}(0, I_d/\sqrt{d})$.
    \item For each cluster around $c_i$, sample $n_i = O(1)$ number of elements around $c_i$, of the form $z_{ij} = c_i + r_{ij}$ for $j = 1, \dots, n_i$ where $r_{ij} \sim \mathcal{N}(0, I_d \Delta/\sqrt{d})$.
    Assume that the total number of elements is $n = n_1 + \dots + n_C$ and $\Delta\leq O(1/\log^{1/4} n)$.
  \end{enumerate}
  Let $Q$ be the matrix whose rows are the vectors $z_{ij}$ where $i = 1, \dots, C$ and $j = 1, \dots, n_i$.
  Let $A = Q Q^\top$ and let the attention matrix be $M_\beta = \exp(\beta \cdot A)$.
\end{process}
We visualize this generative process in~\cref{fig:generative_vis}.

\paragraph{Softmax temperature and approx.\ error}

We characterize when to use sparse, low-rank, or sparse + low-rank to approximate the attention matrices in \cref{ex:generative}, depending on the inverse temperature $\beta$.
The intuition here is that the inverse temperature corresponds to the strength of interaction between the clusters.
If $\beta$ is large, intra-cluster interaction dominates the attention matrix, the softmax distribution is peaked, and so we only need a sparse matrix to approximate the attention.
If $\beta$ is small, then the inter-cluster attention is similar to intra-cluster attention, the softmax distribution is diffuse, and we can approximate it with a low-rank matrix.
In the middle regime of $\beta$, we need the sparse part to cover the intra-cluster attention and the low-rank part to approximate the inter-cluster attention.

We formalize this intuition in \cref{thm:temperature} (in bounds below we think of $\epsilon$ as a constant).
All the proofs are in Appendix~\ref{sec:proofs}.
\begin{theorem}
  \label{thm:temperature}
  Let $M_\beta$, be the attention matrix in~\cref{ex:generative}. Fix $\epsilon\in (0,1)$. Let $R \in \mathbb{R}^{n \times n}$ be a matrix.
  Consider low-rank, sparse, and sparse + low-rank approximations to $M_\beta$.
  \begin{enumerate}[leftmargin=*,nosep,nolistsep]
    \item \textbf{High temperature}: Assume $\beta=o ( {\log n}/{\log d} )$.
    \begin{enumerate}
        \item \textbf{Low-rank}: There exists $R$ with $n^{o(1)}$ rank (and hence $n^{1+o(1)}$ parameters) such that $\|M_\beta-R\|_F\leq \eps n$.
        \item \textbf{Sparse}: If $R$ has sparsity $o(n^2)$, then $\|M_\beta-R\|_F \geq \Omega(n)$.
    \end{enumerate}

    \item \textbf{Mid temperature}: Assume $(1 - \Delta^2) \log n  \leq \beta \leq O(\log n)$.
    \begin{enumerate}
        \item \textbf{Sparse + low-rank}: There exists a sparse + low-rank $R$ with $n^{1+o(1)}$ parameters with $\|M_\beta-R\|_F\leq \eps n$.
        \item \textbf{Low-rank}: If $R$ is such that $n-\rank(R)=\Omega(n)$, then $\|M_\beta-R\|_F\geq \Omega(n)$.
        \item \textbf{Sparse}: If $R$ has sparsity $o(n^2)$, then $\|M_\beta - R\|_F\geq \Omega(n)$.
    \end{enumerate}

    \item \textbf{Low temperature}: Assume $\beta = \Omega(\log n)$.
    \begin{enumerate}
        \item \textbf{Low-rank}: If $R$ is such that $n-\rank(R) = \Omega(n)$, then $\|M_\beta-R\|_F\geq \Omega(e^{\beta(1-\Delta^2)})$.
        \item \textbf{Sparse}: There exists $R$ with sparsity $O(n)$ such that $\|M_\beta-R\|_F\leq\eps\cdot e^{\beta(1-\Delta^2)}$
    \end{enumerate}

  \end{enumerate}
\end{theorem}

\section{Scatterbrain: Unifying Sparse and Low-rank Attention}
\label{sec:algorithm}

\begin{wrapfigure}{}{0.4\textwidth}
  \small
  \iftoggle{arxiv}{}{
    \vspace{-2em}
  }
  \centering
  \includegraphics[width=\linewidth]{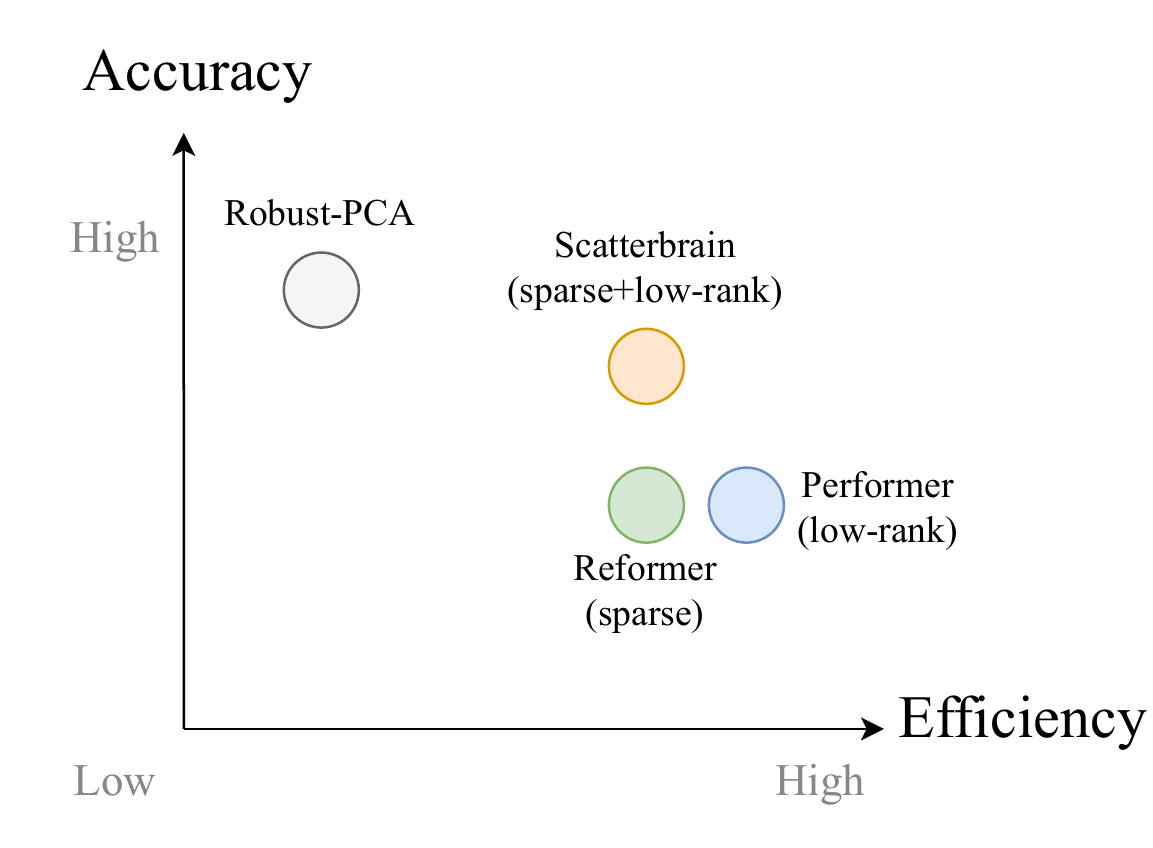}
  \caption{Qualitative comparison of approx.\ accuracy and efficiency, among Robust PCA, sparse (Reformer) and low-rank (Performer) attention, and Scatterbrain.
  Scatterbrain is more accurate while being efficient.}
  \label{fig:qualitative_comparison}
  \iftoggle{arxiv}{}{
    \vspace{-2em}
  }
\end{wrapfigure}
We present Scatterbrain, and show that it approximates attention accurately and efficiently.
\cref{sec:challenges} describes the challenges of designing an accurate and efficient approximation, and how obvious baselines such as Robust PCA or a simple combination of sparse attention and low-rank attention fail to meet both criteria.
\cref{sec:algo_description} demonstrates how Scatterbrain address the challenges (\cref{fig:workflow} contains a schematic of Scatterbrain).
In~\cref{sec:algo_analysis}, we show that Scatterbrain is unbiased with provably lower variance than low-rank baselines such as Performer. 

\cref{fig:qualitative_comparison} shows a qualitative comparison between different methods of approximating the attention matrix: Robust PCA is accurate but slow, sparse (e.g., Reformer), and low-rank (e.g., Performer) attention are fast and memory-efficient but may not be very accurate, while Scatterbrain is more accurate than its sparse and low-rank counterparts while remaining just as efficient.

More details about the efficient implementation of Scatterbrain are in Appendix~\ref{sec:implementation}.

\subsection{Challenges of Designing an Accurate and Efficient Sparse + Low-rank Approximation}
\label{sec:challenges}

We seek a sparse + low-rank approximation of the attention matrix\footnote{For simplicity of discussion, we consider the unnormalized attention matrix $A = \exp(Q K^\top)$, omitting the usual scaling of $\sqrt{d}$ and the softmax normalization constant.} $A$ that is \emph{both} accurate and efficient.
The natural theoretical baseline of Robust PCA is too slow and requires too much memory, while the most straightforward way of combining sparse attention and low-rank attention fails due to double counting on the support of the sparse attention.
\begin{enumerate}[leftmargin=*,nosep,nolistsep]
  \item If the goal is accuracy, Robust PCA is the most studied algorithm to find a sparse + low-rank approximation to a given matrix.
  It relaxes the NP-hard problem of finding the best sparse + low-rank approximation into a convex optimization problem, with the nuclear norm and $\ell_1$ constraints.
  Even though it can be solved in polynomial time, it is orders of magnitude too slow to be used in each iteration of a training loop.
  Moreover, it requires materializing the attention matrix, which defeats the main purpose of reducing compute and memory requirements.
  \item On the other hand, one efficient way to get sparse + low-rank approximation of an attention matrix is to simply add the entries of a sparse approximation $S$ (say, from Reformer) and a low-rank approximation $\wt{Q}{\wt{K}}^\top$ for $\wt{Q}, \wt{K} \in \mathbb{R}^{n \times m}$ (say, from Performer).
  The sparse matrix $S$ typically has support determined randomly~\citep{child2019generating}, by LSH~\citep{kitaev2020reformer,daras2020smyrf}, or by clustering~\citep{roy2021efficient}.
  On the support of S, which likely includes the locations of the large entries of the attention matrix $A$, the entries of $S$ match those of $A$.
  One can multiply $(S + \wt{Q}{\wt{K}}^\top) V = SV + \wt{Q}({\wt{K}}^\top V)$ efficiently because $S$ is sparse, and grouping $\wt{Q}({\wt{K}}^\top V)$ reduces the matrix multiplication complexity when $m \ll n$, from $O(n^2m)$ to $O(nmd)$.
  The approximation $S + \wt{Q}{\wt{K}}^\top$ matches $\wt{Q}{\wt{K}}^\top$ outside $\supp(S)$, hence it could be accurate there if $\wt{Q}{\wt{K}}^\top$ is accurate.
  However, $S + \wt{Q}{\wt{K}}^\top$ will not be accurate on the support of $S$ due to the contributions from both $S$ and from $\wt{Q}{\wt{K}}^\top$.
  Adjusting $\wt{Q}{\wt{K}}^\top$ to discount the contribution from $S$ is difficult, especially if we want to avoid materializing $\wt{Q}{\wt{K}}^\top$ for efficiency.
\end{enumerate}

\subsection{Scatterbrain: Algorithm Intuition and Description}
\label{sec:algo_description}

The simple insight behind our method is that on the support of the sparse matrix $S$, instead of trying to match the entries of the attention matrix $A$, we can set the entries of $S$ to discount the contribution from the low-rank part $\wt{Q} {\wt{K}}^\top$.
This way, the approximation $S + \wt{Q} {\wt{K}}^\top$ will match $A$ exactly on the support of $S$, and will match $\wt{Q} {\wt{K}}^\top$ outside $\supp(S)$, which means it will still be accurate there if $\wt{Q} {\wt{K}}^\top$ is accurate.
We do not need to materialize the full matrix $\wt{Q} {\wt{K}}^\top$ as need a subset of its entries is required, hence our approximation will be compute and memory efficient.

Scatterbrain thus proceeds in three steps: we construct a low-rank approximation $\wt{Q} {\wt{K}}^\top \approx A$, and construct a sparse matrix $S$ such that $S + \wt{Q} {\wt{K}}^\top$ matches $A$ on the support of $S$, then finally multiply $SV$ and $\wt{Q} ({\wt{K}}^\top V)$ and combine the result.
More specifically:
\begin{enumerate}[leftmargin=*,nosep,nolistsep]
  \item {\bf Low-rank Approximation}.
  We define a procedure \textsc{LowRank} that returns two matrices $\wt{Q}, \wt{K} \in \mathbb{R}^{n \times m}$ such that $\wt{Q} {\wt{K}}^\top$ approximates $A$.
  In particular, we use a randomized kernel feature map $\phi \colon \mathbb{R}^d \to \mathbb{R}^m$ where $\phi(x) = \frac{1}{\sqrt{m}} \exp(Wx - \norm{x}^2/2)$ with $W \in \mathbb{R}^{m \times d}$ randomly sampled, entry-wise, from the standard normal distribution $\mathcal{N}(0, 1)$.
  We apply $\phi$ to each row vector of $Q, K$ matrices, and denote $\wt{Q} = \phi(Q)$ and $\wt{K} = \phi(K)$ (row-wise).
  Note that we do not materialize $\wt{Q} {\wt{K}}^\top$.
  \item \textbf{Sparse Approximation}.
  We define a procedure \textsc{Sparse} that returns a sparse matrix $S$ that matches $A - \wt{Q} {\wt{K}}^\top$ on $\supp(S)$.
  In particular, using a family of locality sensitive hash functions, compute the hash codes of each query and key vectors in $Q, K$ matrices (row-wise).
  Let $\mathcal{S}$ be the set of locations $(i, j)$ where $q_i$ and $k_j$ have
  the same hash codes (i.e, fall into the same hash bucket).
  Let $S$ be the sparse matrix whose support is $\mathcal{S}$, and for each
  $(i, j) \in \mathcal{S}$, define
  \begin{equation}
  \label{eq:scatterbrain_subtract}
      S_{i,j} = \exp( q_i^\top k_j ) - \phi(q_i)^\top \phi(k_j) = \exp(q_i^\top k_j) - \wt{q}_i^\top \wt{k}_j,
  \end{equation}
  where $q_i, k_j, \wt{q}_i, \wt{k}_j$ are the $i$-th and $j$-th rows of $Q, K, \wt{Q}, \wt{K}$ respectively.
  Note that we do not materialize $\wt{Q} {\wt{K}}^\top$.
  \item \textbf{Scatterbrain Approximation}. With $\wt{Q}, \wt{K}$ returned from \textsc{LowRank} and $S$ from \textsc{Sparse}, we compute the (unnormalized) attention output with
  \begin{equation}
  \label{eq:scatterbrain_attention}
      \wt{O} = (\wt{Q} {\wt{K}}^\top + S) V  = \wt{Q} ({\wt{K}}^\top V) + SV.
  \end{equation}
\end{enumerate}
The precise algorithm, including the normalization step, as well as the causal/unidirectional variant, is described in Appendix~\ref{sec:implementation}. We also note Scatterbrain's flexibility: it can use different kinds of low-rank and sparse approximation as its sub-components. The combination of Reformer and Performer is simply one instance of Scatterbrain. Instead of using Reformer as a sparse component, we could use local attention~\citep{beltagy2020longformer} or random block-sparse attention~\citep{child2019generating}. Instead of using Performer~\citep{choromanski2020rethinking} as a low-rank component, we could also use Linear attention~\citep{katharopoulos2020transformers} or global tokens as in BigBird~\citep{zaheer2020bigbird}. 

\begin{wrapfigure}{}{0.35\textwidth}
  \small
  \iftoggle{arxiv}{}{
  \vspace{-5mm}
  }
  \centering
  \includegraphics[width=0.9\linewidth]{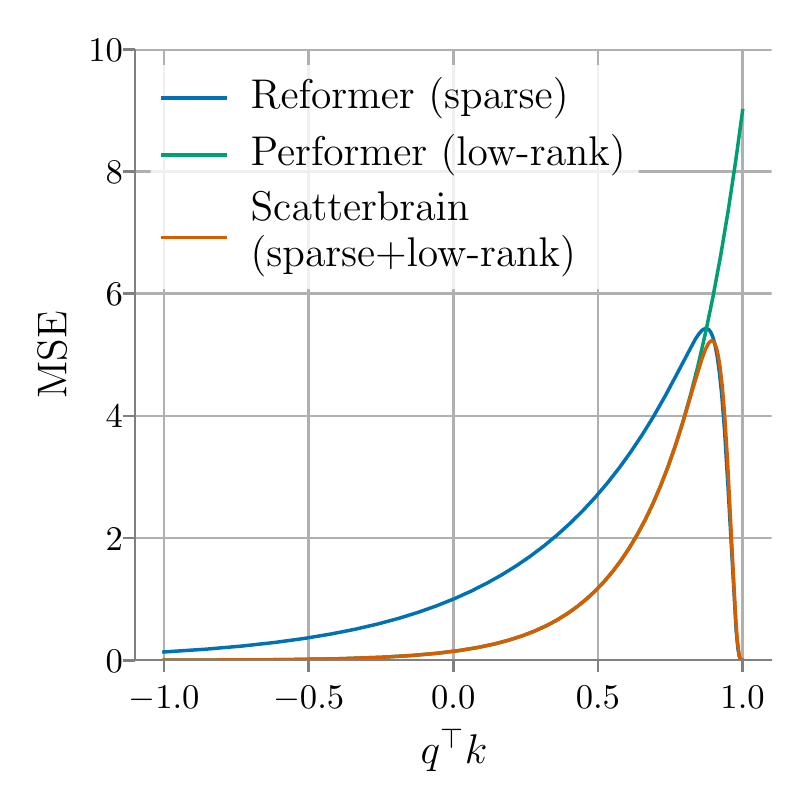}
  \vspace{-2mm}
  \caption{Per-entry MSE for different approximations, across a range of magnitude of $q^\top k$.Scatterbrain has low MSE for both small and large entries, thus outperforming its sparse (Reformer) and low-rank (Performer) counterparts.}
  \label{fig:theory_mse}
  \vspace{-2.4em}
\end{wrapfigure}

The Scatterbrain method would work exactly the same way. As long as the low-rank component is unbiased (e.g., Performer), its combination with any sparse component in Scatterbrain would yield an unbiased estimator of the attention matrix as shown below.

\subsection{Scatterbrain: Analysis}
\label{sec:algo_analysis}

Our method combines a low-rank approximation $\wt{Q} \wt{K}^\top$ (which has rank $m \ll n$) with a sparse approximation $S$. We argue that it is accurate (lower approximation error than baselines) and efficient (scaling the same as sparse or low-rank alone).
The main insight of the analysis is that our approximation is exact for entries on the support of $S$ (picked by LSH), which are likely to be large.
For entries not in the support of $S$ (likely to be small), our approximation matches the low-rank part (Performer) $\wt{Q} {\wt{K}}^\top$, which is unbiased and has low variance for these entries.
As a result, Scatterbrain retains the unbiasedness of Performer~\citep{choromanski2020rethinking} but with strictly lower variance.

We compare Scatterbrain to its low-rank baseline (Performer) and sparse baseline (Reformer).
Performer is also based on the kernel approximation $\phi$, and simply uses $\wt{Q} {\wt{K}}^\top$ to approximate the attention matrix $A$.
Reformer uses LSH to identify large entries of $A$, then compute a sparse matrix $S$ such that $S_{ij} = \exp(q_i^\top k_j)$ for $ij \in \supp(S)$.

\textbf{Accuracy}: Because of the way $S$ is defined in
  \cref{eq:scatterbrain_subtract}, $\wt{Q} {\wt{K}}^\top + S$ matches $A = \exp(QK^\top)$ exactly on locations $(i, j) \in \mathcal{S}$, which are locations with likely large values. This addresses a weakness of low-rank methods (e.g., Performer) where the low-rank estimate is not accurate for locations with large values.
  We analyze the expectation and variance per entry of our estimator below (proof in Appendix~\ref{sec:proofs}).
\begin{theorem}
  \label{thm:unbiased}
  Define $\sigma(q, k) = \exp( q^\top k)$, $\hat{\sigma}^{\mathsf{pfe}}$ as Performer's estimator and $\hat{\sigma}^{\mathsf{sbe}}$ as Scatterbrain estimator.
  Denote $ {\cal S}^{d-1} \subset \mathbb{R}^d$ as the unit sphere. Suppose $q, k \in S^{d-1}$ are such that $\Vert q-k \Vert < \tau$.
  Then:
  \begin{align}
      \mathbb{E}[\hat{\sigma}^{\mathsf{sbe}}(q, k)] = \sigma(q, k), 
      \quad \mathrm{Var}[\hat{\sigma}^{\mathsf{sbe}}(q, k)] = (1-p) \cdot \mathrm{Var}[\hat{\sigma}^{\mathsf{pfe}}(q, k)] < \mathrm{Var}[\hat{\sigma}^{\mathsf{pfe}}(q, k)]
    \end{align}
  where $p = \exp(-\frac{\tau^2}{4-\tau^2}\ln d-O_{\tau}(\ln\ln d))$.
\end{theorem}

Hence Scatterbrain is unbiased, similar to Performer~\citep{choromanski2020rethinking}, but with strictly lower variance.
The variance is small if $\exp( q^\top k )$ is small (since $\mathrm{Var}(\hat{\sigma}^{\mathsf{pfe}}(q, k))$ will be small), or if $\exp( q^\top k)$ is large (since the probability of not being selected by LSH, $1 - p$, will be small).
In \cref{fig:theory_mse}, we plot the per-entry MSE of different methods from \cref{thm:unbiased} when approximating the unnormalized softmax attention $\exp(Q K^\top)$.
Scatterbrain can approximate well both small entries (similar to the low-rank baseline, Performer), as well as large entries (similar to the sparse baseline, Reformer).
Thus Scatterbrain has much lower MSE than Performer for large entries, and lower MSE than Reformer for small entries.

\textbf{Efficiency}: In \cref{eq:scatterbrain_attention}, the computation $SV$ is
efficient because $S$ is sparse, and $\wt{Q} ({\wt{K}}^\top V)$ is efficient because of
the way we associate matrix multiplication (scaling as $O(nmd)$ instead of
$O(n^2 d)$, which is much bigger if $m \ll n$).

We validate these two properties of our approach in \cref{sec:experiments}.

\section{Experiments}
\label{sec:experiments}

We validate three claims that suggest Scatterbrain provides an accurate and efficient approximation to attention matrices, allowing it to outperform its sparse and low-rank baselines on benchmark datasets.
\begin{itemize}[leftmargin=*,nosep,nolistsep]
  \item In \cref{subsec:accurate}, we evaluate the approximation error and testing accuracy of different approximation methods on pre-trained models such as BigGAN and Vision Transformer. 
  We show that the approximation by Scatterbrain is close to the Robust PCA oracle and up to $2.1 \times$ lower approximation error than other efficient baselines.
    \item In \cref{subsec:results}, we validate that when trained end-to-end, Scatterbrain outperforms baselines (sparse or low-rank attention) on a wide variety of benchmark tasks, including language modeling, classification, and the Long-range Arena (LRA) benchmarks. Scatterbrain achieves up to 5 points higher average accuracy on the LRA benchmark compared to Performer and Reformer.
  \item In \cref{subsec:fast}, we demonstrate the scalability of Scatterbrain, showing that it has comparable memory and time usage with simpler baselines (sparse or low-rank alone) across a range of input sequence lengths (\cref{subsec:fast}), while requiring up to $12 \times$ smaller memory than full attention.
\end{itemize}
All details (hyperparameters, data splits, etc.), along with additional experiments, are in Appendix~\ref{sec:experiment_details}.

\subsection{Scatterbrain's Approximation Accuracy}
\label{subsec:accurate}

\begin{wraptable}{}{7.3cm}
\iftoggle{arxiv}{}{
  \vspace{-0.3cm}
}
\scriptsize
\caption{Top-1 Accuracy of pre-trained T2T Vision Transformer on ImageNet with different attention replacements. Error represents the average normalized approximation error to full attention.
}
\centering
\resizebox{\linewidth}{!}{
\begin{tabular}{ c||cc }
\specialrule{.15em}{.05em}{.05em}
Attention &Top-1 Acc  &  Error (avg)\\
\specialrule{.15em}{.05em}{.05em}
Full Attention & $81.7\%$ &  - \\
SMYRF & $79.8\%$ &$11.4\%$ \\
Performer & $80.1\%$ & $7.5\%$ \\
Baseline SMYRF + Performer & $79.7\%$	& $12.6\%$ \\
Scatterbrain & \bf{80.7$\%$}& \textbf{5.3$\%$} \\
\specialrule{.15em}{.05em}{.05em}
\end{tabular}
}
\iftoggle{arxiv}{}{
  \vspace{-0.3cm}
}
\label{table:vit}
\end{wraptable}

We evaluate Scatterbrain's approximation accuracy in three steps: (1) compare it with of Robust PCA (sparse+low-rank), our theoretical foundation and oracle (2) compare it with SMYRF\footnote{SMYRF is a variant of Reformer that does not require the key and query to be the same, which is necessary for experiments in this section.}~\citep{daras2020smyrf}, Performer~\citep{choromanski2020rethinking}, which are popular variants of sparse and low-rank approximation to attention respectively and a naive baseline that directly adds SMYRF and Performer, (3) evaluate the inference accuracy when replacing full attention with Scatterbrain approximation.
Scatterbrain achieves error within 20\% of the oracle robust PCA, and up to $2.1 \times$ lower error than SMYRF and Performer.
When serving as a drop-in replacement for full attention, even without training, Scatterbrain can reduce the attention memory of Vision Transformer by 98\% at the cost of only 0.8\% drop of accuracy.

\textbf{Setup:} We use the attention matrices from pre-trained BigGAN and T2T-ViT. BigGAN is a state-of-the-art model in Image Generation for ImageNet. BigGAN has a single attention layer at resolution 64 × 64 (4096 queries). T2T-ViT has 14 attention layers. Scatterbrain sets the ratio between SMYRF and Performer based on the entropy of an observed subset of attention matrices in different layers. We allocate more memory to the low-rank component compared to the sparse part if the entropy is high.

\begin{figure}
	\begin{center}
	\vspace{-0.6cm}
	\scriptsize
		\begin{tabular}{cccc}
		    \hspace{-0.5cm}
			\includegraphics[width=0.25\linewidth]{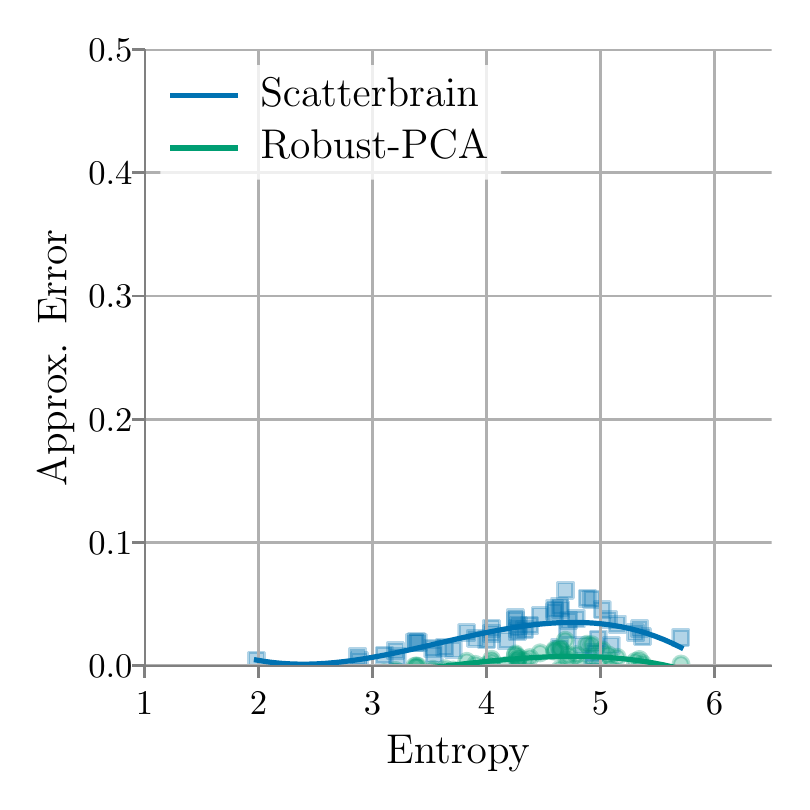}
			\includegraphics[width=0.25\linewidth]{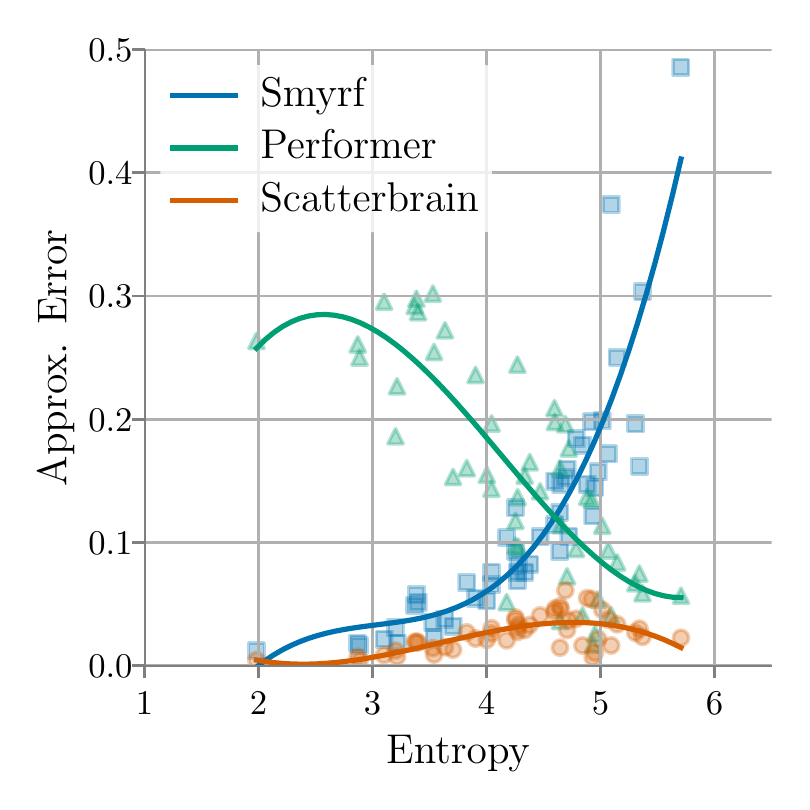}
			\includegraphics[width=0.25\linewidth]{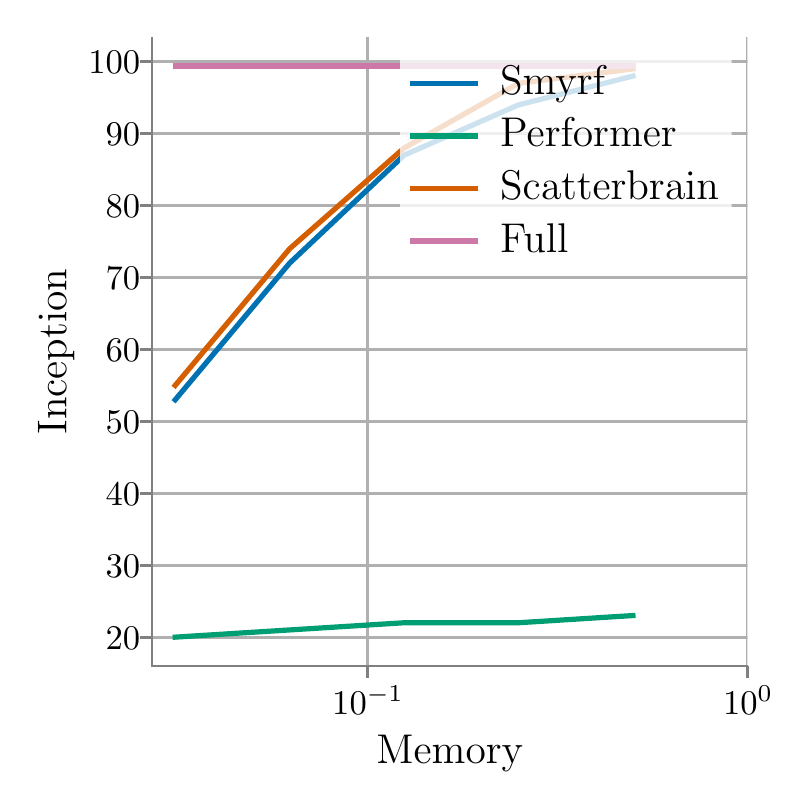}
			\includegraphics[width=0.25\linewidth]{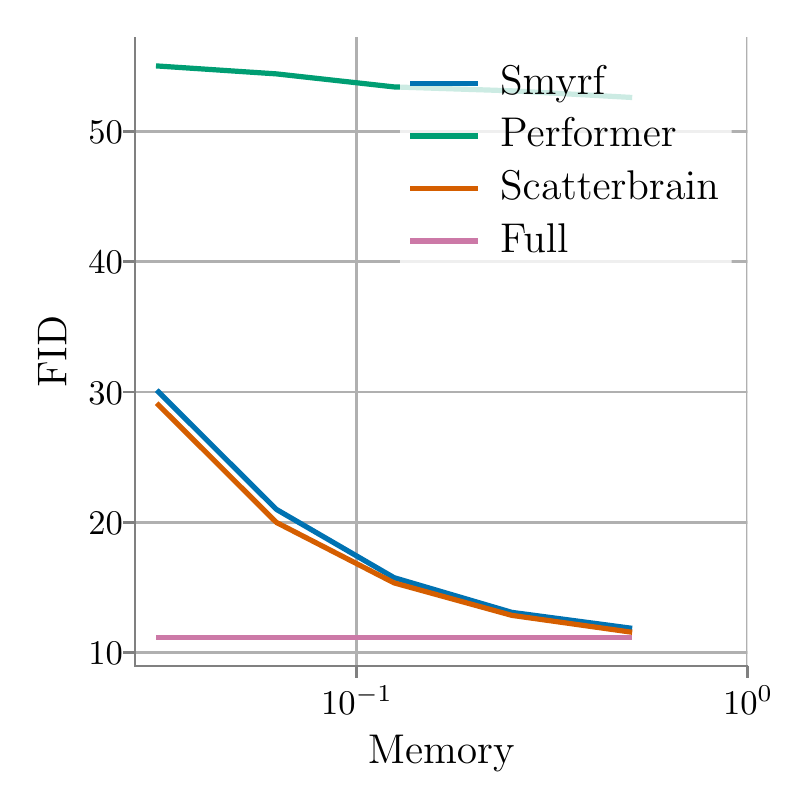}
		\end{tabular}
	\end{center}
	\caption{First: approximation comparison between Scatterbrain and its ``lowerbound" Robust PCA. 
	Second: comparison of error vs.\ entropy among SMYRF, Performer and Scatterbrain, three representatives of sparse, low-rank and sparse+low-rank approximations. 
	Third and forth: Inception score (higher is better) and FID score (lower is better) of different attention variants for pretrained BigGAN.}
	\vspace{-0.6cm}
	\label{fig:upperbound} 
\end{figure}

\textbf{Scatterbrain and Robust PCA:}
We first show that Scatterbrain approximates pre-trained attention matrices $10^5 \times$ faster while its approximation error is within 20\% on average. We also provide an example visualization on 100 attention matrices from the BigGAN generation process in Figure~\ref{fig:upperbound} (left).

\textbf{Scatterbrain vs.\ SMYRF and Performer:} We show that Scatterbrain approximates pre-trained dense attention matrices with very low error compared to sparse (Reformer) or low-rank (Performer).
Measuring Frobenius approx.\ error on the BigGAN image generation task, Scatterbrain achieves $2\times$ lower error compared to Performer.

\textbf{Drop-in replacement for full attention:}
We show that accurate approximation directly leads to efficient Inference. We replace BigGAN’s dense attention with a Scatterbrain layer without other modifications. In \ref{fig:upperbound} (right two), we show Inception and FID scores for Scatterbrain and other baselines under different memory budgets. Similarly, we use T2T-ViT~\citep{yuan2021tokens}, which is a token-to-token vision Transformer pre-trained on ImageNet~\citep{5206848}. In~\cref{table:vit}, we show the average approximation error of Scatterbrain for each layer and the end-to-end testing accuracy after replacing full attention with Scatterbrain and other baselines. Notably, Scatterbrain achieves $80.7\%$ Top-1 accuracy, which is only $1\%$ drop from the original $81.7\%$ by full attention reducing up to $98\%$ of the memory usage.

\subsection{End-to-end Training Performance}
\label{subsec:results}

Scatterbrain's accurate approximation of attention matrices allows it to outperform other efficient Transformer methods on benchmark tasks. Across a range of diverse tasks, both commonly used autoregressive tasks (sequence modeling) and benchmark long-range classification tasks (Long-Range Arena), Scatterbrain outperforms Performer (low-rank baseline) and Reformer (sparse baseline) by up to 4 points.

\subsubsection{Auto-regressive Tasks}
On the standard language modeling task of Wikitext-103, Scatterbrain obtains 1 point better perplexity than Reformer (sparse baseline), coming within 1.5 points of full attention.

\textbf{Settings:} We compare the performance of Scatterbrain against Reformer and Performer on one popular synthetic task, Copy, and one large language modeling task: WikiText103~\citep{merity2016pointer}. Reformer is a representative sparse-approximation-based variant and Performer is a low-rank-approximation-based variant. The base model is vanilla Transformer~\citep{vaswani2017attention}. We observed that generally allocating more memory budget to sparse tends to perform better, so Scatterbrain sets the ratio to 3:1 (sparse: low-rank component) for simplicity. The statistics of each dataset and model hyper-parameters are in Appendix~\ref{sec:experiment_details}. We report the best results of each method in perplexity.

\begin{table}[!tb]
    \caption{The performance of Scatterbrain, Reformer, Performer and Full-Attention on Long-Range-Arena benchmarks and 2 popular language modeling tasks. We fix the same number of parameters ($1/8$ of the full) used for approximating the attention matrix for each method.\vspace{0.3cm}}
    \begin{minipage}{.5\linewidth}
    \hspace{-1cm}
      \centering
          	\resizebox{0.9\linewidth}{!}{
	\centering
	\begingroup
	\setlength{\tabcolsep}{10pt}
	\renewcommand{\arraystretch}{1.25}
	\begin{tabular}{c||c|c}
    \specialrule{.15em}{.05em}{.05em}
    \multirow{1}{*}{ {\bf Attention} } & \multicolumn{1}{c|}{\multirow{1}{*}{Copy (ppl)}} & \multicolumn{1}{c}{\multirow{1}{*}{WikiText-103 (ppl)}} \\
	\hline
	Full Attention& 1 & 25.258  \\
	\cline{1-3}
	Reformer& 6.8 & 27.68 \\
	Performer& 49 & 66  \\
	\cline{1-3}
	Scatterbrain& \textbf{2.58} &\textbf{26.72}  \\
	\specialrule{.15em}{.05em}{.05em}
	\end{tabular}
		\endgroup
	}
    \end{minipage}%
    \begin{minipage}{.6\linewidth}
      \centering
        \hspace{-2cm}
            	\resizebox{0.9\linewidth}{!}{
	\centering
	\Huge
	\begingroup
	\setlength{\tabcolsep}{10pt}
	\renewcommand{\arraystretch}{1.25}
	\begin{tabular}{c||c|c|c|c|c|c}
    \specialrule{.15em}{.05em}{.05em}
    \multirow{1}{*}{ {\bf Attention} }  &
    \multicolumn{1}{c|}{\multirow{1}{*}{ListOps}} &
    \multicolumn{1}{c|}{\multirow{1}{*}{Text}} & 
    \multicolumn{1}{c|}{\multirow{1}{*}{Retrieval}} & 
    \multicolumn{1}{c|}{\multirow{1}{*}{Image}} &
    \multicolumn{1}{c|}{\multirow{1}{*}{Pathfinder}}  & 
    \multicolumn{1}{c}{\multirow{1}{*}{Avg}}\\
	\hline
	\hline
	Full Attention& 38.2& 63.29& 80.85 & 41.78 & 73.98 & 59.62\\
	\cline{1-7}
	\hline
	\hline
	Reformer& 36.85& 58.12& 78.36 & 28.3 & 67.95 & 53.9 \\
	Performer& 35.75& 62.36 & 78.83 & 39.71 & 68.6 & 57.05 \\
	\cline{1-7}
	Scatterbrain& \textbf{38.6} & \textbf{64.55} & \textbf{80.22} & \textbf{43.65} & \textbf{69.91} & \textbf{59.38} \\
	\specialrule{.15em}{.05em}{.05em}
	\end{tabular}
		\endgroup
	}
    \end{minipage} 
    \label{table:main}
    \vspace{-0.2cm}
\end{table}

\textbf{Results:} Table~\ref{table:main} shows the testing perplexity for Scatterbrain and other baselines under the same parameter budget (each approximation is only allowed to compute $\frac{1}{8}$ of the full computation). Scatterbrain achieves comparable perplexity compared to the full attention Transformer model on Copy, and WikiText-103. Notably, Scatterbrain achieves 4 points lower perplexity on Copy and 1 point lower on WikiText-103 compared to Reformer, 
while Performer does not train stably on auto-regressive tasks (loss does not go down).

\textbf{Analysis:} We also analyze the results by visualizing the error of Reformer (sparse), Performer (low-rank), and Scatterbrain (sparse + low-rank) given the same number of parameters when approximating the full attention matrices for each attention layer during training (Appendix~\ref{sec:experiment_details}). The conclusion is for language modeling tasks, sparse+low-rank has the smallest approximation error in most of the cases, and sparse has the largest error, which matches with the end-to-end results. It also confirms the observation in the popular benchmark paper~\citep{tay2020long} that kernel or low-rank based approximations are less effective for hierarchical structured data.

\subsubsection{Classification Tasks}

On a suite of long-range benchmark tasks (Long Range Area), Scatterbrain outperforms Reformer (sparse baseline) and Performer (low-rank baseline) by up to 5 points on average.

\textbf{Settings:} We compare the performance of Scatterbrain against Reformer and Performer on ListOps, two classifications: byte-level IMDb reviews text classification, image classification on sequences of pixels, a text retrieval, and pathfinder tasks. The datasets are obtained from the Long Range Arena (LRA) Benchmark~\citep{tay2020long}, which is a recent popular benchmark designed for testing efficient Transformers. Similar to the auto-regressive tasks above, we use Reformer and Performer as baselines. The base model is also a vanilla Transformer. We follow the evaluation protocol from~\citep{tay2020long}. We report the best accuracy of each method.

\textbf{Results:} Table~\ref{table:main} shows the individual and average accuracy of each task for Scatterbrain and other baselines under the same parameters budget. Specially, each approximation is only allowed to use $12.5\%$ of the full computation. We can see Scatterbrain is very close to full attention even with a large reduction in computation and memory. Further more, it outperforms all the other baselines consistently on every task and achieves more than 5 point average accuracy improvement than sparse-based approximation Reformer and more than 2 point average accuracy improvement than low-rank-based variant Performer.

\begin{wrapfigure}{}{0.66\textwidth}
  \small
  \iftoggle{arxiv}{}{
    \vspace{-0.7cm}
  }
  \centering
      \includegraphics[width=0.45\linewidth]{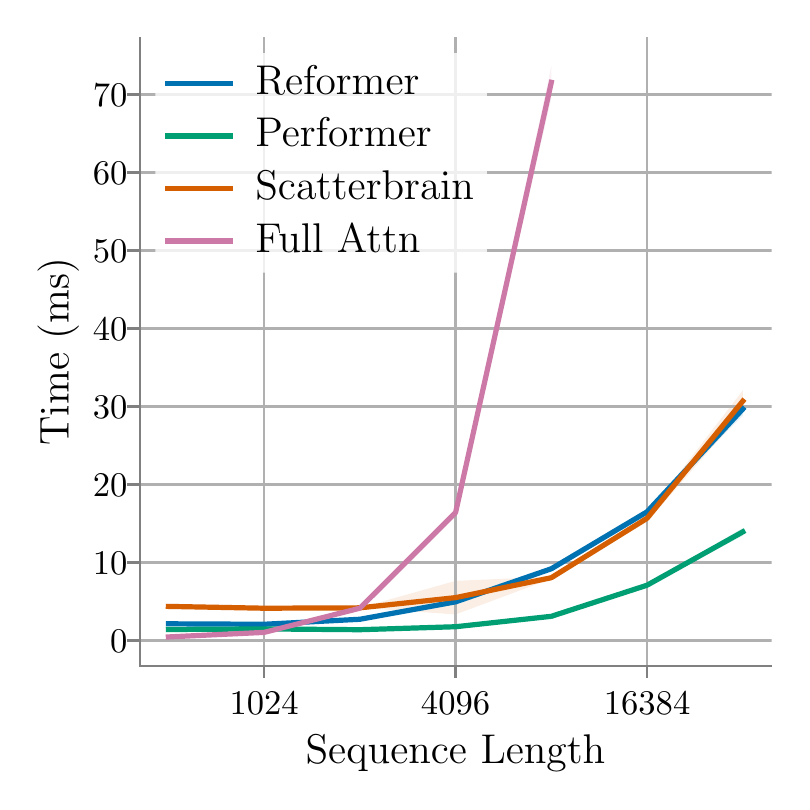}
      \includegraphics[width=0.45\linewidth]{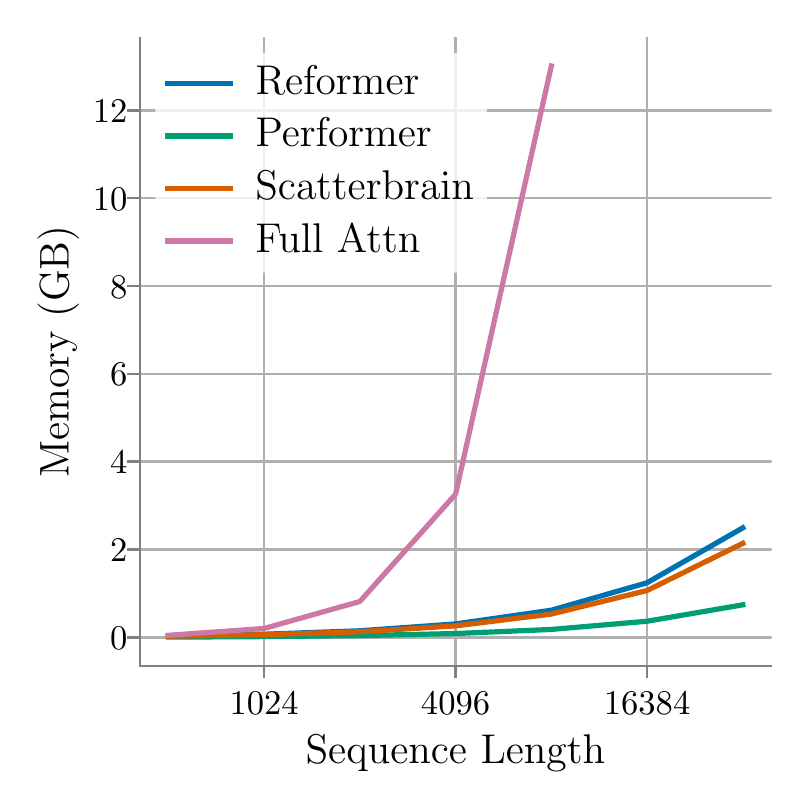}
  \caption{Speed and memory required by different efficient attention methods.
    Scatterbrain is competitive with SMYRF (sparse baseline) and Performer (low-rank baseline), while up to $3 \times$ faster and $12 \times$ more memory efficient than full attention for sequence length 4096.}
  \label{fig:efficiency}
  \vspace{-0.6cm}
\end{wrapfigure}
\textbf{Analysis:} Similarly, in order to analyze the performance of Reformer, Performer and Scatterbrain, we visualize their approximation error given the same number of parameters when approximating the full attention matrices for each attention layer during training (Appendix~\ref{sec:experiment_details}). We again find that Scatterbrain has the smallest approximation error, while Performer is the worst on ListOps and Reformer has the largest error on classification tasks, which matches with the end-to-end results and confirms our observations earlier (sparse and low-rank approximation excel in different regimes).

\subsection{Scatterbrain's Efficiency, Scaling with Input Sequence Length}
\label{subsec:fast}

We include ablation studies on the scalability of Scatterbrain in \cref{fig:efficiency}, showing that it is as computation and memory-efficient as simpler baselines such as SMYRF and Performer, while up to $3 \times$ faster and $12 \times$ more memory efficient than full attention for sequence length 4096.
This demonstrates that our combination of sparse and low-rank inherits their efficiency.

We report run times and memory consumption of the sequence lengths ranging from 512 to 32768. We use a batch size of 16 for all runs and conduct experiments a V100 GPU. Since the efficiency would be largely conditioned on hardware and implementation details, we perform best-effort fair comparisons. We adapt the Pytorch implementation from \texttt{pytorch-fast-transformers} library for our baselines and implement Scatterbrain similarly without any customized cuda kernels.

\section{Discussion}
\label{sec:conclusion}

\textbf{Limitations.}
As Scatterbrain has sparse attention as a component, it is not yet as hardware friendly (on GPUs and TPUs) as the low-rank component, which uses the very optimized dense matrix multiplication.
This is the same limitation suffered by other sparse attention methods, but we are excited that more efficient sparse GPU kernels are being developed~\citep{gray2017gpu,gale2020sparse}.

\textbf{Potential negative societal impacts.}
Our work seeks to understand the role of matrix approximation (and potentially energy savings) in the attention layer, which may improve a wide range of applications, each with their own potential benefits and harms.
For example, making it language modeling more compute and memory efficient might facilitate spreading misinformation, and better image and video processing may
make automatic surveillance easier.
To mitigate these risks, one needs to address application-specific issues such as privacy and fairness, going beyond the error metrics we considered.
Specially, for language models (LMs), while our work partially addresses the issue of environmental cost of LMs raised in~\citep{bender2021on}, it does not address other issues such as unfathomable training data~\citep{bender2021on}.

\textbf{Discussion and future work.}
In this work, we make an observation on the sparse + low-rank structure of the attentions in Transformer models and theoretically characterize the regimes where sparse, low-rank and sparse + low-rank excel, based on the softmax temperature of the attention matrices.
Motivated by this observation, we present Scatterbrain, a novel way to unify the strengths of both sparse and low-rank methods for accurate and efficient attention approximation with provable guarantees.
We empirically verify the effectiveness of Scatterbrain on pretrained BigGAN, vision transformers, as well as end-to-end training of vanilla transformer.
We anticipate that the study of this core approximation problem can prove useful in other contexts, such as generalized attention layers with other non-linearity beside softmax, and wide output layer in language modeling or extreme-classification.

\subsubsection*{Acknowledgments}

We thank Xun Huang, Sarah Hooper, Albert Gu, Ananya Kumar, Sen Wu, Trenton Chang, Megan Leszczynski, and Karan Goel for their helpful discussions and feedback on early drafts of the paper.

We gratefully acknowledge the support of NIH under No.\ U54EB020405 (Mobilize), NSF under Nos.\ CCF1763315 (Beyond Sparsity), CCF1563078 (Volume to Velocity), and 1937301 (RTML); ONR under No.\ N000141712266 (Unifying Weak Supervision); ONR N00014-20-1-2480: Understanding and Applying Non-Euclidean Geometry in Machine Learning; N000142012275 (NEPTUNE); the Moore Foundation, NXP, Xilinx, LETI-CEA, Intel, IBM, Microsoft, NEC, Toshiba, TSMC, ARM, Hitachi, BASF, Accenture, Ericsson, Qualcomm, Analog Devices, the Okawa Foundation, American Family Insurance, Google Cloud, Salesforce, Total, the HAI-AWS Cloud Credits for Research program, the Stanford Data Science Initiative (SDSI), and members of the Stanford DAWN project: Facebook, Google, and VMWare. The Mobilize Center is a Biomedical Technology Resource Center, funded by the NIH National Institute of Biomedical Imaging and Bioengineering through Grant P41EB027060. The U.S.\ Government is authorized to reproduce and distribute reprints for Governmental purposes notwithstanding any copyright notation thereon. Any opinions, findings, and conclusions or recommendations expressed in this material are those of the authors and do not necessarily reflect the views, policies, or endorsements, either expressed or implied, of NIH, ONR, or the U.S.\ Government. 
Atri Rudra’s research is supported by NSF grant CCF-1763481.

\bibliography{ref}

\begin{thebibliography}{71}
\providecommand{\natexlab}[1]{#1}
\providecommand{\url}[1]{\texttt{#1}}
\expandafter\ifx\csname urlstyle\endcsname\relax
  \providecommand{\doi}[1]{doi: #1}\else
  \providecommand{\doi}{doi: \begingroup \urlstyle{rm}\Url}\fi

\bibitem[Alizadeh et~al.(2020)Alizadeh, Farhadi, and
  Rastegari]{alizadeh2019butterfly}
Keivan Alizadeh, Ali Farhadi, and Mohammad Rastegari.
\newblock Butterfly transform: An efficient {FFT} based neural architecture
  design.
\newblock In \emph{The Conference on Computer Vision and Pattern Recognition
  (CVPR)}, 2020.

\bibitem[Andoni et~al.(2015{\natexlab{a}})Andoni, Indyk, Laarhoven,
  Razenshteyn, and Schmidt]{NIPS2015_5893}
Alexandr Andoni, Piotr Indyk, Thijs Laarhoven, Ilya Razenshteyn, and Ludwig
  Schmidt.
\newblock Practical and optimal lsh for angular distance.
\newblock In C.~Cortes, N.~D. Lawrence, D.~D. Lee, M.~Sugiyama, and R.~Garnett,
  editors, \emph{Advances in Neural Information Processing Systems (NeurIPS)},
  pages 1225--1233. 2015{\natexlab{a}}.

\bibitem[Andoni et~al.(2015{\natexlab{b}})Andoni, Indyk, Laarhoven,
  Razenshteyn, and Schmidt]{andoni2015practical}
Alexandr Andoni, Piotr Indyk, Thijs Laarhoven, Ilya Razenshteyn, and Ludwig
  Schmidt.
\newblock Practical and optimal {LSH} for angular distance.
\newblock In \emph{Proceedings of the 28th International Conference on Neural
  Information Processing Systems-Volume 1}, pages 1225--1233,
  2015{\natexlab{b}}.

\bibitem[Artusi et~al.(2002)Artusi, Verderio, and Marubini]{artusi2002bravais}
R~Artusi, P~Verderio, and E~Marubini.
\newblock Bravais-pearson and spearman correlation coefficients: meaning, test
  of hypothesis and confidence interval.
\newblock \emph{The International journal of biological markers}, 17\penalty0
  (2):\penalty0 148--151, 2002.

\bibitem[Beltagy et~al.(2020)Beltagy, Peters, and Cohan]{beltagy2020longformer}
Iz~Beltagy, Matthew~E Peters, and Arman Cohan.
\newblock Longformer: The long-document transformer.
\newblock \emph{arXiv preprint arXiv:2004.05150}, 2020.

\bibitem[Bender et~al.(2021)Bender, Gebru, McMillan-Major, and
  Mitchell]{bender2021on}
Emily~M. Bender, Timnit Gebru, Angelina McMillan-Major, and Margaret Mitchell.
\newblock On the dangers of stochastic parrots: Can language models be too big?
\newblock In \emph{Proceedings of the 2021 ACM Conference on Fairness,
  Accountability, and Transparency}, New York, NY, USA, 2021. Association for
  Computing Machinery.

\bibitem[Brown et~al.(2020)Brown, Mann, Ryder, Subbiah, Kaplan, Dhariwal,
  Neelakantan, Shyam, Sastry, Askell, et~al.]{brown2020language}
Tom~B Brown, Benjamin Mann, Nick Ryder, Melanie Subbiah, Jared Kaplan, Prafulla
  Dhariwal, Arvind Neelakantan, Pranav Shyam, Girish Sastry, Amanda Askell,
  et~al.
\newblock Language models are few-shot learners.
\newblock \emph{arXiv preprint arXiv:2005.14165}, 2020.

\bibitem[Cand{\`e}s and Recht(2009)]{candes2009exact}
Emmanuel~J Cand{\`e}s and Benjamin Recht.
\newblock Exact matrix completion via convex optimization.
\newblock \emph{Foundations of Computational mathematics}, 9\penalty0
  (6):\penalty0 717--772, 2009.

\bibitem[Cand{\`e}s et~al.(2011)Cand{\`e}s, Li, Ma, and
  Wright]{candes2011robust}
Emmanuel~J Cand{\`e}s, Xiaodong Li, Yi~Ma, and John Wright.
\newblock Robust principal component analysis?
\newblock \emph{Journal of the ACM (JACM)}, 58\penalty0 (3):\penalty0 1--37,
  2011.

\bibitem[Carion et~al.(2020)Carion, Massa, Synnaeve, Usunier, Kirillov, and
  Zagoruyko]{carion2020end}
Nicolas Carion, Francisco Massa, Gabriel Synnaeve, Nicolas Usunier, Alexander
  Kirillov, and Sergey Zagoruyko.
\newblock End-to-end object detection with transformers.
\newblock In \emph{European Conference on Computer Vision}, pages 213--229.
  Springer, 2020.

\bibitem[Chen and Shrivastava(2018)]{chen2018densified}
Beidi Chen and Anshumali Shrivastava.
\newblock Densified winner take all (wta) hashing for sparse datasets.
\newblock In \emph{Uncertainty in artificial intelligence}, 2018.

\bibitem[Chen et~al.(2018)Chen, Shrivastava, and Steorts]{chen2018unique}
Beidi Chen, Anshumali Shrivastava, and Rebecca~C Steorts.
\newblock Unique entity estimation with application to the syrian conflict.
\newblock \emph{The Annals of Applied Statistics}, 12\penalty0 (2):\penalty0
  1039--1067, 2018.

\bibitem[Chen et~al.(2019)Chen, Xu, and Shrivastava]{chen2019fast}
Beidi Chen, Yingchen Xu, and Anshumali Shrivastava.
\newblock Fast and accurate stochastic gradient estimation.
\newblock 2019.

\bibitem[Chen et~al.(2020)Chen, Medini, Farwell, Tai, Shrivastava,
  et~al.]{chen2019slide}
Beidi Chen, Tharun Medini, James Farwell, Charlie Tai, Anshumali Shrivastava,
  et~al.
\newblock {SLIDE}: In defense of smart algorithms over hardware acceleration
  for large-scale deep learning systems.
\newblock \emph{Proceedings of Machine Learning and Systems}, 2:\penalty0
  291--306, 2020.

\bibitem[Chen et~al.(2021)Chen, Liu, Peng, Xu, Li, Dao, Song, Shrivastava, and
  R{\'e}]{chen2020mongoose}
Beidi Chen, Zichang Liu, Binghui Peng, Zhaozhuo Xu, Jonathan~Lingjie Li, Tri
  Dao, Zhao Song, Anshumali Shrivastava, and Christopher R{\'e}.
\newblock Mongoose: A learnable lsh framework for efficient neural network
  training.
\newblock In \emph{The International Conference on Learning Representations
  ({ICLR})}, 2021.

\bibitem[Child et~al.(2019)Child, Gray, Radford, and
  Sutskever]{child2019generating}
Rewon Child, Scott Gray, Alec Radford, and Ilya Sutskever.
\newblock Generating long sequences with sparse transformers.
\newblock \emph{arXiv preprint arXiv:1904.10509}, 2019.

\bibitem[Choromanski et~al.(2020)Choromanski, Likhosherstov, Dohan, Song, Gane,
  Sarlos, Hawkins, Davis, Mohiuddin, Kaiser, et~al.]{choromanski2020rethinking}
Krzysztof Choromanski, Valerii Likhosherstov, David Dohan, Xingyou Song,
  Andreea Gane, Tamas Sarlos, Peter Hawkins, Jared Davis, Afroz Mohiuddin,
  Lukasz Kaiser, et~al.
\newblock Rethinking attention with performers.
\newblock \emph{arXiv preprint arXiv:2009.14794}, 2020.

\bibitem[Daghaghi et~al.(2021)Daghaghi, Medini, Meisburger, Chen, Zhao, and
  Shrivastava]{daghaghi2021tale}
Shabnam Daghaghi, Tharun Medini, Nicholas Meisburger, Beidi Chen, Mengnan Zhao,
  and Anshumali Shrivastava.
\newblock A tale of two efficient and informative negative sampling
  distributions.
\newblock In \emph{International Conference on Machine Learning}, pages
  2319--2329. PMLR, 2021.

\bibitem[Dai et~al.(2019)Dai, Yang, Yang, Carbonell, Le, and
  Salakhutdinov]{dai2019transformer}
Zihang Dai, Zhilin Yang, Yiming Yang, Jaime Carbonell, Quoc~V Le, and Ruslan
  Salakhutdinov.
\newblock Transformer-xl: Attentive language models beyond a fixed-length
  context.
\newblock \emph{arXiv preprint arXiv:1901.02860}, 2019.

\bibitem[Dao et~al.(2019)Dao, Gu, Eichhorn, Rudra, and R{\'e}]{dao2019learning}
Tri Dao, Albert Gu, Matthew Eichhorn, Atri Rudra, and Christopher R{\'e}.
\newblock Learning fast algorithms for linear transforms using butterfly
  factorizations.
\newblock In \emph{The International Conference on Machine Learning ({ICML})},
  2019.

\bibitem[Dao et~al.(2020)Dao, Sohoni, Gu, Eichhorn, Blonder, Leszczynski,
  Rudra, and Ré]{dao2020kaleidoscope}
Tri Dao, Nimit Sohoni, Albert Gu, Matthew Eichhorn, Amit Blonder, Megan
  Leszczynski, Atri Rudra, and Christopher Ré.
\newblock Kaleidoscope: An efficient, learnable representation for all
  structured linear maps.
\newblock In \emph{The International Conference on Learning Representations
  ({ICLR})}, 2020.

\bibitem[Daras et~al.(2020)Daras, Kitaev, Odena, and Dimakis]{daras2020smyrf}
Giannis Daras, Nikita Kitaev, Augustus Odena, and Alexandros~G Dimakis.
\newblock Smyrf: Efficient attention using asymmetric clustering.
\newblock \emph{arXiv preprint arXiv:2010.05315}, 2020.

\bibitem[De~Sa et~al.(2015)De~Sa, Re, and Olukotun]{de2015global}
Christopher De~Sa, Christopher Re, and Kunle Olukotun.
\newblock Global convergence of stochastic gradient descent for some non-convex
  matrix problems.
\newblock In \emph{International Conference on Machine Learning}, pages
  2332--2341. PMLR, 2015.

\bibitem[De~Sa et~al.(2018)De~Sa, Gu, Puttagunta, R{\'e}, and Rudra]{de2018two}
Christopher De~Sa, Albert Gu, Rohan Puttagunta, Christopher R{\'e}, and Atri
  Rudra.
\newblock A two-pronged progress in structured dense matrix vector
  multiplication.
\newblock In \emph{Proceedings of the Twenty-Ninth Annual ACM-SIAM Symposium on
  Discrete Algorithms}, pages 1060--1079. SIAM, 2018.

\bibitem[Deng et~al.(2009)Deng, Dong, Socher, Li, Li, and Fei-Fei]{5206848}
Jia Deng, Wei Dong, Richard Socher, Li-Jia Li, Kai Li, and Li~Fei-Fei.
\newblock Imagenet: A large-scale hierarchical image database.
\newblock In \emph{2009 IEEE Conference on Computer Vision and Pattern
  Recognition}, pages 248--255, 2009.
\newblock \doi{10.1109/CVPR.2009.5206848}.

\bibitem[Devlin et~al.(2018)Devlin, Chang, Lee, and Toutanova]{devlin2018bert}
Jacob Devlin, Ming-Wei Chang, Kenton Lee, and Kristina Toutanova.
\newblock Bert: Pre-training of deep bidirectional transformers for language
  understanding.
\newblock \emph{arXiv preprint arXiv:1810.04805}, 2018.

\bibitem[Dong et~al.(2019)Dong, Indyk, Razenshteyn, and
  Wagner]{dong2019learning}
Yihe Dong, Piotr Indyk, Ilya Razenshteyn, and Tal Wagner.
\newblock Learning space partitions for nearest neighbor search.
\newblock In \emph{International Conference on Learning Representations
  (ICLR)}, 2019.

\bibitem[Dosovitskiy et~al.(2020)Dosovitskiy, Beyer, Kolesnikov, Weissenborn,
  Zhai, Unterthiner, Dehghani, Minderer, Heigold, Gelly,
  et~al.]{dosovitskiy2020image}
Alexey Dosovitskiy, Lucas Beyer, Alexander Kolesnikov, Dirk Weissenborn,
  Xiaohua Zhai, Thomas Unterthiner, Mostafa Dehghani, Matthias Minderer, Georg
  Heigold, Sylvain Gelly, et~al.
\newblock An image is worth 16x16 words: Transformers for image recognition at
  scale.
\newblock \emph{arXiv preprint arXiv:2010.11929}, 2020.

\bibitem[Gale et~al.(2020)Gale, Zaharia, Young, and Elsen]{gale2020sparse}
Trevor Gale, Matei Zaharia, Cliff Young, and Erich Elsen.
\newblock Sparse {GPU} kernels for deep learning.
\newblock In \emph{Supercomputing}, 2020.

\bibitem[Gionis et~al.(1999)Gionis, Indyk, Motwani,
  et~al.]{gionis1999similarity}
Aristides Gionis, Piotr Indyk, Rajeev Motwani, et~al.
\newblock Similarity search in high dimensions via hashing.
\newblock In \emph{Vldb}, volume~99, pages 518--529, 1999.

\bibitem[Gray et~al.(2017)Gray, Radford, and Kingma]{gray2017gpu}
Scott Gray, Alec Radford, and Diederik~P Kingma.
\newblock {GPU} kernels for block-sparse weights.
\newblock \emph{arXiv preprint arXiv:1711.09224}, 3, 2017.

\bibitem[Gu et~al.(2020)Gu, Dao, Ermon, Rudra, and R{\'e}]{gu2020hippo}
Albert Gu, Tri Dao, Stefano Ermon, Atri Rudra, and Christopher R{\'e}.
\newblock Hippo: Recurrent memory with optimal polynomial projections.
\newblock In \emph{Advances in neural information processing systems
  (NeurIPS)}, 2020.

\bibitem[Hotelling(1933)]{hotelling1933analysis}
Harold Hotelling.
\newblock Analysis of a complex of statistical variables into principal
  components.
\newblock \emph{Journal of educational psychology}, 24\penalty0 (6):\penalty0
  417, 1933.

\bibitem[Indyk and Motwani(1998)]{indyk1998approximate}
Piotr Indyk and Rajeev Motwani.
\newblock Approximate nearest neighbors: towards removing the curse of
  dimensionality.
\newblock In \emph{Proceedings of the thirtieth annual ACM symposium on Theory
  of computing}, pages 604--613, 1998.

\bibitem[Katharopoulos et~al.(2020)Katharopoulos, Vyas, Pappas, and
  Fleuret]{katharopoulos2020transformers}
Angelos Katharopoulos, Apoorv Vyas, Nikolaos Pappas, and Fran{\c{c}}ois
  Fleuret.
\newblock Transformers are rnns: Fast autoregressive transformers with linear
  attention.
\newblock In \emph{International Conference on Machine Learning}, pages
  5156--5165. PMLR, 2020.

\bibitem[Kitaev et~al.(2020)Kitaev, Kaiser, and Levskaya]{kitaev2020reformer}
Nikita Kitaev, {\L}ukasz Kaiser, and Anselm Levskaya.
\newblock Reformer: The efficient transformer.
\newblock In \emph{The International Conference on Machine Learning ({ICML})},
  2020.

\bibitem[Krizhevsky et~al.(2009)Krizhevsky, Hinton,
  et~al.]{krizhevsky2009learning}
Alex Krizhevsky, Geoffrey Hinton, et~al.
\newblock Learning multiple layers of features from tiny images.
\newblock 2009.

\bibitem[Lan et~al.(2020)Lan, Chen, Goodman, Gimpel, Sharma, and
  Soricut]{lan2019albert}
Zhenzhong Lan, Mingda Chen, Sebastian Goodman, Kevin Gimpel, Piyush Sharma, and
  Radu Soricut.
\newblock Albert: A lite bert for self-supervised learning of language
  representations.
\newblock In \emph{The International Conference on Learning Representations
  ({ICLR})}, 2020.

\bibitem[Likhosherstov et~al.(2020)Likhosherstov, Choromanski, Davis, Song, and
  Weller]{likhosherstov2020sub}
Valerii Likhosherstov, Krzysztof Choromanski, Jared Davis, Xingyou Song, and
  Adrian Weller.
\newblock Sub-linear memory: How to make performers slim.
\newblock \emph{arXiv preprint arXiv:2012.11346}, 2020.

\bibitem[Linsley et~al.(2018)Linsley, Kim, Veerabadran, and
  Serre]{linsley2018learning}
Drew Linsley, Junkyung Kim, Vijay Veerabadran, and Thomas Serre.
\newblock Learning long-range spatial dependencies with horizontal
  gated-recurrent units.
\newblock \emph{arXiv preprint arXiv:1805.08315}, 2018.

\bibitem[Liu et~al.(2020)Liu, Xu, Ji, Li, Chen, and
  Shrivastava]{liu2020climbing}
Zichang Liu, Zhaozhuo Xu, Alan Ji, Jonathan Li, Beidi Chen, and Anshumali
  Shrivastava.
\newblock Climbing the wol: Training for cheaper inference.
\newblock \emph{arXiv preprint arXiv:2007.01230}, 2020.

\bibitem[Luo et~al.(2021)Luo, Zhang, Lei, and Xie]{luo2021simplified}
Haoneng Luo, Shiliang Zhang, Ming Lei, and Lei Xie.
\newblock Simplified self-attention for transformer-based end-to-end speech
  recognition.
\newblock In \emph{2021 IEEE Spoken Language Technology Workshop (SLT)}, pages
  75--81. IEEE, 2021.

\bibitem[Ma et~al.(2021)Ma, Kong, Wang, Zhou, May, Ma, and
  Zettlemoyer]{ma2021luna}
Xuezhe Ma, Xiang Kong, Sinong Wang, Chunting Zhou, Jonathan May, Hao Ma, and
  Luke Zettlemoyer.
\newblock Luna: Linear unified nested attention.
\newblock \emph{arXiv preprint arXiv:2106.01540}, 2021.

\bibitem[Maas et~al.(2011)Maas, Daly, Pham, Huang, Ng, and
  Potts]{maas2011learning}
Andrew Maas, Raymond~E Daly, Peter~T Pham, Dan Huang, Andrew~Y Ng, and
  Christopher Potts.
\newblock Learning word vectors for sentiment analysis.
\newblock In \emph{Proceedings of the 49th annual meeting of the association
  for computational linguistics: Human language technologies}, pages 142--150,
  2011.

\bibitem[Merity et~al.(2016)Merity, Xiong, Bradbury, and
  Socher]{merity2016pointer}
Stephen Merity, Caiming Xiong, James Bradbury, and Richard Socher.
\newblock Pointer sentinel mixture models.
\newblock \emph{arXiv preprint arXiv:1609.07843}, 2016.

\bibitem[Nangia and Bowman(2018)]{nangia2018listops}
Nikita Nangia and Samuel~R Bowman.
\newblock Listops: A diagnostic dataset for latent tree learning.
\newblock \emph{arXiv preprint arXiv:1804.06028}, 2018.

\bibitem[Parmar et~al.(2018)Parmar, Vaswani, Uszkoreit, Kaiser, Shazeer, Ku,
  and Tran]{parmar2018image}
Niki Parmar, Ashish Vaswani, Jakob Uszkoreit, Lukasz Kaiser, Noam Shazeer,
  Alexander Ku, and Dustin Tran.
\newblock Image transformer.
\newblock In \emph{International Conference on Machine Learning}, pages
  4055--4064. PMLR, 2018.

\bibitem[Radev et~al.(2013)Radev, Muthukrishnan, Qazvinian, and
  Abu-Jbara]{radev2013acl}
Dragomir~R Radev, Pradeep Muthukrishnan, Vahed Qazvinian, and Amjad Abu-Jbara.
\newblock The acl anthology network corpus.
\newblock \emph{Language Resources and Evaluation}, 47\penalty0 (4):\penalty0
  919--944, 2013.

\bibitem[Rae et~al.(2020)Rae, Potapenko, Jayakumar, and
  Lillicrap]{rae2019compressive}
Jack~W Rae, Anna Potapenko, Siddhant~M Jayakumar, and Timothy~P Lillicrap.
\newblock Compressive transformers for long-range sequence modelling.
\newblock In \emph{The International Conference on Learning Representations
  ({ICLR})}, 2020.

\bibitem[Raffel et~al.(2019)Raffel, Shazeer, Roberts, Lee, Narang, Matena,
  Zhou, Li, and Liu]{raffel2019exploring}
Colin Raffel, Noam Shazeer, Adam Roberts, Katherine Lee, Sharan Narang, Michael
  Matena, Yanqi Zhou, Wei Li, and Peter~J Liu.
\newblock Exploring the limits of transfer learning with a unified text-to-text
  transformer.
\newblock \emph{arXiv preprint arXiv:1910.10683}, 2019.

\bibitem[Ramsauer et~al.(2020)Ramsauer, Sch{\"a}fl, Lehner, Seidl, Widrich,
  Adler, Gruber, Holzleitner, Pavlovi{\'c}, Sandve,
  et~al.]{ramsauer2020hopfield}
Hubert Ramsauer, Bernhard Sch{\"a}fl, Johannes Lehner, Philipp Seidl, Michael
  Widrich, Thomas Adler, Lukas Gruber, Markus Holzleitner, Milena Pavlovi{\'c},
  Geir~Kjetil Sandve, et~al.
\newblock Hopfield networks is all you need.
\newblock \emph{arXiv preprint arXiv:2008.02217}, 2020.

\bibitem[Recht(2011)]{recht2011simpler}
Benjamin Recht.
\newblock A simpler approach to matrix completion.
\newblock \emph{Journal of Machine Learning Research}, 12\penalty0 (12), 2011.

\bibitem[Roy et~al.(2021)Roy, Saffar, Vaswani, and Grangier]{roy2021efficient}
Aurko Roy, Mohammad Saffar, Ashish Vaswani, and David Grangier.
\newblock Efficient content-based sparse attention with routing transformers.
\newblock \emph{Transactions of the Association for Computational Linguistics},
  9:\penalty0 53--68, 2021.

\bibitem[Shrivastava and Li(2014)]{shrivastava2014asymmetric}
Anshumali Shrivastava and Ping Li.
\newblock Asymmetric lsh (alsh) for sublinear time maximum inner product search
  (mips).
\newblock In \emph{Advances in Neural Information Processing Systems
  (NeurIPS)}, pages 2321--2329, 2014.

\bibitem[Sindhwani et~al.(2015)Sindhwani, Sainath, and
  Kumar]{sindhwani2015structured}
Vikas Sindhwani, Tara~N. Sainath, and Sanjiv Kumar.
\newblock Structured transforms for small-footprint deep learning.
\newblock In \emph{Advances in Neural Information Processing Systems}, pages
  3088--3096, 2015.

\bibitem[Sukhbaatar et~al.(2019)Sukhbaatar, Grave, Bojanowski, and
  Joulin]{sukhbaatar2019adaptive}
Sainbayar Sukhbaatar, Edouard Grave, Piotr Bojanowski, and Armand Joulin.
\newblock Adaptive attention span in transformers.
\newblock In \emph{Proceedings of the Annual Meeting of the Association for
  Computational Linguistics}, 2019.

\bibitem[Tay et~al.(2020{\natexlab{a}})Tay, Dehghani, Abnar, Shen, Bahri, Pham,
  Rao, Yang, Ruder, and Metzler]{tay2020long}
Yi~Tay, Mostafa Dehghani, Samira Abnar, Yikang Shen, Dara Bahri, Philip Pham,
  Jinfeng Rao, Liu Yang, Sebastian Ruder, and Donald Metzler.
\newblock Long range arena: A benchmark for efficient transformers.
\newblock \emph{arXiv preprint arXiv:2011.04006}, 2020{\natexlab{a}}.

\bibitem[Tay et~al.(2020{\natexlab{b}})Tay, Dehghani, Bahri, and
  Metzler]{tay2020efficient}
Yi~Tay, Mostafa Dehghani, Dara Bahri, and Donald Metzler.
\newblock Efficient transformers: A survey.
\newblock \emph{arXiv preprint arXiv:2009.06732}, 2020{\natexlab{b}}.

\bibitem[Taylor(1990)]{taylor1990interpretation}
Richard Taylor.
\newblock Interpretation of the correlation coefficient: a basic review.
\newblock \emph{Journal of diagnostic medical sonography}, 6\penalty0
  (1):\penalty0 35--39, 1990.

\bibitem[Tewarson and Tewarson(1973)]{tewarson1973sparse}
Reginald~P Tewarson and Reginald~P Tewarson.
\newblock \emph{Sparse matrices}, volume~69.
\newblock Academic Press New York, 1973.

\bibitem[Thomas et~al.(2018)Thomas, Gu, Dao, Rudra, and
  R{\'e}]{thomas2018learning}
Anna Thomas, Albert Gu, Tri Dao, Atri Rudra, and Christopher R{\'e}.
\newblock Learning compressed transforms with low displacement rank.
\newblock In \emph{Advances in neural information processing systems
  (NeurIPS)}, pages 9052--9060, 2018.

\bibitem[Udell and Townsend(2019)]{udell2019big}
Madeleine Udell and Alex Townsend.
\newblock Why are big data matrices approximately low rank?
\newblock \emph{SIAM Journal on Mathematics of Data Science}, 1\penalty0
  (1):\penalty0 144--160, 2019.

\bibitem[Vaswani et~al.(2017)Vaswani, Shazeer, Parmar, Uszkoreit, Jones, Gomez,
  Kaiser, and Polosukhin]{vaswani2017attention}
Ashish Vaswani, Noam Shazeer, Niki Parmar, Jakob Uszkoreit, Llion Jones,
  Aidan~N Gomez, Lukasz Kaiser, and Illia Polosukhin.
\newblock Attention is all you need.
\newblock \emph{arXiv preprint arXiv:1706.03762}, 2017.

\bibitem[Wang et~al.(2018)Wang, Singh, Michael, Hill, Levy, and
  Bowman]{wang2018glue}
Alex Wang, Amanpreet Singh, Julian Michael, Felix Hill, Omer Levy, and Samuel~R
  Bowman.
\newblock Glue: A multi-task benchmark and analysis platform for natural
  language understanding.
\newblock \emph{arXiv preprint arXiv:1804.07461}, 2018.

\bibitem[Wang et~al.(2020)Wang, Li, Khabsa, Fang, and Ma]{wang2020linformer}
Sinong Wang, Belinda Li, Madian Khabsa, Han Fang, and Hao Ma.
\newblock Linformer: Self-attention with linear complexity.
\newblock \emph{arXiv preprint arXiv:2006.04768}, 2020.

\bibitem[Wu et~al.(2019)Wu, Fan, Baevski, Dauphin, and Auli]{wu2019pay}
Felix Wu, Angela Fan, Alexei Baevski, Yann~N Dauphin, and Michael Auli.
\newblock Pay less attention with lightweight and dynamic convolutions.
\newblock In \emph{The International Conference on Learning Representations
  ({ICLR})}, 2019.

\bibitem[Xiong et~al.(2021)Xiong, Zeng, Chakraborty, Tan, Fung, Li, and
  Singh]{xiong2021nystr}
Yunyang Xiong, Zhanpeng Zeng, Rudrasis Chakraborty, Mingxing Tan, Glenn Fung,
  Yin Li, and Vikas Singh.
\newblock Nystromformer: A {N}ystrom-based algorithm for approximating
  self-attention.
\newblock \emph{arXiv preprint arXiv:2102.03902}, 2021.

\bibitem[Yang et~al.(2019)Yang, Dai, Yang, Carbonell, Salakhutdinov, and
  Le]{yang2019xlnet}
Zhilin Yang, Zihang Dai, Yiming Yang, Jaime Carbonell, Ruslan Salakhutdinov,
  and Quoc~V Le.
\newblock Xlnet: Generalized autoregressive pretraining for language
  understanding.
\newblock \emph{arXiv preprint arXiv:1906.08237}, 2019.

\bibitem[Yuan et~al.(2021)Yuan, Chen, Wang, Yu, Shi, Tay, Feng, and
  Yan]{yuan2021tokens}
Li~Yuan, Yunpeng Chen, Tao Wang, Weihao Yu, Yujun Shi, Francis~EH Tay, Jiashi
  Feng, and Shuicheng Yan.
\newblock Tokens-to-token vit: Training vision transformers from scratch on
  imagenet.
\newblock \emph{arXiv preprint arXiv:2101.11986}, 2021.

\bibitem[Zaheer et~al.(2020)Zaheer, Guruganesh, Dubey, Ainslie, Alberti,
  Ontanon, Pham, Ravula, Wang, Yang, et~al.]{zaheer2020bigbird}
Manzil Zaheer, Guru Guruganesh, Kumar~Avinava Dubey, Joshua Ainslie, Chris
  Alberti, Santiago Ontanon, Philip Pham, Anirudh Ravula, Qifan Wang, Li~Yang,
  et~al.
\newblock Big bird: Transformers for longer sequences.
\newblock \emph{Advances in Neural Information Processing Systems}, 33, 2020.

\bibitem[Zhu et~al.(2021)Zhu, Ping, Xiao, Shoeybi, Goldstein, Anandkumar, and
  Catanzaro]{zhu2021long}
Chen Zhu, Wei Ping, Chaowei Xiao, Mohammad Shoeybi, Tom Goldstein, Anima
  Anandkumar, and Bryan Catanzaro.
\newblock Long-short transformer: Efficient transformers for language and
  vision.
\newblock \emph{arXiv preprint arXiv:2107.02192}, 2021.

\end{thebibliography}
\bibliographystyle{plainnat}

\newpage

\appendix
\addcontentsline{toc}{section}{Appendix} %
\renewcommand \thepart{} %
\renewcommand \partname{}
\newpage
\part{Appendix} %
\parttoc %
\newpage
\section{Extended Related Work}
\label{sec:extended_related_work}

\subsection{Robust PCA}

Robust Principle Component Analysis (robust PCA) is the problem of finding a
composition of a matrix $M$ into a sum of sparse and low-rank components:
$M = S + L$.
It is a modification of PCA to accommodate corrupted observations (aka, noise).
The sparse part covers the noise, while the low-rank part recovers the principle
components.
The most popular method to solve the problem is convex relaxation~\citep{candes2009exact}, where one
minimizes the error $\norm{M - S - L}_F^2$ subject to $\ell_1$ constraint on
$\norm{S}_1$ and nuclear norm constraint on $\norm{L}_*$, in order to promote
the sparsity of $S$ and the low-rankness of $L$.
This convex problem can be solved with a variety of methods, such as interior
point methods or the method of Augmented Lagrange Multipliers.

In our context, to find a sparse + low-rank decomposition of the attention
matrix, one can also heuristically ``peel off'' the sparse part by finding the
large entries of the attention matrix, then find a low-rank decomposition of the
remainder.
To avoid materializing the full attention matrix, one can use LSH to find
potential locations of large entries, and use matrix
completion~\citep{recht2011simpler} to find a low-rank decomposition.
Gradient descent can find global optimum for this matrix completion
problem~\citep{de2015global}.
However, it still requires too many iterations to be used in each training step.

\subsection{Efficient Transformers}

\textbf{Sparse, Low-rank Approx.: }Transformer-based model such as BERT~\citep{lan2019albert} has achieved unprecedented performance in natural language processing. Recently, Vision Transformers~\citep{dosovitskiy2020image,yuan2021tokens} has also achieved comparable performance to the traditional convolutional neural network in computer vision tasks~\citep{wu2019pay}. However, the quadratic computation of the attention layers constrains the scalability of Transformers. There are many existing directions to overcome this bottleneck, including attention matrix approximation such as Reformer~\citep{kitaev2020reformer}, Performer~\citep{choromanski2020rethinking},  leveraging a side memory module that can access multiple tokens at once~\citep{sukhbaatar2019adaptive,likhosherstov2020sub,lan2019albert} such as Longformer~\citep{beltagy2020longformer} and BigBird~\citep{zaheer2020bigbird}, segment-based recurrence such as
Transformer-XL~\citep{dai2019transformer} and Compressive Transformer~\citep{rae2019compressive}. Please refer to a recent survey~\citep{tay2020efficient} for more details. In this paper, we mainly explore within the scope of approximating dense or full attention matrices.

\textbf{Existing combination of Sparse and Low-rank Attention: }
Our focus on the classical and well-defined problem of matrix approximation, as opposed to simply designing an efficient model that performs well on downstream tasks (e.g., Longformer, Luna, Long-short transformer, etc.) affords us several advantages: (i) Easier understanding and theoretical analysis (Section 3, 4). We see that Scatterbrain yields an unbiased estimate of the attention matrix, and we can also understand how its variance changes. (ii) Clear-cut evaluation based on approximation error, as well as the ability to directly replace a full attention layer with Scatterbrain attention without re-training (Section 5). This setting is increasingly important as transformer models are getting larger and training them from scratch has become prohibitively costly. Other methods such as Luna and Long-short transformer are not backward compatible with pre-trained models.

Here we compare Scatterbrain with other work mentioned by the reviewer, showing how most of them are special cases of Scatterbrain. We will also add this discussion in the updated version of the manuscript.
\begin{itemize}[leftmargin=*,nosep,nolistsep]
    \item Longformer~\citep{beltagy2020longformer}: a special case of Scatterbrain where the sparse component is local attention, and the low-rank component is the global tokens. Global tokens can be considered a restricted form of low-rank approximation.
    \item BigBird~\citep{zaheer2020bigbird}: a special case of Scatterbrain where the sparse component is local + random sparse attention, and the low-rank component is the global tokens. The use of global tokens makes the model unsuited for autoregressive modeling. On the other hand, Scatterbrain’s generality allows it to use other kinds of low-rank attention (e.g., Performer), and thus Scatterbrain works on both the causal/autoregressive and the bidirectional/non-causal attention settings. BigBird’s motivation is also quite different from ours: they aim to design efficient attention such that the whole Transformer model is still a universal approximator and is Turing complete. Our goal is more concrete and easier to evaluate: we approximate the attention matrices, to get a small Frobenius error between the Scatterbrain attention and the full attention matrices.
    \item Luna~\citep{ma2021luna} (concurrent work): they use a fixed-length extra sequence and two consecutive attention steps: the context sequence attends to the extra sequence, and then the query sequence attends to the extra sequence. This is similar in spirit to low-rank attention (Linformer) and global tokens, but it is not a low-rank approximation due to the non-linearity between the two attention steps. It is not clear to us that it combines different kinds of attention.
    \item Long-short transformer\citep{zhu2021long} (concurrent work): a special case of Scatterbrain where the sparse component is local attention and the low-rank component is Linformer.
\end{itemize}

\subsection{Locality Sensitive Hashing for Efficient Neural Network Training}
Locality Sensitive Hashing (LSH) has been well-studied in approximate nearest-neighbor search~\cite{gionis1999similarity,indyk1998approximate,shrivastava2014asymmetric,NIPS2015_5893,dong2019learning,chen2018densified}. Since the brute-force approach for similarity search is computationally expensive, researchers have come up with various indexing structures to expedite the search process. Usually this comes with trade-offs on the search quality. Based on these indexing structures, one can achieve sub-linear search time. LSH has been used in estimation problem as well~\citep{chen2019fast,chen2018unique}.

Recently, there has been several work taking advantage of LSH data structures for efficient neural network training. During training process, the weight matrices are slowly modified via gradients derived from objective functions. If we consider the weights as the search data and input as queries, we can view neural network training as a similarity search problem. For example, \citep{chen2019slide,daghaghi2021tale,liu2020climbing} proposes an algorithm which performs sparse forward and backward computations via maximum inner product search during training. It is based on the observation that the model is usually over-parameterized so the activation for a given input could be sparse and LSH is used to find or impose the sparse structure. Similarly, LSH based algorithms have also been used in Transformers~\citep{chen2019slide, chen2020mongoose}, where LSH is used to capture the sparse structure of the attention matrices. They can largely reduce the memory bottleneck of self-attention modules especially over long sequences in Transformer. Though~\citep{chen2020mongoose} has done some exploration to improve LSH accuracy-efficiency trade-offs through learnable LSH, most of the above works have limited understanding on when and where LSH can perform well. 

\subsection{Structured Matrices for Efficient Machine Learning Models}

Sparse + low-rank is an example of a class of structured matrices: those with
asymptotically fast matrix-vector multiplication algorithm ($o(n^2)$ time
complexity) and few parameters ($o(n^2)$ space complexity).
Common examples include sparse, low-rank matrices, and matrices based on fast
transforms (e.g., Fourier transform, circulant, Toeplitz, Legendre transform,
Chebyshev transform, and more generally orthogonal polynomial transforms).
These classes of matrices, and their generalization, have been used in machine
learning to replace dense matrices in fully connected, convolutional, and
recurrent layers~\citep{sindhwani2015structured,thomas2018learning,gu2020hippo}.
\citet{de2018two} shows that any structured matrix can be written as product of
sparse matrices, and products of sparse matrices even with fixed sparsity
pattern have been shown to be effective at parameterizing compressed
models~\citep{dao2019learning,alizadeh2019butterfly,dao2020kaleidoscope}.

In our setting, it remains difficult to approximate the attention matrix with
these more general classes of structured matrices.
This is because many of them are fixed (e.g., Fourier transform, orthogonal
polynomial transforms), and there lacks efficient algorithms to find the closest
structured matrix to a given attention matrix.

\newpage
\section{Motivating Observations: Low-rank and Sparse Structures of Attention Matrices}
\label{sec:observation_details}

We aim to build a deeper understanding of sparse and low-rank structures in real attention matrices: where each of them excel, and the potential for their combination.
Specifically, we
\begin{itemize}[leftmargin=*,nosep,nolistsep]
\item show that sparse and low-rank approximation errors are negatively correlated (through statistical tests),
\item characterize regimes where each of sparse and low-rank approximation are well-suited, as dictated by the entropy of the softmax attention distribution, and
\item demonstrate that sparse + low-rank has the potential to achieve better approximation than either.
\end{itemize}

\subsection{Setup}

\begin{figure}
\captionsetup{font=small}
	\begin{center}
	\vspace{-0.6cm}
	\scriptsize
		\begin{tabular}{cc}
		    \hspace{-0.5cm}
			\includegraphics[width=0.43\linewidth]{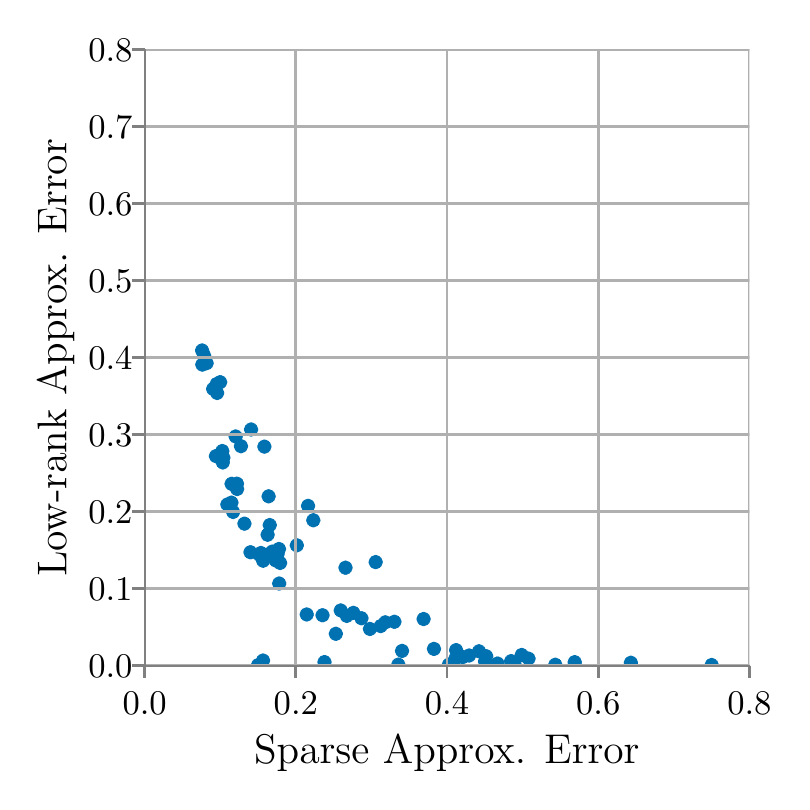}
			\includegraphics[width=0.43\linewidth]{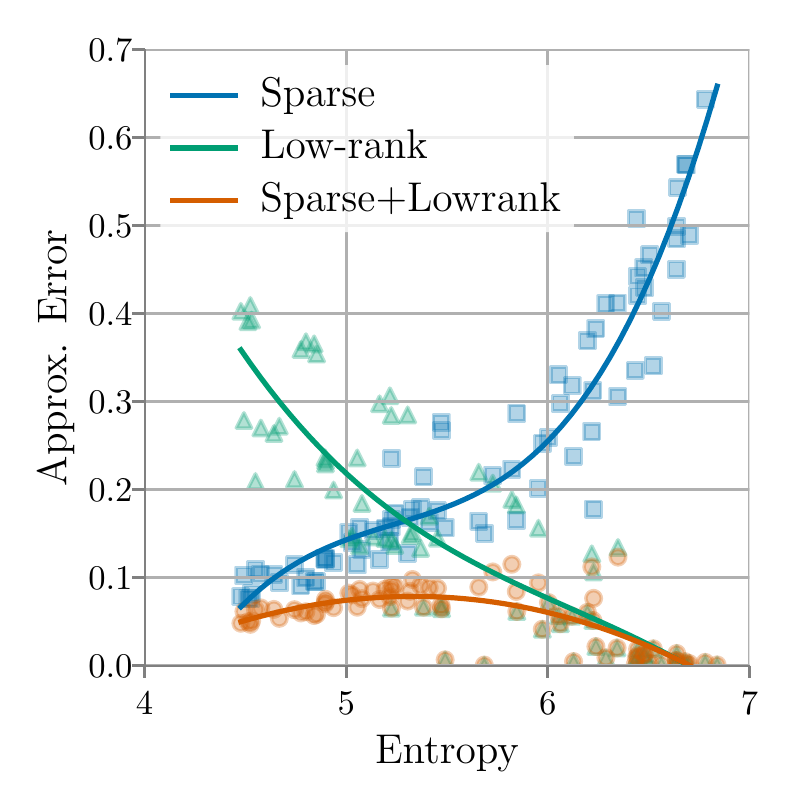}
			\\
			\hspace{-0.5cm}
			\includegraphics[width=0.43\linewidth]{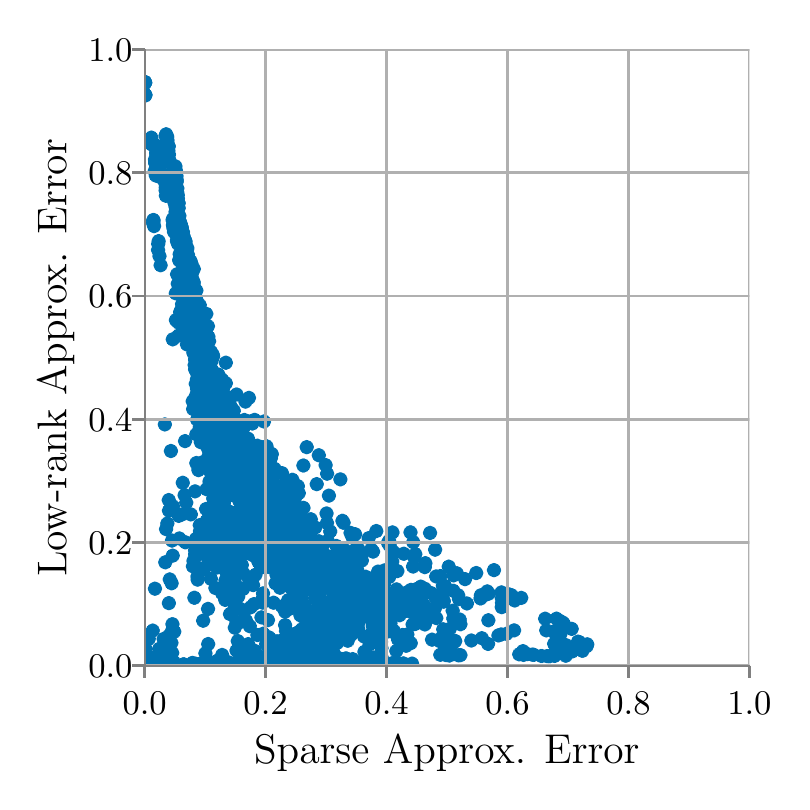}
			\includegraphics[width=0.43\linewidth]{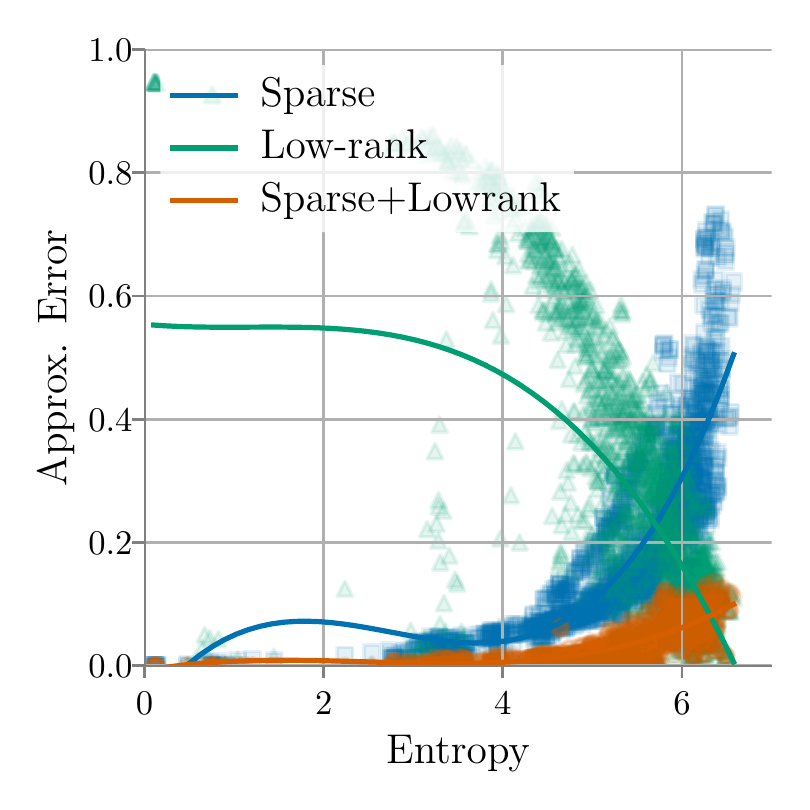}
						\\
			\hspace{-0.5cm}
			\includegraphics[width=0.43\linewidth]{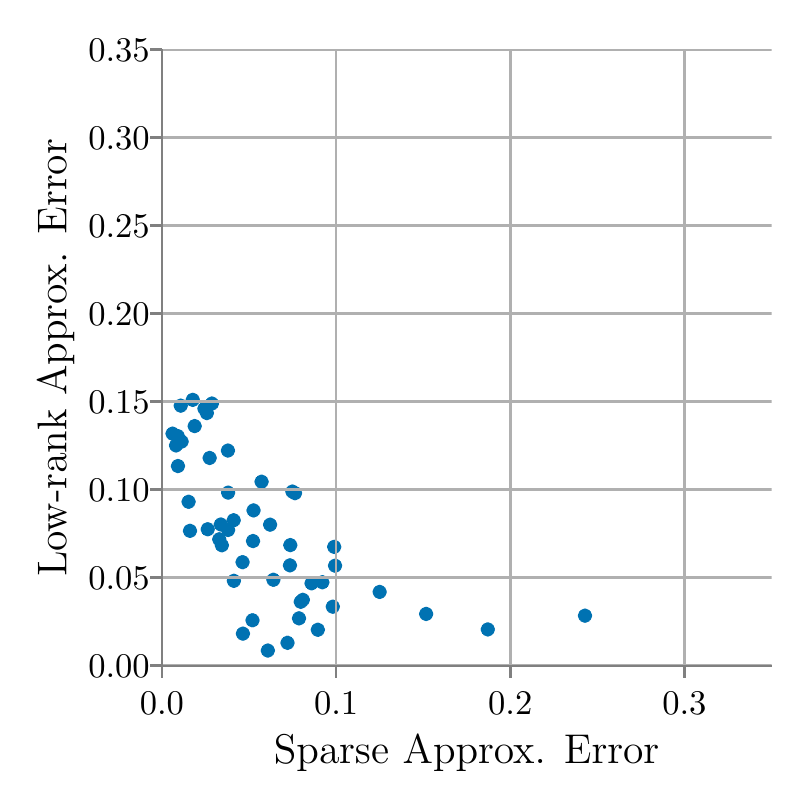}
			\includegraphics[width=0.43\linewidth]{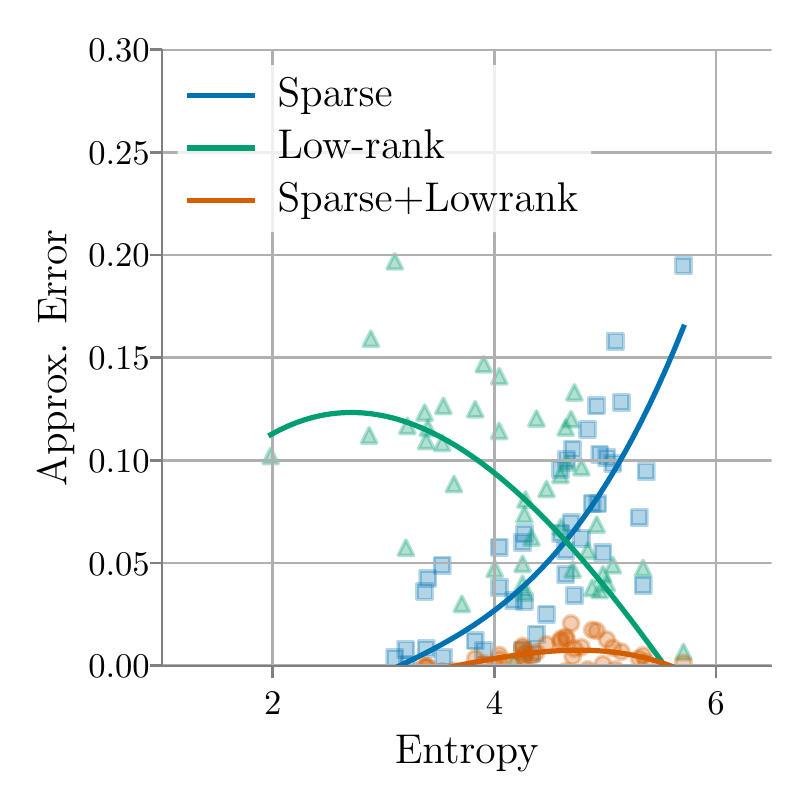}
		\end{tabular}
	\end{center}
	\caption{Characterization of the relationship between the softmax distribution of each attention matrix row and approximation error of sparse, low-rank and sparse+low-rank. The top, middle and bottom plots are for IMDb, WikiText103 and BigGAN-ImageNet respectively. Left: The approximation error of sparse and low-rank are negatively
      correlated. Sparse performs well when low-rank does not, and vice versa. Right: Entropy of the softmax attention distribution (i.e., scale of
      logits) determines the regimes where sparse, low-rank, or sparse +
      low-rank perform well. Sparse + low-rank yields better approximation than
      sparse or low-rank alone, across the board.}
	\vspace{-0.6cm}
	\label{fig:obs_full} 
\end{figure}

Denote $M$ as the attention matrix (after softmax) and $\mathcal{H}$ as entropy.
We measure approximation error by the Frobenius norm or the original matrix and the approximation (sparse or low-rank).
All the observed attention matrices in this section are from (1) a 4-layer vanilla Transformer trained from scratch on char-level IMDb reviews classification~\citep{tay2020long} (2) a 16-layer vanilla Transformer trained from scratch on WikiText103~\citep{merity2016pointer} (3) a 1-layer (attention) pre-trained BigGAN on ImageNet~\citep{5206848}. To collect attention matrices for IMDb and WikiText103, we first save checkpoint of the models in every epoch; then evaluate 100 samples from validate data for each checkpoint and collect attention matrices from each layer each head. Note we take the median of the stats (error) for those 100 samples if it is difficult to visualize. To collect attention matrices for BigGAN, we generate 100 samples and collect the attention on the fly.

\subsection{Observation 1: Sparse and low-rank approximation errors are negatively correlated}

\begin{table*}[h]
    \scriptsize
\captionsetup{font=small}
	\centering
    	\caption{The Spearman's rank, Pearson and Kendall's Tau correlation coefficients between
  Sparse and Low-rank approx.\ error on IMDb, WikiText-103, and BigGAN-ImageNet.
  P-values of $< 0.05$ indicate statistical significance.
  The two errors are negatively correlated.}
    	\resizebox{\linewidth}{!}{
	\centering
	\begingroup
	\setlength{\tabcolsep}{10pt}
	\renewcommand{\arraystretch}{1.3}
	\begin{tabular}{c|cc|cc|cc}
    \specialrule{.15em}{.05em}{.05em}
    \multirow{1}{*}{}& \multicolumn{2}{c|}{\multirow{1}{*}{IMDb}} & \multicolumn{2}{c|}{\multirow{1}{*}{WikiText103}}&
    \multicolumn{2}{c}{\multirow{1}{*}{BigGAN-ImageNet}}\\
    \cline{2-7}
    &  Coef & p-value & Coef & p-value & Coef & p-value  \\
		\hline
	\multirow{1}{*}{Spearman's rank} & -0.89 &$<$ .00001 &  -0.63 & $<$ .00001 &  -0.21&$<$ .00001 \\
	\cline{1-7}
	\multirow{1}{*}{Pearson} & -0.78& $<$ .00001& -0.61 & $<$ .00001 & -0.31 &$<$ .00001 \\
	\cline{1-7}
	\multirow{1}{*}{Kendall's Tau} &-0.74 & $<$ .00001 & -0.51& $<$ .00001 & -0.18 &$<$ .00001 \\
	\specialrule{.15em}{.05em}{.05em}
	\end{tabular}
	\endgroup
	}
	\label{table:test}
\end{table*}

We fixed the number of parameters, $K$, allowed for each attention matrix
approximation and collect the errors from ideal sparse and low-rank
approximations: top$-K$ entries for each row of the matrix for sparse and
top$-K$ eigenvalues for low-rank. Then we run three standard statistical correlation tests~\citep{artusi2002bravais,taylor1990interpretation}, Spearman, Pearson and Kendall's Tau on sparse and low-rank approximation error for all the matrices.
We can see from Table~\ref{table:test} that errors are significantly negatively
correlated (p-value $< 0.05$). 
Further more, the left three plots on Figure~\ref{fig:obs_full} visualizes the correlation between the two errors on three datasets. 

This negative correlation suggests that there is some property of the softmax attention distribution which determines when sparse or low-rank excels. We validate this claim in the next observation.

\subsection{Observation 2: Sparse approximation error is lower when softmax entropy is low and low-rank approximation error is lower error when entropy is high}

We visualize the sparse and low-rank approximation error against the entropy of
attention matrices $\mathcal{H}(M)$ (applied to each row, then averaged) on the
right plot in Figure~\ref{fig:obs_full}.
The attention matrices are $\in \mathbb{R}^{1024 \times 1024}$ (padded) so the x-axis has
range from $[0, \ln(1024)]$.
For high-entropy distributions (more diffused) low-rank matrices approximates
the attention matrix well.
For low-entropy distributions (more peaked), sparse matrices are better-suited. 

This implies that sparse and low-rank approximations could be complementary: if we can combine the strength of both, it is possible to come up with a better approximation across more general scenarios. Therefore, in the next observation, we try to combine sparse and low-rank approximations.

\subsection{Observation 3: Sparse + Low-rank achieves better approximation error
  than sparse or low-rank alone}
We find an approximation of the attention matrix of the form $S + L$, where $S$
is sparse and $L$ is low-rank.
This problem has a rich history and is commonly solved with Robust PCA.
As shown in~\ref{fig:obs_full}, across the range of entropy, sparse +
low-rank approximation can achieve lower error than either sparse or low-rank
when choosing the correct mix ratio of sparse and low rank approximation ideally
(with robust-PCA).

Motivated by the fact that sparse and low-rank approximations of attention matrices have complementary strengths (Observations 1 and 2),
one might want to combine them (Observation 3) in hope of yielding a more robust approximation that works well across different kinds of attention matrices.
The above introduces three main challenges that we have addressed in the main paper:
\begin{itemize}[leftmargin=*,nosep,nolistsep]
\item how to find sparse + low-rank decomposition of an attention matrix that is compute efficient (the most studied algorithm, robust PCA, is orders of magnitude too slow to be done at each training iteration) and memory efficient (i.e., without materializing the full matrix) (\cref{sec:algorithm}),
\item if we can find such a sparse + low-rank decomposition, how accurate is the approximation (\cref{sec:algo_analysis}),
\item how expressive is the sparse + low-rank parameterization, i.e., are there natural classes of matrices where sparse + low-rank yields asymptotically better approximation than sparse or low-rank alone) (\cref{sec:theory})?
\end{itemize}

\newpage
\section{Scatterbrain Algorithm and Implementation Details}
\label{sec:implementation}

Let $Q, K \in \mathbb{R}^{n \times d}$ be the
query and key matrices respectively, and $V \in \mathbb{R}^{n \times d}$ be the value matrix. Let the rows of $Q$ be $q_1, \dots, q_n$, and the rows of $K$ be $k_1, \dots, k_n$. The attention computes:
\begin{equation*}
  \softmax(Q K^\top) V,
\end{equation*}
with $\softmax$ applied row-wise, where for each vector $v \in \mathbb{R}^n$,
$\softmax(v) = \frac{1}{\sum_{j=1}^{n} e^{v_j}} \begin{bmatrix} e^{v_1}, \dots, e^{v_n} \end{bmatrix}^\top.$
Here we omit the usual scaling of $\frac{Q K^\top}{\sqrt{d}}$ for simplicity
since that could be folded into $Q$ or $K$.
Note that $\softmax(Q K^\top) = D^{-1} \exp(Q K^\top)$, where the exponential
function is applied element-wise and $D$ is a diagonal matrix containing the
softmax normalization constants
($D_{i, i} = \sum_{j=1}^{n} \exp(q_i^\top k_j)$).
Then attention has the form $D^{-1} \exp(Q K^\top) V$.

We describe the Scatterbrain approximation algorithm in~\cref{alg:scatterbrain}.
This includes the normalization step.
\begin{algorithm}[H]
\begin{algorithmic}[1]
{\small
\State {\bf Input: }$Q, K, V \in \R^{n\times d}$, hyper-parameters $m, k, l$
  \Procedure{\textsc{Init}}{$m, k, l$}
  \State Sample $W \in \mathbb{R}^{m \times d}$ where
  $W_{i} \sim \mathcal{N}(0, 1)$ i.i.d.
  \State Kernels $\phi \colon \mathbb{R}^{d} \mapsto \mathbb{R}^m$, $\phi(x) = \frac{\exp \left(Wx - \norm{x}^2/2\right)}{\sqrt{m}}$
  \State Hash $\forall l\in [L]$, ${\cal H}_l = \{h_{l,k}\}_{k \in [K]}$, ${\cal H} = \cup_{l \in [L] } {\cal H}_l $
  \EndProcedure
  \Procedure{\textsc{LowRankApprox}}{$Q, K, V, \phi$}
    \State $\wt{Q} = \phi(Q)$, $\wt{K} = \phi(K)$ \Comment{applied to each row}
    \State \Return $\wt{Q} ({\wt{K}}^\top V)$, $\wt{Q} ({\wt{K}}^\top) 1_n$. %
  \EndProcedure
  \Procedure{\textsc{SparseApprox}}{$Q, K, V, \phi, \mathcal{H}$}
  \State $\mathcal{S} = \{(i, j)| \mathcal{H}(Q_i)=\mathcal{H}(K_j)\}$
  \State $S \leftarrow$ sparse matrix whose support is $\mathcal{S}$
  \For {$(i, j) \in \mathcal{S}$}
  \State $S_{ij} = \exp(q_i^\top k_j) - \phi(q_i)^\top \phi(k_j)$.
  \EndFor
  \State \Return $SV$, $S 1_n$. %
  \EndProcedure
  \Procedure{ScatterbrainApprox}{$Q, K, V$}
  \State $\phi$, $h$ $\leftarrow$ \textsc{Init}($m,k,l$).
  \State $O_\mathrm{lr}, D_\mathrm{lr} \leftarrow$ \textsc{LowRankApprox}($Q, K, V, \phi$).
  \State $O_\mathrm{s}, D_\mathrm{s} \leftarrow$ \textsc{SparseApprox}($Q, K, V, \phi, h$).
  \State \Return
  $\diag(D_\mathrm{lr} + D_\mathrm{s})^{-1} (O_\mathrm{lr} + O_\mathrm{s})$.
  \EndProcedure
}
\end{algorithmic}
\caption{Scatterbrain Approximation of Attention}\label{alg:scatterbrain}
\end{algorithm}

\paragraph{Autoregressive / Causal / Unidirectional Attention}
To approximate autoregressive attention, we simply use the autoregressive
variant of low-rank attention, and apply the autoregressive mask to the sparse
attention.
In particular, let $M \in \mathbb{R}^{n \times n}$ be the autoregressive mask,
whose lower triangle is all ones and the rest of the entries are zero.
The unnormalized attention matrix is $\exp((Q K^\top) \odot M)$, and the
unnormalized output is $\exp((Q K^\top) \odot M) V$, where $\odot$ is
elementwise multiplication.

The low-rank autoregressive variant computes $((\wt{Q} {\wt{K}}^\top) \odot M) V$,
though with a custom GPU kernel / implementation so as not to materialize the
$n \times n$ matrix.
For the sparse component, we simply mask out locations $S_{ij}$ where $i > j$.
That is, we can perform $S \odot M$ efficiently.
As a result, we can compute the Scatterbrain output
$((\wt{Q} {\wt{K}}^\top) \odot M) V + (S \odot M) V$ efficiently.
\newpage
\section{Proofs}
\label{sec:proofs}

\subsection{Expressiveness of Sparse + Low-rank Matrices}
\label{subsec:expressiveness_proofs}

To motivate the use of sparse + low-rank matrices, we describe a family of attention matrices where sparse + low-rank matrices need asymptotically fewer parameters to approximate the attention matrix, compared to sparse or low-rank matrices alone.
For there cases, either sparse or low-rank alone requires a quadratic number of parameters ($O(n^2)$, where $n \times n$ is the dimension of the attention matrix) to get $\epsilon$ approximation error in Frobenius norm, while sparse + low-rank only requires $O(n)$ parameters.

We construct a matrix family that shows the separation between the approximation
capability of sparse + low-rank vs.\ sparse or low-rank alone.
More specifically, we will use diagonal + low-rank (a special case of sparse +
low-rank).

\begin{example}
  \label{ex:separation_1}
  Let $\epsilon$ denote a parameter that satisfies $\epsilon \in (0,1/2]$. %
  Consider the following randomized construction of a matrix
  $Q \in \mathbb{R}^{n \times d}$ with $d \geq 6 \epsilon^{-2} \log n$ and $d = \Theta(\epsilon^{-2} \log n)$,
  where each entry of $Q$ is picked
  independently and uniformly at random from
  $\{\pm 1/\sqrt{d} \}$.
  Let $M = \sigma(Q Q^{\top})$ where $\sigma$ is the elementwise exponential
  function (we first ignore the normalization term of softmax here).
\end{example}

It can be shown (e.g.\ by Hoeffding's inequality) that with high probability
\begin{align*}
  (Q Q^\top)_{i, j} =
  \begin{cases}
    1, & \text{if } i=j;\\
    \in [-\epsilon, \epsilon], & \text{otherwise} .
  \end{cases}
\end{align*}

Since $M=\sigma(QQ^{\top})$ where $\sigma$ is the elementwise exponential function,
\begin{align*}
  M_{i, j} =
  \begin{cases}
    e, & \text{if } i=j ;\\
    \in [1-O(\epsilon), 1+O(\epsilon)], & \text{otherwise} .
  \end{cases}
\end{align*}

Intuitively, as the attention matrix $M$ has large diagonal entries, low-rank
matrices will not be able to approximate it well.
However, the off-diagonals are also of reasonable size, thus making sparse
approximation difficult.
With sparse + low-rank, we can use the sparse part to represent the diagonal,
and the low-rank part to represent the remaining elements, allowing it to
approximate this matrix well.
We formalize this separation in the theorem below.

\begin{theorem}
  \label{thm:sparse_lowrank_1}
  Let $M$ be the attention matrix from~\cref{ex:separation_1}.
  For any $\gamma \in [0, 1]$, with probability at least $1 - n^{-1}$, there exists a sparse + low-rank estimator with
  $O(\gamma^{-1} n^{3/2} \log n)$ parameters that achieve $\gamma\sqrt{n}$
  Frobenius error.
  For any matrix $R \in \R^{n \times n}$ with rank such that $n-\rank=\Omega(n)$
  (e.g., $R$ has $o(n^2)$ parameters), with probability at least $1 - n^{-1}$, we have
  $\| M-R \|_F \geq \Omega(\sqrt{n})$.
  Moreover, any matrix $E_\mathrm{S}$ that has row sparsity $k$ (each row has less than
  $k$ non-zeros) such that $n - k = \omega(1)$ (e.g., $E_\mathrm{S}$ has
  $o(n^2)$ parameters) will have error $\| M-E_S \|_F \geq \Omega(\sqrt{n})$
  with probability at least $1 - n^{-1}$.
\end{theorem}
We see that for any $\gamma \in [0, 1]$, any low-rank or sparse estimator for
$M$ with $\o(n^2)$ parameters has $\Omega( \gamma^{-1} )$ times the error
of the sparse + low-rank estimator with $O( \gamma^{-1} n^{1.5} \log n)$
parameters.

\begin{proof}[Proof of \cref{thm:sparse_lowrank_1}]

For each $i \in [n]$, let $q_i$ denote the $i$-th row of
$Q \in \R^{n \times d}$.
Define $J \in \mathbb{R}^{n \times  n}$ to be the all 1s matrix.
Define $T = M-J-QQ^{\top}$. Therefore,
\begin{align*}
T_{i, j} =
\begin{cases}
  e-2 & \text{if } i=j\\
  e^{ q_i^\top q_j} - 1 - q_i^\top q_j & \text{otherwise}
\end{cases}.
\end{align*}
By Hoeffding's inequality, for a pair $i \neq j$, we have that
\begin{equation*}
  \P \left( \abs{q_i^\top q_j - \E[q_i^\top q_j]} \geq \epsilon \right) \leq 2 \exp \left( -\frac{2\epsilon^2}{ \left( \frac{1}{\sqrt{d}} - \frac{-1}{\sqrt{d}} \right)^2} \right) = 2 \exp(-d \epsilon^2 / 2).
\end{equation*}
Note that $\E[q_i^\top q_j] = 0$.

By a union bound over all pairs $i \neq j$ (there are $n(n-1) / 2$ such pairs),
with probability at least $1 - n^2 \exp \left( -d\epsilon^2 / 2 \right)$, we
have that
\begin{equation*}
  q_i^\top q_j \in [-\epsilon, \epsilon] \quad \text{for all } i \neq j.
\end{equation*}
Since we assume that $d \geq 6 \epsilon^{-2} \log n$, we have that
\begin{equation*}
  n^2 \exp (-d \epsilon^2/2) \leq n^2 \exp (-3 \log n) = n^{-1}.
\end{equation*}
Hence $q_i^\top q_j \in [-\epsilon, \epsilon]$ for all $i \neq j$ with
probability at least $1 - n^{-1}$.
For the rest of the proof, we only consider this case (where
$q_i^\top q_j \in [-\epsilon, \epsilon]$ for all $i \neq j$).

Since $1+x\leq e^x \leq 1+x+ x^2$ for $|x|<1$, we can bound the off diagonal
elements $|T_{i, j}| \leq \epsilon^2$.
In particular, for all $i \neq j$,
\begin{equation}
  \label{eq:bound_t}
  \abs{T_{ij}} = \abs{e^{q_i^\top q_j} - 1 - q_i^\top q_j} \leq \left(q_i^\top q_j\right) \leq \epsilon^2.
\end{equation}

\paragraph{Sparse + low-rank estimator:}

We use the following sparse + low-rank estimator:
\begin{align*}
E_{\mathrm{SL}} = \underbrace{(e-2)\cdot I}_{ \mathrm{sparse}} + \underbrace{J+ QQ^{\top}}_{ \mathrm{low-rank}},
\end{align*}
where $(e-2)I$ has row sparsity 1 and
$\rank(J + Q Q^\top) \leq d+1 = O\left(\epsilon^{-2} \log n\right)$.

Notice that the $E_\mathrm{SL}$ estimate matches $M$ exactly on the diagonal,
and on the off-diagonal it differs from $M$ by $T_{ij}$.
Thus, the Frobenious error of the sparse + low-rank estimator is
$$\Vert M - E_\mathrm{SL} \Vert_F \leq \epsilon^2 \sqrt{n(n-1)} \leq \epsilon^2n.$$

Set $\epsilon = \frac{\sqrt{\gamma}}{n^{1/4}}$ for $0 \leq \gamma \leq 1$, Then

(i) The sparse + low-rank parameter count is $n + n \cdot \rank \leq n \cdot O( \epsilon^{-2} \log n ) \leq  O(\gamma^{-1} n^{1.5} \log n )$.

(ii) The Frobenius error is $\leq \gamma \sqrt{n}$.

\paragraph{Low-rank estimator:} We want to argue that low-rank approximation would
require more parameters.
If we approximate the matrix $(e - 2)I$ by a matrix $R$ with rank $r$, then the
difference matrix will have at least $n - d$ singular values of magnitude
$e - 2 \geq 1/2$.
As a result, by the Eckart–Young–Mirsky theorem,
$$\Vert (e-2) \cdot I - R \Vert_F \geq \frac{1}{2} \sqrt{n-r}.$$

Define $T' = T-(e-2) \cdot I$, then $T'$ is all 0 on the diagonal and has
absolute value $\leq \epsilon^2$ on off-diagonal entries.
Thus $\Vert T'\Vert_F \leq \epsilon^2 n = \gamma\sqrt{n}$.

We want to show that if $R^{'}$ is a rank $r'$ matrix, then
$\Vert M-R' \Vert_F \geq \frac{1}{2} \sqrt{n-r'-d-1} - \Vert T' \Vert_F$.
We argue by contradiction.
Suppose that there exists some matrix $R'$ with rank $r'$ such that
$$\Vert M-R' \Vert_F \leq \frac{1}{2} \sqrt{n-r'-d-1} - \Vert T' \Vert_F.$$
Define $R = R'-J-QQ^{\top}$, so $M-R' = (e-2) \cdot I - R + T'$.
We see that:
\begin{align*}
    \| (e-2) \cdot I - R \|_F &= \| M - R' - T' \|_F \notag \\
    & \leq \| M - R' \|_F + \| T' \|_F \notag \\
    & \leq \frac{1}{2} \sqrt{n-r'-d-1} \notag \\
    & \leq \frac{1}{2} \sqrt{n-\rank(R)}.
\end{align*}
This contradicts the result above, which states that
$\norm{(e - 2)\cdot I - R}_F \geq \frac{1}{2} \sqrt{n - \rank(R)}$.

Therefore any low-rank estimator with rank $r$ such that $n - r = \Omega(n)$,
which has $\Omega(n^2)$ parameters, will have error at least
$\Omega(\sqrt{n - r - d - 1}) - \norm{T'}_F = \Omega(\sqrt{n})$, which is
$\Omega(\gamma^{-1})$ times the error of the sparse + low-rank estimator above.

\paragraph{Sparse estimator:} For our sparse estimator, it is easy to see that for any $E_\mathrm{S} \in \mathbb{R}^{n \times n}$ that has row sparsity $k$ (each row has fewer than $k$ non-zeros),
\begin{align*}
  \| M-E_\mathrm{S} \|_F \geq \Omega(\sqrt{n(n-k)}).
\end{align*}
This implies that in order to achieve error $O(\sqrt{n})$, we would need
$n - k = O(1)$, which requires $\Omega(n^2)$ parameters.

\end{proof}

Now we construct a matrix that shows better separation between the approximation capability of sparse + low-rank vs sparse or low-rank alone.

\begin{example}
  \label{ex:separation_2}
  Consider the following randomized construction of matrix
  $Q \in \mathbb{R}^{n \times d}$ with $d \geq 6 \epsilon^{-2} r \log n$ and $d = \Theta(\epsilon^{-2} r \log n)$
  ($\epsilon \in (0, 1]$ and close to 0 and $r$ is $\Theta(\log{n})$): each
  entry of $Q$ is picked independently and uniformly at random from
  $\{\pm \sqrt{r/d} \}$.
  Let $M = \sigma(Q Q^{\top})$ where $\sigma$ is the elementwise exponential
  function.
\end{example}

Similar to~\cref{ex:separation_1}, with high probability, we have:
\begin{align*}
(Q Q^\top)_{i, j} =
\begin{cases}
  r, & \text{if } i=j;\\
  \in [-\epsilon , \epsilon], & \text{otherwise} .
\end{cases}
\end{align*}
We also have:
\vspace{-0.2cm}
\begin{align*}
M_{i, j} =
\begin{cases}
  e^r, & \text{if } i=j ;\\
  \in [1-O(\epsilon), 1+O(\epsilon)], & \text{otherwise} .
\end{cases}
\end{align*}

By setting $r$ appropriately, we can formalize the separation between the
approximation ability of sparse, low-rank, and sparse + low-rank matrices:
\begin{theorem}
  \label{thm:sparse_lowrank_2}
  Let $M$ be the attention matrix from~\cref{ex:separation_2}.
  Any sparse or low-rank estimator of $M$ needs $\Omega(n^2)$ parameters for
  $O(n)$ error with probability at least $1 - n^{-1}$ while a sparse + low-rank estimator needs $O(n)$ parameters for
  $O(n)$ error with probability at least $1 - n^{-1}$.
\end{theorem}

\begin{proof}[Proof of \cref{thm:sparse_lowrank_2}]
Similar to the proof of \cref{thm:sparse_lowrank_1}, by Hoeffding's inequality, for a pair $i \neq j$, we have that
\begin{equation*}
  \P \left( \abs{q_i^\top q_j - \E[q_i^\top q_j]} \geq \epsilon \right) \leq 2 \exp \left( -\frac{2\epsilon^2}{ \left( \frac{r}{\sqrt{d}} - \frac{-r}{\sqrt{d}} \right)^2} \right) = 2 \exp\left(-\frac{d \epsilon^2}{2r}\right).
\end{equation*}
Note that $\E[q_i^\top q_j] = 0$.
By a union bound over all pairs $i \neq j$ (there are $n(n-1) / 2$ such pairs),
with probability at least $1 - n^{-1}$ (since $d \geq 6 \epsilon^{-2} r \log n$), we
have that
\begin{equation*}
  q_i^\top q_j \in [-\epsilon, \epsilon] \quad \text{for all } i \neq j.
\end{equation*}
Since we assume that $d \geq 6 \epsilon^{-2} \log n$, we have that
For the rest of the proof, we only consider this case (where
$q_i^\top q_j \in [-\epsilon, \epsilon]$ for all $i \neq j$).

Let $T = M - (e^{r} - 1) \cdot I + J$, where $J$ is the all one matrix.
We see that $T$ is zero on the diagonal.
Moreover, using the fact that $e^x \leq 1 + 2 \abs{x}$ for all $x \in [-1, 1]$,
the off-diagonal entries of $T$ have of magnitude at most $2\epsilon$.

We consider 3 different estimators.

\paragraph{Sparse + low-rank estimator:}
Our estimator is
\begin{align*}
E_\mathrm{SL} = \underbrace{(e^r-1)\cdot I}_{ \mathrm{sparse}} + \underbrace{J}_{ \mathrm{low-rank}},
\end{align*}
where $(e-1)I$ has row sparsity 1 and $\rank(J) = 1$.

The Frobenious error of sparse + low-rank approximation is
\[\Vert M - E_\mathrm{SL} \Vert_F \leq O(\sqrt{\epsilon^2n(n-1)})\leq O(\epsilon n).\]

We have that:

(i) Sparse + low-rank parameter count is $n \cdot (1+1) \leq O(n)$.

(ii) Its Frobenius error is $\leq O(n)$.

\textbf{Low-rank estimator:} We want to argue that low-rank approximation would
require more parameters.
From a similar observation that any matrix $R$ with rank that $n-\rank=\Omega(1)$,
$$\Vert (e^r-1)I - R \Vert_F \geq \Omega(e^r),$$ (by Eckart–Young–Mirsky theorem),
we obtain a similar result to the proof of~\cref{thm:sparse_lowrank_1}.

If $R^{'}$ is a matrix with rank such that $n-\rank = \Omega(1)$, then
$\Vert M-R{'} \Vert_F \geq \Omega(n) - \norm{T}_F \geq \Omega(n) -O(\epsilon n) \geq \Omega(n)$.
Hence any low-rank matrix with $O(n^2)$ parameters would have error $\Omega(n)$.

\paragraph{Sparse estimator:} Similar to the proof of~\cref{thm:sparse_lowrank_1}, for our sparse estimator, it is easy to see that for any $E_\mathrm{S} \in \mathbb{R}^{n \times n}$ that has row sparsity $k$ (each row has fewer than $k$ non-zeros),
\begin{align*}
\| M-E_\mathrm{S} \|_F \geq \Omega(\sqrt{n(n-k)}).
\end{align*}
This implies that to get $O(n)$ error, we would need $\Omega(n^2)$ parameters.
\end{proof}

\subsection{Generative Model, Softmax Temperature, and Matrix Approximation}
\label{subsec:softmax_temperature}

Here we show 3 cases where depending on the softmax temperature, either we'll need low-rank, low-rank + sparse, or sparse to
approximate the attention matrix.

We start with some notation first.
Given a matrix $B$, let $B[i,j]$ be the entry in the $i$th row and $j$th column.
For a range $[l,r]$, we define a matrix $B_{[l,r]}$ where
$B_{[l,r]}[i,j]=B[i,j]$ if $B[i,j]\in[l,r]$ and $B_{[l,r]}=0$ otherwise (that
is, $B_{[l, r]}$ only keep entries for $B$ that are in the range $[l, r]$, with
other entries zeroed out).
We write $\supp(C)$ for the set of locations of non-zeros in $C$.
We let $\lambda_i(D)$ be the $i$-th largest (in absolute value) eigenvalue of
$D$.

To prove~\cref{thm:temperature}, we first define a more general matrix class,
prove that the attention matrix in~\cref{ex:generative} is a subset of this
class (with high probability),
and then show that~\cref{thm:temperature} holds for this more general class.
We introduce an extra parameter $l \in \mathbb{R}$, in addition to the inverse
temperature $\beta$ and the intro-cluster distance $\Delta$.

\begin{matrixclass}
  \label{ex:temperature}
  Let $Q \in \mathbb{R}^{n \times d}$ with every row of $Q$ having $\ell_2$-norm
  in $[1 - O(\Delta), 1 + O(\Delta)]$, and
  let $A = Q Q^\top$.
  Further:
  \begin{enumerate}[leftmargin=*,nosep,nolistsep]
    \item Let $H$ = $A_{ \left[ 1/l, 2 - 1/l \right]}$ for some
    $l \geq \Omega(1)$.
    Assume that $H$ is block diagonal with $\Omega(n)$ blocks, and $\supp(H)$ is
    $o(n^2)$.
    That is, the large entries of $Q Q^\top$ form a block diagonal matrix.
    \item Let $L = A - H$ then $L = A_{[-\Delta, \Delta]}$ where
    $\Delta = o ( {1}/{\log d} )$.
    Assume that there is a constant fraction of elements in $\supp(L)$ falling
    in $[0, \Delta]$.
    Assume that $\supp(A_{[0, \Delta]})$ is $\Omega(n^2)$.
  \end{enumerate}
  Let $M_\beta = \exp(\beta \cdot A)$.
\end{matrixclass}

We now show that~\cref{ex:generative} is a subset of~\cref{ex:temperature}, with
high probability.
\begin{lemma}
  The matrix $M_\beta$ in~\cref{ex:generative} is a subset
  of~\cref{ex:temperature}, where $l = \frac{1}{1 - \Delta^2}$.
\end{lemma}

\begin{proof}
  We first bound the norm of each row in $Q$ in~\cref{ex:generative}.
  For any $i, j$, we have
  \begin{equation*}
    \norm{z_{ij}}^2 = \norm{c_i + r_{ij}}^2 = \norm{c_i}^2 + 2 c_i^\top r_{ij} + \norm{r_{ij}}^2.
  \end{equation*}
  Since $c_i \sim \mathcal{N}(0, I_d / \sqrt{d})$,
  $\norm{c_i}^2 \in [1 - \Delta^2, 1 + \Delta^2]$ with probability at least
  $1 - 2e^{-d \Delta^2/8}$ (by the standard argument using the fact that
  $\chi^2$-random variables are sub-exponential).
  Similarly, $\norm{r_{ij}}^2 \in [\Delta^2 - \Delta^4, \Delta^2 + \Delta^4]$
  with probability at least $1 - 2e^{-d \Delta^2/8}$.
  By concentration of measure, we can also bound
  $2 c_i^\top r_{ij} \in [2 \Delta - 2 \Delta^3, 2\Delta + 2 \Delta^3]$ as well.
  Therefore, we have that $\norm{z_{ij}}^2 \in [1 - O(\Delta), 1 + O(\Delta)]$.

  Now we show that the large entries of $Q Q^\top$ form a block diagonal matrix.
  With high probability, the large entries come from intra-cluster dot product,
  and the small entries come from inter-cluster dot product.

  We bound the intra-cluster dot product:
  \begin{align*}
    z_{ij}^\top z_{ik}
    &= (c_i + r_{ij})^\top (c_i + r_{ik}) \\
    &= \norm{c_{i}}^2 + c_i^\top r_{ij} + c_i^\top r_{ik} + r_{ij}^\top r_{ik}.
  \end{align*}
  Similar to the argument above, by concentration of measure,
  $\norm{c_i}^2 \in [1 + \epsilon\Delta, 1 - \epsilon\Delta]$ with high
  probability (we will pick $\epsilon = \theta(\Delta)$).
  The cross terms $c_i^\top r_{ij}$ and $c_i^\top r_{ik}$ can be bounded using
  Cauchy-Schwarz inequality to be in $[-\epsilon \Delta, \epsilon \Delta]$ with
  high probability.
  And the fourth term $r_{ij}^\top r_{ik}$ is in
  $[-\epsilon \Delta^2, \epsilon \Delta^2]$ with high probability.
  Therefore, the inner product is in $1 \pm O(\epsilon \Delta)$ with high
  probability.
  This satisfies the first condition in~\cref{ex:temperature}, for
  $l = \frac{1}{1 - \Delta^2}$, assuming $\epsilon \leq \Delta$.

  We use a similar argument to bound the inter-cluster dot product.
  For $i \neq i'$
  \begin{align*}
    z_{ij}^\top z_{i'k}
    &= (c_i + r_{ij})^\top (c_{i'} + r_{i'k}) \\
    &= c_i^\top c_{i'}^\top + c_i^\top r_{i'k} + c_{i'}^\top r_{ij} + r_{ij}^\top r_{i'k}.
  \end{align*}
  By concentration of measure, $c_i^\top c_{i'} \in [-\epsilon, \epsilon]$.
  Similar to the argument in the intra-cluster case, we can bound the other
  three terms, so this dot product is in $[-O(\epsilon), O(\epsilon)]$.
  This satisfies the second condition in~\cref{ex:temperature}.

\end{proof}

To prove~\cref{thm:temperature} for~\cref{ex:temperature}, we start with some
technical lemmas.
\begin{lemma}
  \label{lem:lambda_min}
  Let $F \in \mathbb{R}^{N \times N}_{\geq 0}$ be a symmetric matrix.
  Let $\lambda_{\max}$ be the largest eigenvalue of $F$.
  Assuming $N \geq 2$, we have that
  \begin{equation*}
    \lambda_{\max} \geq \min_{i \neq j} F[i, j].
  \end{equation*}
\end{lemma}

\begin{proof}
  Since $F$ is symmetric, $\lambda_{\max}$ is real and
  \begin{equation*}
    \lambda_{\max} = \max_{u \neq 0} \frac{u^\top F u}{u^T u}.
  \end{equation*}
  Let $u$ be the all 1's vector, then
  \begin{align*}
    \lambda_{\max}
    \geq & ~ \frac{1}{N} \sum_{i=j} F[i, j] \\
    \geq & ~ \frac{1}{N} \sum_{i \neq j} F[i, j] \\
    \geq & ~ \frac{1}{N} \cdot N(N-1) \min_{i \neq j} F[i, j] \\
    \geq & ~ \min_{i \neq j} F[i, j],
  \end{align*}
where the second step follows from all the diagonal entries are non-negative, the last step follows from $N \geq 2$
\end{proof}

The above implies the following result:
\begin{corollary}
  \label{cor:block_diagonal}
  Let $F \in \mathbb{R}^{N \times N}_{\geq 0}$ be a block diagonal matrix.
  Let $r$ be the number of $m \times m$ blocks in $F$ for some $m \geq 2$.
  The $\lambda_r(F)$ is at least the smallest non-diagonal element in any
  $m \times m$ block ($m \geq 2$) in $F$.
\end{corollary}

\begin{proof}
  By~\cref{lem:lambda_min}, each $m \times m$ block $B$ ($m \geq 2$) by itself
  has max eigenvalue at least $\min_{i \neq j \in [m]} B[i, j]$.
  The claim then follows from the fact that any eigenvalue of $B$ is also an
  eigenvalue of $F$.
\end{proof}

We'll need the following function for our low-rank argument:
\begin{equation*}
  f_k(x) = \sum_{i=0}^{k} \frac{x^i}{i!}.
\end{equation*}
Note that $f_\infty(x) = e^x$.
\begin{definition}
Let $\epsilon \in (0,1/10)$ and $L > 0$.
We say a function $f : \R \rightarrow \R$ is $(\epsilon,L)$-close to $e^y$ if
  \begin{equation*}
    | e^y - f(y) | \leq \epsilon \quad \text{ for any } y \in [-L, L].
  \end{equation*}
\end{definition}

\begin{lemma}
  \label{lem:k_star}
  For any $\epsilon \in (0,1/10)$ and $L > 0$. If
  \begin{equation*}
    D \geq 10( L +  \log(1/\epsilon) )
  \end{equation*}
  then function $f_D(y)$ is $(\epsilon,L)$-close to $e^y$.
\end{lemma}

\begin{proof}
Recall the definition of function $f_D$,
  \begin{align*}
  e^x = f_{D}(x) + \sum_{i=D+1}^{\infty} \frac{x^i}{i!},
  \end{align*}
It is sufficient to show that $|e^y - f(y) | < \epsilon$ if we have
\begin{align*}
  \frac{x^{D + 1}}{ (D + 1)!}
  \leq \frac{\epsilon}{2},
\end{align*}
We can show that
\begin{align*}
    \frac{y^D}{D!}
    \leq & ~ \frac{L^D}{D!} \\
    \leq & ~ \frac{L^D}{ (D/4)^D} \\
    = & ~ ( \frac{4L}{D} )^D \\
    \leq & ~ (1/2)^D \\
    \leq & ~ \epsilon/10
\end{align*}
where the first step follows from $|y| \leq L$, the second step follows $n! \geq (n/4)^n$, the forth step follows from $D \geq 10 L$, the last step follows $D \geq 10 \log(1/\epsilon)$ and $\epsilon \in (0,1/10)$.

\end{proof}

We'll also use the following fact:
\begin{lemma}
  \label{lem:rank_k_star}
  For any $D = o(\log n / \log d)$, we have
  \begin{equation*}
    \rank(f_{D}) \leq n^{o(1)}.
  \end{equation*}
\end{lemma}

\begin{proof}
We can upper bound $\rank(f_D(A))$ in the following sense:
  \begin{align*}
    \rank(f_{D}(A))
    \leq & ~ (\rank(A))^{D} \\
    \leq & ~ d^{D} \\
    = & ~ 2^{D \cdot \log d } \\
    = & ~ 2^{o(\log n)} \\
    = & ~ n^{o(1)}.
  \end{align*}
where the second step follows from $\rank(A) \leq d$, the forth step follows from $D = o( \frac{\log n}{\log d} )$.
\end{proof}

Finally we're ready to prove the theorem:

\begin{proof}
The basic idea is: (i) Use $f_{k^*}(b \cdot A)$ to get the low-rank
approximation (ii) Use $\exp(b \cdot H)$ to get the sparse part.

\paragraph{Small $\beta$ range,}
  i.e., $\beta$ is $o \left( \frac{\log n}{\log d} \right)$.

  Low rank approximation: $R = f_{k^*}(b \cdot A)$.

  Since each entry of $A$ is in $[-1, 1]$, each entry of $\beta \cdot A$ is in
  $[-\beta, \beta]$.
  But note that $\beta$ in this case is
  $o \left( \frac{\log n}{d} \right) = O(\log n \cdot \Delta)$.
  By the definition of $k^*$, each entry of
  $\exp(\beta \cdot A) - f_{k^*}(\beta \cdot A)$ has absolute value $\leq \epsilon$.
  Therefore the overall error is $\leq \epsilon n$.

  For sparse only: By assumption, $m = \Omega(\norm{L}_0)$ entries in $A$ are
  $\geq 0$, which are exactly the entries in $\exp(\beta \cdot A)$ that are
  $\geq 1$.
  Hence any (say) $\frac{m}{2}$ sparse approximation has error
  $\geq \sqrt{\frac{m}{2}} \geq \Omega(\sqrt{\norm{L}_0})$.
  By our assumption, $\norm{L}_0 = \Omega(n^2)$.

\paragraph{Mid-range $\beta$,} i.e., $\beta \geq \frac{1}{l} \cdot \log n$ and $\beta$
  is $O(\log n)$.

  Sparse only: the argument is the same as in the low $\beta$ range.

  Sparse + low-rank: The low-rank part $R = fst(\beta \cdot A)$.
  By \cref{lem:rank_k_star}, this has rank $n^{o(1)}$, so it has
  $n^{(1 + o(1))}$ parameters.

  The sparse part is $S = e^{\beta \cdot H} - R_{\supp(H)}$.
  Clearly this needs $\abs{\supp(H)}$ parameters.

  Let $E = M_\beta - (S + R)$.
  Then (i) in $\supp(H)$, $E$ is all $0$.
  (ii) output of $\supp(H)$, by definition, entries of $\beta \cdot A$ are in
  $[-\beta \Delta, \beta \Delta]$, which in the current range of $\beta$ is
  $[-O(\log n \Delta), O(\log n \Delta)]$.
  Therefore all the entries of $E$ have absolute value $\leq \epsilon$.
  By the definition of $k^*$, we have that $\norm{E}_F \leq \epsilon n$.

  Low-rank only: Let $\tilde{R}$ be rank $r - n^{o(1)} - 1$ that approximates
  $M_\beta$.
  Then using the same argument as our existing lower bound argument, we get that
  $\tilde{R} - R \approx_E S$ (this means that the error
  $\leq \norm{E}_F + \norm{M_\beta - \tilde{R}}_F$).
  Now note that $S = e^{\beta \cdot H} - (f_{k^*}(\beta \cdot A))_{\supp{H}}$ is a
  symmetric, block diagonal matrix with $r = \Omega(n)$ blocks.
  \Cref{cor:block_diagonal} implies that $\lambda_r(S)$ is at least the smallest
  non-diagonal value in $S$.
  Now the smallest non-diagonal value in $e^{\beta \cdot H}$ is
  $\geq e^{\frac{1}{l} \log n} = n$.
  On the other hand, the largest value in $(f_{k^*}(\beta \cdot A))_{\supp{H}}$ is
  \begin{align*}
    &\leq k^* \frac{\beta^{k^*}}{k^*!} \leq \beta \cdot \left( \frac{e\beta}{k^* - 1} \right)^{k^* - 1} \\
    &\lesssim \log n \left(\frac{e \cdot \log n}{\log n \cdot \Delta} \right)^{O(\log n \cdot \Delta)} \\
    &\lesssim \log n e^{O(\log n \cdot \Delta \cdot \log \frac{ 1}{\Delta})} \\
    &\lesssim \log n \cdot n^{o(1)} \\
    &= n^{o(1)}.
  \end{align*}
  Hence $\lambda_r(S)$ is $\Omega(n)$.
  The claimed result then follows since $\norm{E}_F \leq \epsilon n$ and
  $\rank{\tilde{R} - R} \leq r - 1$ (Eckart-Young-Mirsky theorem).

\paragraph{Large $\beta$ range,}
i.e., $\beta \geq \omega(\log n)$.

  Sparse only: $S = e^{\beta \cdot H}$.
  Note that each entry in $E = M_\beta - S$ is upper bounded by
  $e^{\Delta \cdot \beta} \leq e^{o \left( \frac{\beta}{\log d} \right)}$.
  Then
  \begin{align*}
    \norm{E}_F
    &\leq n \cdot e^{o \left( \frac{\beta}{\log d} \right)} \\
    &\leq \epsilon \cdot e^{\log \frac{n}{\epsilon} + o \left( \frac{\beta}{\log d} \right)} \\
    &\leq \epsilon \cdot e^{o(\beta) + o \left( \frac{\beta}{\log d} \right)} \\
    &\leq \epsilon \cdot e^{o(\beta)} \\
    &\leq \epsilon \cdot e^{\beta / l}.
  \end{align*}

  Low-rank only: since $\norm{E}_F$ is $\leq \epsilon e^{\beta / l}$, it is enough
  to argue that any rank $r$-approximation to $S$ has error $\geq e^{\beta / l}$.
  But the latter follows since $\lambda_r(S) \geq e^{\beta/ l}$.
  This is because $e^{b \cdot H}$ is symmetric and each entry in $H$ is
  $\geq \frac{1}{\lambda}$.
  Then we can use \cref{cor:block_diagonal}.
  Eckart-Young-Mirsky then completes the proof.
\end{proof}

\subsection{Scatterbrain: Analysis}
\label{sec:scatterbrain_analysis_proof}

Here we prove~\cref{thm:unbiased}, which shows that Scatterbrain approximation
is unbiased and analyses its variance.
We restate the theorem here for the reader's convenience.

\begin{theorem*}
  Define $\sigma(q, k) = \exp( q^\top k)$, $\hat{\sigma}^{\mathsf{pfe}}$ as Performer's estimator and $\hat{\sigma}^{\mathsf{sbe}}$ as Scatterbrain estimator.
  Denote $ {\cal S}^{d-1} \subset \mathbb{R}^d$ as the unit sphere. Suppose $q, k \in S^{d-1}$ are such that $\Vert q-k \Vert < \tau$.
  Then:
  \begin{align*}
    \mathbb{E}[\hat{\sigma}^{\mathsf{sbe}}(q, k)] = \sigma(q, k),
    \quad \mathrm{Var}[\hat{\sigma}^{\mathsf{sbe}}(q, k)] = (1-p) \cdot \mathrm{Var}[\hat{\sigma}^{\mathsf{pfe}}(q, k)] < \mathrm{Var}[\hat{\sigma}^{\mathsf{pfe}}(q, k)]
  \end{align*}
  where $p = \exp(-\frac{\tau^2}{4-\tau^2}\ln d-O_{\tau}(\ln\ln d))$.
\end{theorem*}

\begin{proof}
  Let $A_{ij} = \exp(q_k^\top k_j)$ be $ij$-entry of the unnormalized attention
  matrix, $A^{\mathrm{lr}}_{ij} = \phi(q_i)^\top \phi(k_j)$ the entry of the
  low-rank approximation (Performer), and let $A^{\mathrm{sb}}_{ij}$ be the
  entry of the Scatterbrain (sparse + low-rank) approximation.
  By the construction of the Scatterbrain attention matrix
  (\cref{eq:scatterbrain_subtract}), if $ij \in \mathcal{S}$, where
  $\mathcal{S}$ is the set of indices selected by the LSH, then:
  \begin{equation*}
    A^{\mathrm{sb}}_{ij} = (\wt{Q} \wt{K}^\top + S)_{ij} = \phi(q_i)^\top \phi(k_j) + \exp(q_i^\top k_j) - \phi(q_i)^\top \phi(k_j) = \exp(q_i^\top k_j).
  \end{equation*}
  If $ij \notin \mathcal{S}$, then
  \begin{equation*}
    A^{\mathrm{sb}}_{ij} = (\wt{Q} \wt{K}^\top + S)_{ij} = \phi(q_i)^\top \phi(k_j) + 0 = \phi(q_i)^\top \phi(k_j).
  \end{equation*}
  In other words, $A^{\mathrm{sb}}$ matches $A$ on the indices in $\mathcal{S}$,
  and matches $A^{\mathrm{lr}}$ on the indices not in $\mathcal{S}$.

  To show that $A^{\mathrm{sb}}$ is an unbiased estimator of $A$, we simply use
  the fact that $A^{\mathrm{lr}}$ is also an unbiased estimator of
  $A$~\citep[Lemma 1]{choromanski2020rethinking}:
  \begin{align*}
    \E[A^{\mathrm{sb}}_{ij}]
    &= \P(ij \in \mathcal{S}) \E[A_{ij} \mid ij \in \mathcal{S}] + \P(ij \notin \mathcal{S}) \E[A^{\mathrm{lr}}_{ij} \mid ij \notin \mathcal{S}] \\
    &= \P(ij \in \mathcal{S}) A_{ij} + \P(ij \notin \mathcal{S}) A_{ij} \\
    &= A_{ij}.
  \end{align*}
  In other words, $\E[\hat{\sigma}^{\mathsf{sbe}}(q, k)] = \sigma(q, k)$.

  Now we analyze the per-entry variance of $A^{\mathrm{sb}}$.
  Since $A^{\mathrm{sb}}$ is an unbiased estimator of $A$, by the law of
  total variance,
  \begin{align*}
    \var(A^{\mathrm{sb}}_{ij})
    &= \P(ij \in \mathcal{S}) \var(A_{ij} \mid ij \in \mathcal{S}) + \P(ij \notin \mathcal{S}) \var(A^{\mathrm{lr}}_{ij} \mid ij \notin \mathcal{S}) \\
    &= \P(ij \in \mathcal{S}) \cdot 0 + \P(ij \notin \mathcal{S}) \var(A^{\mathrm{lr}}_{ij} ) \\
    &= \P(ij \notin \mathcal{S}) \var(A^{\mathrm{lr}}_{ij} ).
  \end{align*}
  To compute the probability that the index $ij$ is not in $\mathcal{S}$ (i.e.,
  not selected by LSH), we use the standard bound on cross-polytope
  LSH~\citep[Theorem 1]{andoni2015practical}:
  \begin{equation*}
    p \defeq \P(ij \in \mathcal{S}) = \exp(-\frac{\tau^2}{4-\tau^2}\ln d-O_{\tau}(\ln\ln d)).
  \end{equation*}
  Therefore,
  \begin{equation*}
    \var(A^{\mathrm{sb}}_{ij}) = (1 - p) \var(A^{\mathrm{lr}}_{ij}) < \var(A^{\mathrm{lr}}_{ij}).
  \end{equation*}
  In other words, $\var[\hat{\sigma}^{\mathsf{sbe}}(q, k)] = (1-p) \cdot \var[\hat{\sigma}^{\mathsf{pfe}}(q, k)] < \var[\hat{\sigma}^{\mathsf{pfe}}(q, k)]$.

  More explicitly, by plugging in the variance of $A^{\mathrm{lr}}$~\citep[Lemma
  2]{choromanski2020rethinking}, we have
  \begin{equation*}
    \var(A^{\mathrm{sb}}_{ij}) = (1 - p) \frac{1}{m} \exp \left( \norm{q_i + k_j}^2 \right) \exp(2q_i^\top k_j)  \left( 1 - \exp \left( -\norm{q_i + k_j}^2 \right) \right),
  \end{equation*}
  where $p = \exp(-\frac{\tau^2}{4-\tau^2}\ln d-O_{\tau}(\ln\ln d))$

\end{proof}

\newpage
\section{Additional Experiments and Details}
\label{sec:experiment_details}

\subsection{Datasets}

\textbf{ImageNet~\citep{5206848}:} ImageNet is one of the most widely-used image classification benchmarks. In our experiments in Section~\ref{subsec:accurate} of evaluating the approximation accuracy of Scatterbrain, both BigGAN and Vision Transformer are pre-trained on this dataset. It has roughly 1.2 million training images and 50,000 validation images.

\textbf{WikiText103~\citep{merity2016pointer} and Copy~\citep{kitaev2020reformer}:} WikiText103 is a popular dataset for auto-regressive models. It is from a collection of over 100 million tokens extracted from the set of verified good and featured articles on Wikipedia. It has 28,475 training articles, 60 for validation and 60 for testing. 

Copy is a synthetic a synthetic sequence duplication task where inputs are of the form $0w0w$ and  $w\in\left\{0,...,N\right\}^*$. It is previously used in~\citep{kitaev2020reformer,chen2020mongoose}. This task is useful for demonstrating the effectiveness of long range attention: it requires non-local attention lookups. It cannot be solved by any model relying on sparse attention with a limited range such as, local attention.

\textbf{Long Range Arena (LRA)~\citep{tay2020long}:} This is a recent benchmark for evaluating efficient transformers with long input sequence. We used ListOps~\citep{nangia2018listops}, byte-level IMDb reviews text classification~\citep{maas2011learning},  byte-level document retrieval~\citep{radev2013acl},  image classification on sequences of pixels~\citep{krizhevsky2009learning} and Pathfinder~\citep{linsley2018learning}. We follow the same evaluation mechanism from~\citep{tay2020long} but implement our own version in Pytorch (like data loader).

\textbf{GlUE~\citep{wang2018glue}:} GLUE is a standard multi-task benchmark in NLP. It has single-sentence tasks, CoLA and SST-2; similarity and paraphrase tasks, MRPC, STS-B, QQP; and inference tasks, MNLI, QNLI, RTE and WNLI. For our additional experiments below (not enough space to be included in the main paper), we follow the tradition from~\citep{devlin2018bert, yang2019xlnet, daras2020smyrf} and truncate all the input sequences to 128 tokens. 

\subsection{Settings}

\textbf{BigGAN: }We adapt the same pre-trained BigGAN model from~\citep{daras2020smyrf} with no additional training. The model has a single attention layer at resolution $64 \times 64$ (4096). Similar to the prior work, we also replace its
full attention layer with Scatterbrain at the same resolution. Figure~\ref{fig:upperbound} in the main paper shows the best-effort comparison with [1/32, 1/16, 1/8, 1/4, 1/2] of the parameter budget. For example, if given parameter budget 1/2, we report the best performance of Smyrf from choice of 32/64/128 hash round 64/32/16 cluster size. 

\textbf{T2-ViT:} We use the pre-trained vision transformer model T2T-ViT-14 from~\citep{yuan2021tokens} with $224 \times 224$ image size. Without any additional training, we just replace the attention layer with Scatterbrain and other baselines and evaluate the approximation error and classification accuracy on ImageNet testings. Again, we report the best-effort best performance of each approximation given the certain parameter budget.

\textbf{Auto-regressive Model: } We follow the settings from the popular repo \url{https://github.com/NVIDIA/DeepLearningExamples} for training vanilla Transformer from scratch on WikiText103, except for chunking WikiText103 into sequence length 1024 in order to simulate long input sequences. The model is 16 layer with 8 head and 512 model dimension. We train all the models for 30 epochs and report the best Testing Perplexity. The model we use for Copy task is simply a 2-layer-4-head transformer and sequence length is also 1024. We make 5 runs and report average. Table~\ref{table:main_error_bar} presents the results with standard deviation.

\textbf{Classification Model: } We follow the model setting from~\citep{tay2020long, xiong2021nystr}. We share the same finding with~\citep{xiong2021nystr} that the acuracy for the Retrieval tasks is actually higher than reported in~\citep{tay2020long}.

\textbf{Ratio between Sparse and Low-rank components:} There are some rules that we used in our experiments to set this ratio. For inference, we set this ratio based on the entropy of an observed subset of attention matrices in different layers: we allocate more memory to the low-rank component compared to the sparse component if the entropy is high.
For training, generally allocating more memory budget to sparse tends to perform better, so in the experiment, we set the ratio to 3:1 (sparse: low-rank component) for simplicity.
Moreover, in future work, it could be useful to make this ratio adaptive during training. For example, in the early stage of the training and early layers, attention matrices are usually more uniform (higher entropy). Thus, the approximation error could be even lower if the ratio favors low-rank-based components. One approach could be to monitor the approximation error of sparse and low-rank components compared to full attention regularly and adjust the memory budget accordingly. We will add the above discussion to the updated manuscript.

\begin{table}[h]
    \caption{The performance of Scatterbrain, \textsc{reformer}, \textsc{performer} and Full-Attention on Long-Range-Arena benchmarks and 2 popular language modeling tasks. We fix the same number of parameters ($1/8$ of the full) used for approximating the attention matrix for each method.}
    \begin{minipage}{.45\linewidth}
      \centering
              \hspace{-1cm}
          	\resizebox{0.9\linewidth}{!}{
	\centering
	\begingroup
	\setlength{\tabcolsep}{10pt}
	\renewcommand{\arraystretch}{1.15}
	\begin{tabular}{c||c|c}
    \specialrule{.15em}{.05em}{.05em}
    \multirow{1}{*}{ {\bf Attention} } & \multicolumn{1}{c|}{\multirow{1}{*}{Copy (ppl)}} & \multicolumn{1}{c}{\multirow{1}{*}{WikiText-103 (ppl)}} \\
	\hline
	Full Attention& 1 & 25.258$\pm$0.37  \\
	\cline{1-3}
	Reformer& 6.8$\pm$0.64  & 27.68$\pm$0.53  \\
	Performer& 49$\pm$2.7  & 66$\pm$5.8   \\
	\cline{1-3}
	Scatterbrain& \textbf{2.58}$\pm$0.21 &\textbf{26.72}$\pm$0.44   \\
	\specialrule{.15em}{.05em}{.05em}
	\end{tabular}
		\endgroup
	}
    \end{minipage}%
    \begin{minipage}{.6\linewidth}
      \centering
        \hspace{-1.5cm}
            	\resizebox{\linewidth}{!}{
	\centering
	\Huge
	\begingroup
	\setlength{\tabcolsep}{10pt}
	\renewcommand{\arraystretch}{1.1}
	\begin{tabular}{c||c|c|c|c|c|c}
    \specialrule{.15em}{.05em}{.05em}
    \multirow{1}{*}{ {\bf Attention} }  &
    \multicolumn{1}{c|}{\multirow{1}{*}{ListOps}} &
    \multicolumn{1}{c|}{\multirow{1}{*}{Text}} & 
    \multicolumn{1}{c|}{\multirow{1}{*}{Retrieval}} & 
    \multicolumn{1}{c|}{\multirow{1}{*}{Image}} &
    \multicolumn{1}{c|}{\multirow{1}{*}{Pathfinder}}  & 
    \multicolumn{1}{c}{\multirow{1}{*}{Avg}}\\
	\hline
	\hline
	Full Attention& 38.2$\pm$0.17& 63.29$\pm$0.38& 80.85$\pm$0.12 & 41.78$\pm$0.26 & 73.98$\pm$0.31 & 59.62\\
	\cline{1-7}
	\hline
	\hline
	Reformer& 36.85$\pm$0.37& 58.12$\pm$0.42& 78.36$\pm$0.29 & 28.3$\pm$0.39 & 67.95$\pm$0.28 & 53.9 \\
	Performer& 35.75$\pm$0.29& 62.36$\pm$0.49 & 78.83$\pm$0.33 & 39.71$\pm$0.48 & 68.6$\pm$0.36 & 57.05 \\
	\cline{1-7}
	Scatterbrain& \textbf{38.6}$\pm$0.22 & \textbf{64.55}$\pm$0.34 & \textbf{80.22}$\pm$0.31 & \textbf{43.65}$\pm$0.46 & \textbf{69.91}$\pm$0.25 & \textbf{59.38} \\
	\specialrule{.15em}{.05em}{.05em}
	\end{tabular}
		\endgroup
	}
    \end{minipage} 
    \label{table:main_error_bar}
\end{table}

\afterpage{
\begin{figure}[h]
\captionsetup{font=small}
	\begin{center}
	\scriptsize
		\begin{tabular}{cc}
		    \hspace{-0.5cm}
			\includegraphics[width=0.43\linewidth]{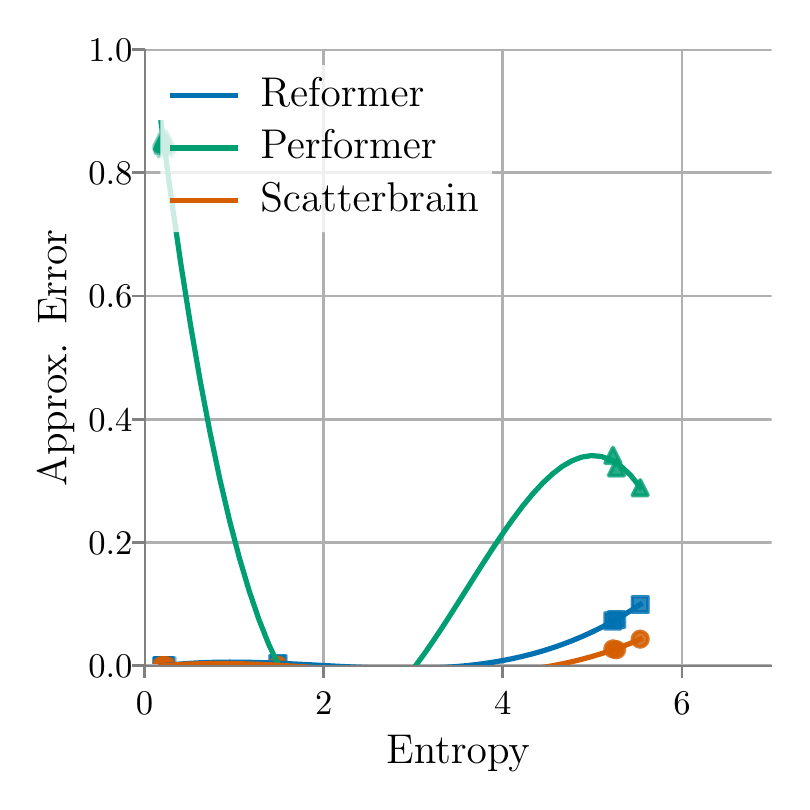}
			\includegraphics[width=0.43\linewidth]{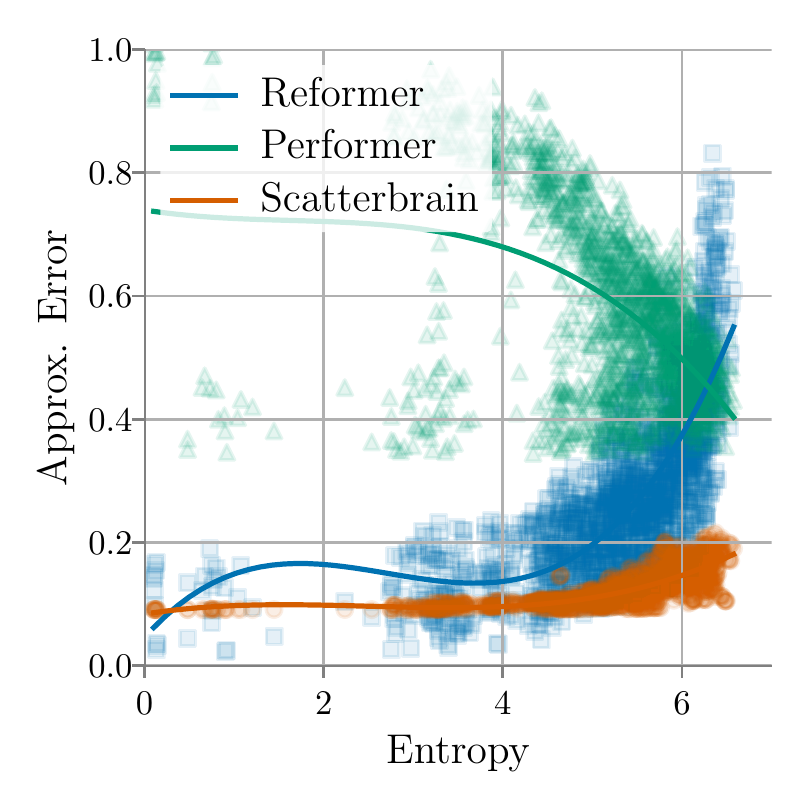}
			\\
			\hspace{-0.5cm}
			\includegraphics[width=0.43\linewidth]{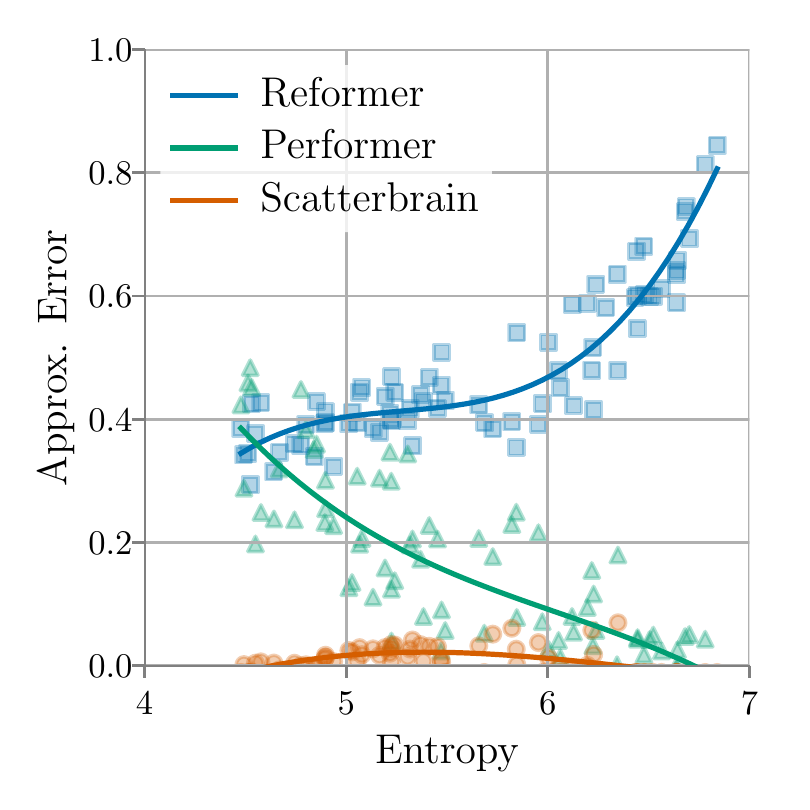}
			\includegraphics[width=0.43\linewidth]{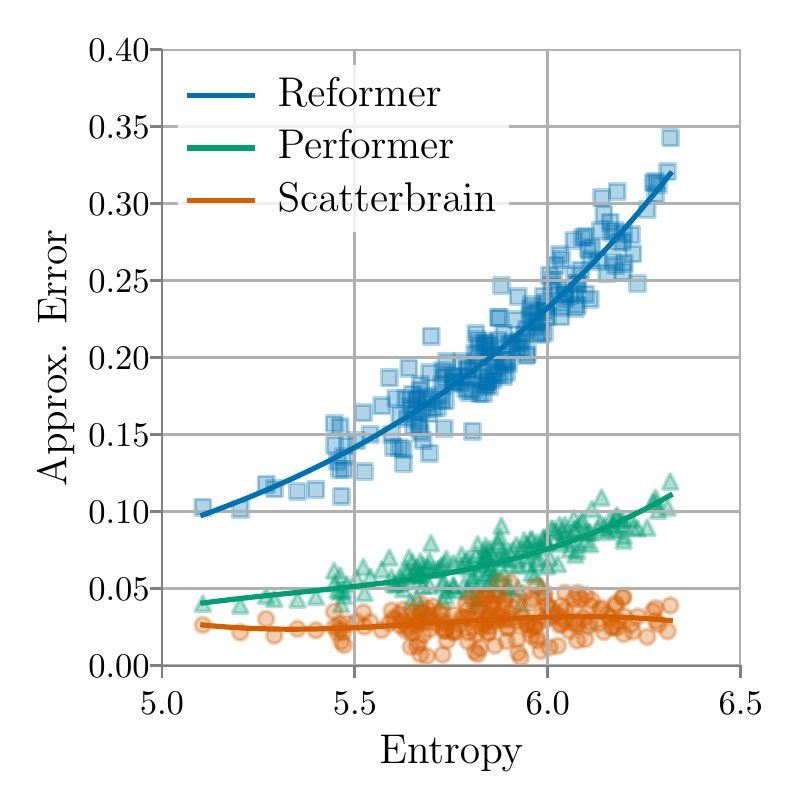}
		\end{tabular}
	\end{center}
	\caption{ Top two plots present Approximation Error vs. Entropy of attention matrices for \textsc{reformer}, \textsc{performer} and Scatterbrain on Copy (left) and WikiText103 (right). Bottom two plots present Approximation Error vs. Entropy of attention matrices for \textsc{reformer}, \textsc{performer} and Scatterbrain on Text-IMDb (left) and Image-Cifar10 (right). Recall we observe that entropy of the softmax attention distribution (i.e., scale of
      logits) determines the regimes where sparse, low-rank, or sparse +
      low-rank perform well. Scatterbrain yields better approximation than
      \textsc{reformer} or \textsc{performer} in most of the cases; \textsc{performer} performs the worst on language modeling tasks while \textsc{reformer} performs the worst on classification tasks. These plots for approximation error analysis match with their performance on downstream tasks.}
	\label{fig:error_analysis} 
\end{figure}
\clearpage
}

\subsection{More Ablation Studies}
\label{appendix:ablation}

\subsubsection{Memory Budget}
We present an ablation study on the parameter budget for the WikiText-103 language modeling task. We show that Scatterbrain outperforms its sparse and low-rank baselines across a range of parameter budgets. The results are presented in~\cref{table:memory}.

\textbf{Analysis:} We have observed that Scatterbrain outperforms its sparse and low-rank baselines under different memory budgets. Similar to what we found in~\cref{subsec:results}, Performer does not train stably even with $\frac{1}{4}$ of the full attention memory. However, under the Scatterbrain framework, Performer can be combined with Reformer in an elegant way to achieve the same accuracy while using only half of the memory and faster than Reformer by exploiting the sparse+low-rank structure in attention matrices.

\begin{table*}[ht]
    \scriptsize
\captionsetup{font=small}
	\centering
    	\caption{We run WikiText-103 LM with a sweep of 1/4, 1/8, 1/16 memory budget. We show the validation perplexity and speed-up with respect to the full attention with different efficient Attention layers.}
    	\resizebox{0.7\linewidth}{!}{
	\centering
	\begingroup
	\setlength{\tabcolsep}{10pt}
	\renewcommand{\arraystretch}{1.5}
	\begin{tabular}{c|c|c|c}
    \specialrule{.15em}{.05em}{.05em}
    \multirow{1}{*}{}& \multicolumn{1}{c|}{\multirow{1}{*}{$\frac{1}{4}$ Mem}} & 
    \multicolumn{1}{c|}{\multirow{1}{*}{$\frac{1}{8}$ Mem}}&
    \multicolumn{1}{c}{\multirow{1}{*}{$\frac{1}{16}$ Mem}}\\
    \cline{2-4}
    & Perplexity (Speed-up) & Perplexity & Perplexity    \\
		\hline
	\multirow{1}{*}{\textsc{Smyrf}} & 26.76 (1.6$\times$) & 27.68 (1.39$\times$) & 28.7(1.85$\times$) \\
	\cline{1-4}
	\multirow{1}{*}{\textsc{Performer}} & 58(2.13$\times$)	 & 66 (2.01$\times$) & 85(1.77$\times$) \\
	\cline{1-4}
	\multirow{1}{*}{Scatterbrain} & \textbf{26.26(1.58$\times$)} & \textbf{26.72 (1.87$\times$)} & \textbf{27.74(2.03$\times$)}\\
	\specialrule{.15em}{.05em}{.05em}
	\end{tabular}
	\endgroup
	}
	\label{table:memory}
\end{table*}

\subsubsection{Different Sparse and Low-rank baselines}
Scatterbrain is general enough to accommodate different kinds of sparse and low-rank approximations as its sub-components. In particular, we can combine Local attention or block sparse (from Sparse Transformer and BigBird) + Performer (instead of Reformer + Performer) in a similar fashion. The support of the sparse matrix S will thus be fixed and not adaptive to input, but all the other steps are exactly the same.

We have run additional experiments on the Local attention + Performer combination and BigBird. Recall that in~\cref{sec:experiment_details}, we have shown Scatterbrain can reduce the attention memory of Vision Transformer by 98\% at the cost of only 0.8\% drop of accuracy when serving as a drop-in replacement for full attention without training on ImageNet. We show the results for local+performer variation with the same memory budget in~\cref{table:vit2}.

We have also run additional experiments on Local attention on Copy and Wikitext-103 language modeling task (~\cref{table:local}). 
We see that Local attention is reasonably competitive on Wikitext-103 but does not perform well on Copy. The results are not surprising as noted in the Reformer paper that Copy requires non-local attention lookups.

    \begin{table*}
\vspace{-0.3cm}
\scriptsize
\captionsetup{font=small}
\caption{Top-1 Accuracy of pre-trained T2T Vision Transformer on ImageNet with different attention replacements. Error represents the average normalized approximation error to full attention.
}
\centering
\resizebox{0.5\linewidth}{!}{
\begin{tabular}{ c||c }
\specialrule{.15em}{.05em}{.05em}
Attention &Top-1 Acc  \\
\specialrule{.15em}{.05em}{.05em}
Full Attention & $81.7\%$  \\
SMYRF & $79.8\%$  \\
Local & $79.6\%$  \\
Performer & $80.1\%$  \\
BigBird & $80.3\%$ \\
Scatterbrain (Local + Performer) & $80.3\%$ \\
Scatterbrain (SMYRF + Performer) & \bf{80.7$\%$} \\
\specialrule{.15em}{.05em}{.05em}
\end{tabular}
}
\vspace{-0.3cm}
\label{table:vit2}
\end{table*}

\subsubsection{Different Sparse and Low-rank baselines}
\begin{table*}
      \caption{Additional experiments for Local attention on the Copy and Wikitext-103 language modeling task.}
      \centering
              \hspace{-1cm}
          	\resizebox{0.5\linewidth}{!}{
	\centering
	\begingroup
	\setlength{\tabcolsep}{10pt}
	\renewcommand{\arraystretch}{1.15}
	\begin{tabular}{c||c|c}
    \specialrule{.15em}{.05em}{.05em}
    \multirow{1}{*}{ {\bf Attention} } & \multicolumn{1}{c|}{\multirow{1}{*}{Copy (ppl)}} & \multicolumn{1}{c}{\multirow{1}{*}{WikiText-103 (ppl)}} \\
	\hline
	Full Attention& 1 & 25.258  \\
	\cline{1-3}
	Reformer& 6.8  & 27.68  \\
	Performer& 49  & 66   \\
	Local& 53  & 30.72   \\
	\cline{1-3}
	Scatterbrain& \textbf{2.58} &\textbf{26.72}  \\
	\specialrule{.15em}{.05em}{.05em}
	\end{tabular}
		\endgroup
	}
	\label{table:local}
    \end{table*}

\subsection{Analysis}
Recall in Section~\ref{sec:experiments}, we have reported the analysis after visualizing the error of \textsc{reformer} (sparse), \textsc{performer} (low-rank), and Scatterbrain (sparse + low-rank) given the same number of parameters when approximating the full attention matrices for each attention layer during training. In Figure~\ref{fig:error_analysis}, we present the visualization.

The conclusion for language modeling tasks is that sparse+low-rank has the smallest approximation error in most of the cases, and sparse has the largest error, which matches with the end-to-end results. It also confirms the observation in the popular benchmark paper~\citep{tay2020long} that kernel or low-rank based approximations are less effective for hierarchical structured data. For classification tasks, we again find that Scatterbrain has the smallest approximation error, while \textsc{performer} is the worst on ListOps and \textsc{reformer} has the largest error on classification tasks, which matches with the end-to-end results and confirms our observations earlier (sparse and low-rank approximation excel in different regimes).

\subsection{Additional Experiments of Fine-tuning Bert on GLUE}

We provide additional experiments of fine-tuning Bert on GLUE in Table~\ref{table:bert}. We follow the similar setting as~\citep{daras2020smyrf}. We replace all the attention layers in Bert base model with Scatterbrain and other baselines. Then we fine-tune Bert on 9 downstream tasks for 3 epochs with batch size 32 and learning rate 3e-5. The parameter budget is 1/2 of the full attention because sequence length 128 is not very long. We can see Scatterbrain outperforms all the other baselines in most of the downstream tasks.  

\begin{table*}[ht]
    \scriptsize
\captionsetup{font=small}
	\centering
    	\caption{Results of GLUE when replacing dense attention matrices with \textsc{smyrf}, \textsc{performer} and Scatterbrain in BERT base model. We fix the same number ofparameters (1/2 of the full) used for approximating the attention matrix for each method.}
    	\resizebox{\linewidth}{!}{
	\centering
	\begingroup
	\setlength{\tabcolsep}{10pt}
	\renewcommand{\arraystretch}{1.5}
	\begin{tabular}{c|c|c|c|c|c|c|c|c|c}
    \specialrule{.15em}{.05em}{.05em}
    \multirow{1}{*}{}& \multicolumn{1}{c|}{\multirow{1}{*}{CoLA}} & 
    \multicolumn{1}{c|}{\multirow{1}{*}{SST-2}}&
    \multicolumn{1}{c|}{\multirow{1}{*}{MRPC}}&
    \multicolumn{1}{c|}{\multirow{1}{*}{STS-B}}&
    \multicolumn{1}{c|}{\multirow{1}{*}{QQP}}&
    \multicolumn{1}{c|}{\multirow{1}{*}{MNLI}}&
    \multicolumn{1}{c|}{\multirow{1}{*}{QNLI}}&
    \multicolumn{1}{c|}{\multirow{1}{*}{RTE}}&
    \multicolumn{1}{c}{\multirow{1}{*}{WNLI}}\\
    \cline{2-10}
    & mcc & acc & acc & corr & acc & acc & acc & acc& acc   \\
		\hline
	\multirow{1}{*}{\textsc{Full}} & 0.576& 0.934 & 0.874 & 0.879  & 0.905 & 0.813 &  0.916 & 0.668 & 0.43 \\
	\cline{1-10}
	\multirow{1}{*}{\textsc{Smyrf}} & 0.538&  0.912& 0.833 & 0.856  & 0.898 &0.775 & 0.879  & \textbf{0.626}  & 0.412 \\
	\multirow{1}{*}{\textsc{Performer}} & 0.508 & 0.838 &0.782& 0.203 & 0.831 & 0.563 & 0.763& 0.556& \textbf{0.449}\\
	\cline{1-10}
	\multirow{1}{*}{Scatterbrain} & \textbf{0.569} & \textbf{0.927} & \textbf{0.863}& \textbf{0.867}& \textbf{0.902} & \textbf{0.813}& \textbf{0.893}&0.619 & 0.428\\
	\specialrule{.15em}{.05em}{.05em}
	\end{tabular}
	\endgroup
	}
	\label{table:bert}
\end{table*}
\section{Further Discussions and Future Work}

In this paper, we present Scatterbrain, unifying the strength of sparse and low-rank approximation. It is inspired by the observations on the attention matrix structures induced by the data and softmax function as well as the classical robust-PCA algorithm. In our implementation and analysis, we have \textsc{reformer}/Smyrf and \textsc{performer} as the back-bone for sparse and low-rank approximations because of their properties, e.g. Performer is unbiased. Scatterbrain is fundamentally a framework for combining the strength of sparse and low-rank variants, so it can be easily extended to other variants, such as Routing Transformer~\citep{roy2021efficient} or Nystromformer~\citep{xiong2021nystr}. Further more, our observations on the connection between entropy and low-rank/sparse approximation error also provide an opportunity for efficiently detecting the approximation or compression method to choose for different architectures or benchmarks.  

\end{document}